\documentclass[final,5p,twocolumn]{elsarticle}

\usepackage{amssymb}
\usepackage{amsmath}
\usepackage{booktabs}
\usepackage{graphicx}
\usepackage{subcaption}
\usepackage[export]{adjustbox}
\usepackage{lscape}
\usepackage{float}
\usepackage{arydshln}
\usepackage{pifont}   
\usepackage{multirow}
\usepackage{enumitem}
\usepackage[hyphens]{url}  
\usepackage{hyperref}
\usepackage{lineno}

\usepackage{tikz}
\usetikzlibrary{shapes.geometric}
\usepackage{xstring}
\usepackage{xcolor}
\usepackage{stfloats}

\newcommand{\cmark}{\ding{51}} 
\newcommand{\xmark}{\ding{55}} 

\makeatletter
\def\ps@pprintTitle{\let\@oddhead\@empty \let\@evenhead\@empty \let\@oddfoot\@empty \let\@evenfoot\@empty}
\makeatother

\begin{document}
\newcommand{\checkcolor}[1]{
    \IfStrEq{#1}{black}{
        \def\fillcolor{white} 
    }{
        \def\fillcolor{black} 
    }
}

\newcommand{\cblacksquare}[2][0.3]{
\begin{tikzpicture}[scale=#1]
    \filldraw[fill=#2, draw=black] (0,0) -- (1,0) -- (1,1) -- (0,1) -- (0,0);
\end{tikzpicture}
}

\newcommand{\cwhitesquare}[2][0.3]{
\begin{tikzpicture}[scale=#1]
 \draw[#2] (0,0) -- (1,0) -- (1,1) -- (0,1) -- (0,0);
\end{tikzpicture}
}

\newcommand{\cblacksquaredot}[2][0.3]{
\begin{tikzpicture}[scale=#1]
\checkcolor{#2}
 \filldraw[fill=#2, draw=black] (0,0) -- (1,0) -- (1,1) -- (0,1) -- (0,0);
 \fill[\fillcolor] (0.5,0.5) circle (4pt);
\end{tikzpicture}
}

\newcommand{\cwhitesquaredot}[2][0.3]{
\begin{tikzpicture}[scale=#1]
 \draw[#2] (0,0) rectangle (1,1);
 \filldraw[#2] (0.5,0.5) circle (4pt);
\end{tikzpicture}
}

\newcommand{\cblacktriangleup}[2][0.3]{
\begin{tikzpicture}[scale=#1]
    \filldraw[fill=#2, draw=black] (0,0) -- (1,0) -- (0.5, 1) -- cycle;
\end{tikzpicture}
}

\newcommand{\cwhitetriangleup}[2][0.3]{
\begin{tikzpicture}[scale=#1]
    \draw[#2] (0,0) -- (1,0) -- (0.5, 1) -- cycle;
\end{tikzpicture}
}

\newcommand{\cblacktriangleleft}[2][0.3]{
\begin{tikzpicture}[scale=#1, rotate=90]
    \filldraw[fill=#2, draw=black] (0,0) -- (1,0) -- (0.5, 1) -- cycle;
\end{tikzpicture}
}

\newcommand{\cwhitetriangleleft}[2][0.3]{
\begin{tikzpicture}[scale=#1, rotate=90]
    \draw[#2] (0,0) -- (1,0) -- (0.5, 1) -- cycle;
\end{tikzpicture}
}

\newcommand{\cblacktriangleright}[2][0.3]{
\begin{tikzpicture}[scale=#1, rotate=-90]
    \filldraw[fill=#2, draw=black] (0,0) -- (1,0) -- (0.5, 1) -- cycle;
\end{tikzpicture}
}

\newcommand{\cwhitetriangleright}[2][0.3]{
\begin{tikzpicture}[scale=#1, rotate=-90]
    \draw[#2] (0,0) -- (1,0) -- (0.5, 1) -- cycle;
\end{tikzpicture}
}

\newcommand{\cblacktriangledown}[2][0.3]{
\begin{tikzpicture}[scale=#1, rotate=180]
    \filldraw[fill=#2, draw=black] (0,0) -- (1,0) -- (0.5, 1) -- cycle;
\end{tikzpicture}
}

\newcommand{\cwhitetriangledown}[2][0.3]{
\begin{tikzpicture}[scale=#1, rotate=180]
    \draw[#2] (0,0) -- (1,0) -- (0.5, 1) -- cycle;
\end{tikzpicture}
}

\newcommand{\cwhitesquarex}[2][0.3]{
\begin{tikzpicture}[scale=#1]
 \draw[#2] (0,0) -- (1,0) -- (1,1) -- (0,1) -- (0,0);
 \draw[#2] (1,0) -- (0,1);
 \draw[#2] (0,0) -- (1,1);
\end{tikzpicture}
}

\newcommand{\cblacksquarex}[2][0.3]{
\begin{tikzpicture}[scale=#1]
\checkcolor{#2}
 \filldraw[fill=#2, draw=black] (0,0) -- (1,0) -- (1,1) -- (0,1) -- (0,0);
 \draw[\fillcolor, line width=0.4mm] (1,0) -- (0,1);
 \draw[\fillcolor, line width=0.4mm] (0,0) -- (1,1);
\end{tikzpicture}
}

\newcommand{\cwhitesquarecross}[2][0.3]{
\begin{tikzpicture}[scale=#1]
 \draw[#2] (0,0) -- (1,0) -- (1,1) -- (0,1) -- (0,0);
 \draw[#2] (1,0.5) -- (0,0.5);
 \draw[#2] (0.5,0) -- (0.5,1);
\end{tikzpicture}
}

\newcommand{\cblacksquarecross}[2][0.3]{
\begin{tikzpicture}[scale=#1]
\checkcolor{#2}
 \filldraw[fill=#2, draw=black] (0,0) -- (1,0) -- (1,1) -- (0,1) -- (0,0);
 \draw[\fillcolor, line width=0.4mm] (1,0.5) -- (0,0.5);
 \draw[\fillcolor, line width=0.4mm] (0.5,0) -- (0.5,1);
\end{tikzpicture}
}

\newcommand{\cwhitediamondcross}[2][0.3]{
\begin{tikzpicture}[scale=#1]
 \draw[#2] (0,0.5) -- (0.5, 1) -- (1, 0.5) -- (0.5, 0) -- (0,0.5);
 \draw[#2] (1,0.5) -- (0,0.5);
 \draw[#2] (0.5,0) -- (0.5,1);
\end{tikzpicture}
}

\newcommand{\cblackdiamondcross}[2][0.3]{
\begin{tikzpicture}[scale=#1]
\checkcolor{#2}
 \filldraw[fill=#2, draw=black] (0,0.5) -- (0.5, 1) -- (1, 0.5) -- (0.5, 0) -- (0,0.5);
 \draw[\fillcolor, line width=0.4mm] (1,0.5) -- (0,0.5);
 \draw[\fillcolor, line width=0.4mm] (0.5,0) -- (0.5,1);
\end{tikzpicture}
}

\newcommand{\cwhitediamondx}[2][0.3]{
\begin{tikzpicture}[scale=#1]
 \draw[#2] (0,0.5) -- (0.5, 1) -- (1, 0.5) -- (0.5, 0) -- (0,0.5);
 \draw[#2] (0.25,0.75) -- (0.75,0.25);
 \draw[#2] (0.25,0.25) -- (0.75,0.75);
\end{tikzpicture}
}

\newcommand{\cblackdiamondx}[2][0.3]{
\begin{tikzpicture}[scale=#1]
\checkcolor{#2}
 \filldraw[fill=#2, draw=black] (0,0.5) -- (0.5, 1) -- (1, 0.5) -- (0.5, 0) -- (0,0.5);
 \draw[\fillcolor, line width=0.4mm] (0.25,0.75) -- (0.75,0.25);
 \draw[\fillcolor, line width=0.4mm] (0.25,0.25) -- (0.75,0.75);
\end{tikzpicture}
}

\newcommand{\cblackcircledot}[2][0.3]{
\begin{tikzpicture}[scale=#1]
\checkcolor{#2}
 \filldraw[fill=#2] (0.5,0.5) circle (14pt);
 \filldraw[\fillcolor] (0.5, 0.5) circle (4pt);
\end{tikzpicture}
}

\newcommand{\cwhitecircledot}[2][0.3]{
\begin{tikzpicture}[scale=#1]s
 \draw[#2] (0.5,0.5) circle (14pt);
 \filldraw[#2] (0.5, 0.5) circle (4pt);
\end{tikzpicture}
}

\newcommand{\cblackcircle}[2][0.3]{
\begin{tikzpicture}[scale=#1]
 \filldraw[fill=#2] (0.5,0.5) circle (14pt);
\end{tikzpicture}
}

\newcommand{\cwhitecircle}[2][0.3]{
\begin{tikzpicture}[scale=#1]
 \draw[#2] (0.5,0.5) circle (14pt);
\end{tikzpicture}
}

\newcommand{\cblackstartriangleup}[2][0.3]{
\begin{tikzpicture}[scale=#1]
    \filldraw[fill=#2] (0,0) to[out=20, in=160] (1,0) to[out=160, in=90] (0.5, 0.9) to[out=90, in=20] cycle;
\end{tikzpicture}
}

\newcommand{\cwhitestartriangleup}[2][0.3]{
\begin{tikzpicture}[scale=#1]
    \draw[#2] (0,0) to[out=20, in=160] (1,0) to[out=160, in=90] (0.5, 0.9) to[out=90, in=20] cycle;
\end{tikzpicture}
}

\newcommand{\cblackstartriangleupdot}[2][0.3]{
\begin{tikzpicture}[scale=#1]
\checkcolor{#2}
    \filldraw[fill=#2] (0,0) to[out=20, in=160] (1,0) to[out=160, in=90] (0.5, 0.9) to[out=90, in=20] cycle;
    \filldraw[\fillcolor] (0.5,0.4) circle (3pt);
\end{tikzpicture}
}

\newcommand{\cwhitestartriangleupdot}[2][0.3]{
\begin{tikzpicture}[scale=#1]
    \draw[#2] (0,0) to[out=20, in=160] (1,0) to[out=160, in=90] (0.5, 0.9) to[out=90, in=20] cycle;
    \filldraw[#2] (0.5,0.4) circle (3pt);
\end{tikzpicture}
}

\newcommand{\cwhitestartriangledown}[2][0.3]{
\begin{tikzpicture}[scale=#1]
    \draw[#2] (0,1) to[out=-20, in=-160] (1,1) to[out=-160, in=65] (0.5, 0) to[out=115, in=-20] cycle;
\end{tikzpicture}
}

\newcommand{\cblackstartriangledown}[2][0.3]{
\begin{tikzpicture}[scale=#1]
    \filldraw[fill=#2] (0,1) to[out=-20, in=-160] (1,1) to[out=-160, in=65] (0.5, 0) to[out=115, in=-20] cycle;
\end{tikzpicture}
}

\newcommand{\cwhitestartriangledowndot}[2][0.3]{
\begin{tikzpicture}[scale=#1]
    \draw[#2] (0,1) to[out=-20, in=-160] (1,1) to[out=-160, in=65] (0.5, 0) to[out=115, in=-20] cycle;
    \fill[fill=#2] (0.5,0.6) circle (3pt);
\end{tikzpicture}
}

\newcommand{\cblackstartriangledowndot}[2][0.3]{
\begin{tikzpicture}[scale=#1]
\checkcolor{#2}
    \filldraw[fill=#2] (0,1) to[out=-20, in=-160] (1,1) to[out=-160, in=65] (0.5, 0) to[out=115, in=-20] cycle;
    \filldraw[\fillcolor] (0.5,0.6) circle (3pt);
\end{tikzpicture}
}

\newcommand{\cblackstar}[2][0.7]{
\begin{tikzpicture}
    \node[star, star points=5, star point ratio=2.25, fill=#2, draw=black, scale=#1] at (0,0) {};
\end{tikzpicture}
}

\newcommand{\cwhitestar}[2][0.7]{
\begin{tikzpicture}
    \node[star, star points=5, star point ratio=2.25, draw=#2, scale=#1] at (0,0) {};
\end{tikzpicture}
}

\begin{frontmatter}

\title{Prompting with the human-touch: \\ evaluating model-sensitivity of foundation models for musculoskeletal CT segmentation}

\author[inst1,inst2,inst3]{Caroline Magg}
\author[inst2,inst3]{Maaike A. ter Wee}
\author[inst2]{Johannes G.G. Dobbe}
\author[inst2]{Geert J. Streekstra}
\author[inst3]{Leendert Blankevoort}
\author[inst1]{Clara I. S\'anchez}
\author[inst1]{Hoel Kervadec}

\affiliation[inst1]{organization={Quantitative Healthcare Analysis (QurAI) Group, University of Amsterdam},
addressline={Science Park 900},
city={Amsterdam},
postcode={1098 XH},
country={The Netherlands}}
\affiliation[inst2]{organization={Department Biomedical Engineering and Physics, Amsterdam UMC},
addressline={Meibergdreef 9},
city={Amsterdam},
postcode={1105 AZ},
country={The Netherlands}}
\affiliation[inst3]{organization={Department Orthopaedics, Amsterdam UMC},
addressline={Meibergdreef 9},
city={Amsterdam},
postcode={1105 AZ},
country={The Netherlands}}

\begin{abstract}
Promptable Foundation Models (FMs), initially introduced for natural image segmentation, have also revolutionized medical image segmentation. The increasing number of models, along with evaluations varying in datasets, metrics, and compared models, makes direct performance comparison between models difficult and complicates the selection of the most suitable model for specific clinical tasks.
In our study, 11 promptable FMs are tested using non-iterative 2D and 3D prompting strategies on a private and public dataset focusing on bone and implant segmentation in four anatomical regions (wrist, shoulder, hip and lower leg). The Pareto-optimal models are identified and further analyzed using human prompts collected through a dedicated observer study. 
Our findings are: 
1) The segmentation performance varies a lot between FMs and prompting strategies;
2) The Pareto-optimal models in 2D are SAM and SAM2.1, in 3D nnInteractive and Med-SAM2; 
3) Localization accuracy and rater consistency vary with anatomical structures, with higher consistency for simple structures (wrist bones) and lower consistency for complex structures (pelvis, tibia, implants);
4) The segmentation performance drops using human prompts, suggesting that performance reported on ``ideal'' prompts extracted from reference labels might overestimate the performance in a human-driven setting;
5) All models were sensitive to prompt variations. While two models demonstrated intra-rater robustness, it did not scale to inter-rater settings.
We conclude that the selection of the most optimal FM for a human-driven setting remains challenging, with even high-performing FMs being sensitive to variations in human input prompts.
Our code base for prompt extraction and model inference is available: \url{https://github.com/CarolineMagg/segmentation-FM-benchmark/}
\end{abstract}


\end{frontmatter}

\section{Introduction}\label{sec:introduction}

Foundation models (FMs) for medical image segmentation have gained significant attention as a promising paradigm for developing promptable methods, which allow users to guide segmentation through simple interactions such as selected points, bounding boxes or scribbles (i.e. prompts). Inspired by the Segment Anything Model (SAM) \cite{kirillov2023sam}, a growing number of medical variants tried to transfer the benefits of broad generalization and promptable inference to the medical domain. These models aim to reduce annotation burden, accelerate data curation, and enhance clinical usability by human-guided interactions and refinements.

Despite extensive interest and many evaluation efforts, most studies rely on synthetic or algorithmically generated prompts based on reference segmentation masks. While these represent ``ideal'' prompts, they fail to account for the inherent variability of human annotations. This results in a \emph{limited understanding of real-world prompting behavior}, where prompts are not ``perfect'' but still correct, provided by humans with varying levels of expertise and experience. Consequently, a significant gap exists in analyzing how human-generated prompt variability impacts the final segmentation performance. To address this gap, we conducted an observer study to collect and analyze human prompts. This allowed us to quantify intra- and inter-rater variability and, more importantly, to evaluate \emph{model sensitivity to input prompt variations}.

In addition to addressing the limited understanding of real-world prompt behavior, this study incorporates specific experimental design choices to overcome three critical challenges in the current evaluation landscape:
\paragraph{Challenge 1 -- Scalability of multi-rater evaluations} The growing number of promptable FMs makes exhaustive benchmarking across all available architectures computationally demanding, especially when accounting for multiple human-annotator prompt sets. Furthermore, analyzing the sensitivity of under-performing models offers limited insight. 
\textit{Solution:} We implemented a two-stage evaluation strategy. First, 11 models were compared using standardized ``ideal'' prompts to identify the Pareto-optimal models with the least model parameters, offering the best trade-offs between segmentation performance and parameter efficiency. Second, analysis with human prompts was focused exclusively on these top-performing models, ensuring that our sensitivity evaluation targeted the most relevant candidates.

\paragraph{Challenge 2 -- Benchmarking fairness and data contamination} While public datasets drive progress in the field, many medical FMs are trained on publicly available datasets, making it difficult to compare models fairly when the same data cannot be reused for testing. As a result, ensuring fair, unbiased evaluation often depends on private datasets, as these provide the necessary independence. At the same time, private datasets hinder reproducibility of a study, creating an inherent tension between the need for unbiased assessment and the desire for open, community-verifiable research.
\textit{Solution:} We utilize a hybrid data strategy. By combining private, task-specific data for independent assessment with public data, our study maintains a balance between independent validation and scientific reproducibility.

\paragraph{Challenge 3 -- Disconnect between benchmark dataset diversity and clinical requirements} While recent large-scale benchmarks aim to showcase the generalization of FMs across diverse datasets \cite{HUANG2024eval}, they often lack the depth required to validate performance on specialized clinical tasks. In clinical practice, the integration of a model depends on its performance for specific tasks, such as wrist bone segmentation for osteoarthritis assessment \cite{tenBerg2017wrist}, tibia and implant segmentation for loosening quantification \cite{magg2025knee}, or shoulder joint analysis for humeral head positioning \cite{verweij2024shoulder}. Evaluating models across heterogeneous tasks can dilute the focus on these task-specific requirements.
\textit{Solution:} Rather than distributing efforts across many modalities and heterogeneous tasks, we performed a targeted, task-focused investigation on musculoskeletal (MSK) CT scans for bone and implant segmentation.

Interested in FM performance in human-driven settings, our work contributes an extensive evaluation of FMs in bone and implant segmentation that moves beyond idealized simulations. By (1) integrating the human-in-the-loop variability, (2) using both public and private data, and (3) focusing on clinically relevant MSK tasks, we provide a more realistic assessment of FM performance.
We make our code base for prompt extraction and model inference publicly available\footnote{\url{https://github.com/CarolineMagg/segmentation-FM-benchmark/}}.

\section{Related Work}\label{sec:related_work} 

\paragraph{Segment Anything Model} The Segment Anything Model (SAM) \cite{kirillov2023sam} enables image segmentation from sparse or dense prompts (bounding boxes, positive and negative points, masks) using three components: an image encoder, a prompt encoder, and a lightweight mask decoder that fuses image and prompt embeddings into a binary mask. Applied to 3D medical scans, SAM operates slice-by-slice and requires a prompt for each slice.
SAM2 \cite{ravi2024sam2} extends SAM to video by replacing the image encoder backbone and adding a memory attention module that merges image embeddings, prompt encodings, and predicted masks into a joint representation. Stored in a first-in, first-out (FIFO) memory bank, this representation is queried when segmenting neighboring video frames. Treating CT slices as video frames allows SAM2 to propagate information across slices, enabling volumetric segmentation without prompting each slice individually.
Recently, SAM3 \cite{carion2025sam3} was released as concept-driven foundation model that unifies  image, video, and volumetric segmentation and object tracking by using short noun phrases or image examples (i.e., concept) instead of geometric prompts as used in SAM and SAM2.

\paragraph{Medical Foundation Models} A wide range of geometric prompt-based interactive FMs have been proposed for medical image segmentation: Spanning from fine-tuned 2D SAM variants (Med-SAM \cite{jun2024medsam}, SAM-Med2D \cite{cheng2023sammed2d}, ScribblePrompt \cite{wong2024scribbleprompt}, MedicoSAM \cite{archit2025medicosam}) and fine-tuned 3D SAM2-based models (Medical-SAM2 \cite{zhu2024medicalsam2}, Med-SAM2 \cite{ma2025medsam2}) to SAM-based extensions to 3D (SAM-Med3D \cite{wang2024sammed3d}, SegVol \cite{du2025segvol}), as well as non-SAM CNN-based methods like Vista3D \cite{he2024vista3d} and nnInteractive \cite{isensee2025nninteractive}. For a comprehensive overview of medical foundation models, their variations and applications, we refer the reader to the dedicated literature \cite{azad2023review, zhang2024samformis, noa2025review, ali2025review, koishiyeva2025review, liang2025review}.

\paragraph{Independent evaluation studies for Promptable Foundation Models} Since the release of SAM, multiple evaluation studies \cite{roy2023sammd, he2023eval, MAZUROWSKI2023eval, HUANG2024eval, cheng2023eval, mattjie2023eval, dong2024eval, sengupta2024eval, yu2024eval, shen2024eval} have shown that the performance of SAM-based models varies widely across datasets and tasks - generally favoring large, well-defined structures while struggling with small, irregular, or low-contrast ones. Most of these studies assessed only a limited set of models and prompts. Ali et al. \cite{ali2025review} complement these findings by examining SAM, MedSAM, and SAM-Med2D in fine-tuning scenarios, showing that fine-tuning and prompt optimization improves performance for Automated Breast Ultrasound (ABUS) tumor dataset and a pregnant pelvis MRI dataset.
Magg et al. \cite{magg2025zeroshot} evaluated bone CT segmentation employing four 2D SAM-based models (SAM, SAM2, Med-SAM, SAM-Med2D) and 32 prompting strategies, finding that bounding box and combination of bounding box with center point yielded the best performance across models. 
Noh et al. \cite{noa2025review} provide a broader comparison of seven foundation models for medical image segmentation (SAM, Med-SAM, SAM-Med2D, UniverSeg, SAM-Med3D, SegVol, and SAT-Pro), evaluating visual, text, and reference prompts across diverse datasets.
The RadioActive benchmark \cite{ulrich2025radioactive} focuses its evaluation on 3D interactive segmentation, testing seven models (SAM, SAM2, Med-SAM, SAM-Med2D, SAM-Med3D, SegVol, and ScribblePrompt) on CT and MRI data under an iterative refinement workflow. Its findings indicate that SAM2 outperforms all assessed 2D and 3D medical foundation models, and that bounding box prompts are generally superior to point-based ones. 
All named studies in this section relied on synthetic and algorithmically generated prompts, based on an available reference label. 

\section{Methodology}\label{sec:method}

\subsection{Promptable Foundation Models}

Eleven foundation models that, to our knowledge, were available as of July 30, 2025 -- while supporting training- and adaption-free open-set medical image segmentation using sparse geometric prompts (i.e., bounding boxes and points) --were included in our study (see \ref{sec:appendix_models} for implementation details).
The models were divided into four categories per prompt type based on prediction dimensionality (2D vs. 3D) and training data domain (medical vs natural images)(see Table \ref{tab:model_prompt_overview} for a model overview): 
\begin{itemize}[noitemsep, topsep=0pt]
    \item \textbf{2D FM trained on natural images}: SAM \& SAM2.1 2D
    \item \textbf{2D FM trained on medical images}: Med-SAM, SAM-Med2D, ScribblePrompt, MedicoSAM 2D
    \item \textbf{3D FMs trained on natural images}: SAM2.1 3D 
    \item \textbf{3D FMs trained on medical images}: SAM-Med3D, SegVol, MedicoSAM 3D, Vista3D, nnInteractive, Med-SAM2
\end{itemize}
Two sources of prompts were used: 1.) Automatically extracted prompts generated based on the \emph{reference mask}, i.e., following previous work \cite{magg2025zeroshot}, called reference prompts for short; 2.) Human-generated prompts created by participants of the observer study following annotation guidelines aligned with the automatic extraction procedure, called \emph{human prompts} for short. To ensure consistency, both prompts were derived on the same selected slices.

\begin{table*}[ht]
\centering

\caption{Overview of promptable FMs: Model backbone architecture, prediction dimensionality (2D vs. 3D), training data domain (Medical vs. Natural) and the supported prompting strategies. \newline
{\protect\scriptsize The prompting strategies are: single (1) or multiple ($N_P$) boxes, points, and their combinations, for single (1) or multiple ($N_S$) slices, with or without volumetric limitations (for 3D predictions). Boxed settings are our default settings, as they are possible across different models. (\cmark)* denotes that authors explicitly stated that the test set of \cite{wasserthal2023totalsegmentator} was excluded from training. $N_P$ prompts denotes that multiple prompts (in our work, up to 5 prompts) per initial slice were used. $N_S$ slices denotes that multiple initial slices (in our work, all selected slices) were used.}}
\label{tab:model_prompt_overview}

\setlength{\tabcolsep}{3pt}
\renewcommand{\arraystretch}{1.3}
    \begin{tabular}{l: l c c c :c c c c c}
    \toprule
    \textbf{Model} & \textbf{Arch.} & \textbf{Dim.} & \multicolumn{2}{c:}{\textbf{Data}} &
    \textbf{Box} & \textbf{Point} & \textbf{Pt+Box} & \textbf{Slice} & \textbf{Vol.} \\
    &                &                     & Domain & \textbf{\cite{wasserthal2023totalsegmentator}} & \textbf{(1/$N_P$)} & \textbf{(1/$N_P$)} & \textbf{(1/$N_P$)} & \textbf{(1/$N_S$)} & \textbf{Limits} \\
    \hline
    \textsc{Sam}  \cite{kirillov2023sam} & ViT & 2D & N & \xmark & \cmark/\fbox{\cmark} & \cmark/\fbox{\cmark} & \cmark/\fbox{\cmark} & - & - \\
    \textsc{Sam}2 2D  \cite{ravi2024sam2} & Hiera & 2D & N & \xmark & \cmark/\fbox{\cmark} & \cmark/\fbox{\cmark} & \cmark/\fbox{\cmark} & - & - \\
    \hdashline
    Med-\textsc{Sam} \cite{jun2024medsam} & \textsc{Sam} & 2D & M & \cmark & \cmark/\fbox{\cmark} & \textcolor{black!60}{\xmark}/\textcolor{black!60}{\xmark} & \textcolor{black!60}{\xmark}/\textcolor{black!60}{\xmark} & - & - \\
    \textsc{Sam}-Med2D \cite{cheng2023sammed2d} & \textsc{Sam} & 2D & M & \cmark & \cmark/\fbox{\cmark} & \cmark/\fbox{\cmark} & \cmark/\fbox{\cmark} & - & - \\
    ScribblePrompt-U \cite{wong2024scribbleprompt} & UNet & 2D & M & (\cmark)* & \cmark/\fbox{\cmark} & \cmark/\fbox{\cmark} & \cmark/\fbox{\cmark} & - & - \\
    ScribblePrompt-\textsc{Sam} \cite{wong2024scribbleprompt} & \textsc{Sam} & 2D & M & (\cmark)* & \cmark/\fbox{\cmark} & \cmark/\fbox{\cmark} & \cmark/\textcolor{black!60}{\xmark} & - & - \\
    Medico\textsc{Sam} 2D \cite{archit2025medicosam} & \textsc{Sam} & 2D & M & \cmark & \cmark/\fbox{\cmark} & \cmark/\fbox{\cmark} & \cmark/\fbox{\cmark} & - & - \\
    \hline
    \textsc{Sam}2 3D  \cite{ravi2024sam2} & Hiera & 3D & N & \xmark & \fbox{\cmark}/\textcolor{black!60}{\xmark} & \fbox{\cmark}/\cmark & \fbox{\cmark}/\textcolor{black!60}{\xmark} & \fbox{\cmark}/\cmark & \cmark \\
    \hdashline
    \textsc{Sam}-Med3D \cite{wang2024sammed3d} & 3D ViT & 3D & M & (\cmark)* & \textcolor{black!60}{\xmark}/\textcolor{black!60}{\xmark} & \fbox{\cmark}/\cmark & \textcolor{black!60}{\xmark}/\textcolor{black!60}{\xmark} & \fbox{\cmark}/\textcolor{black!60}{\xmark} & \textcolor{black!60}{\xmark} \\
    SegVol  \cite{du2025segvol} & 3D ViT & 3D & M & \cmark & \textcolor{black!60}{\xmark}/\textcolor{black!60}{\xmark} & \fbox{\cmark}/\cmark & \textcolor{black!60}{\xmark}/\textcolor{black!60}{\xmark} & \fbox{\cmark}/\cmark & \textcolor{black!60}{\xmark} \\
    Medico\textsc{Sam} 3D  \cite{archit2025medicosam} & \textsc{Sam} & 3D & M & \cmark & \fbox{\cmark}/\textcolor{black!60}{\xmark} & \fbox{\cmark}/\cmark & \fbox{\cmark}/\textcolor{black!60}{\xmark} & \fbox{\cmark}/\textcolor{black!60}{\xmark} & \textcolor{black!60}{\xmark} \\
    Vista3D \cite{he2024vista3d} & SegResNet & 3D & M & \cmark & \textcolor{black!60}{\xmark}/\textcolor{black!60}{\xmark} & \fbox{\cmark}/\cmark & \textcolor{black!60}{\xmark}/\textcolor{black!60}{\xmark} & \fbox{\cmark}/\cmark & \textcolor{black!60}{\xmark} \\
    nnInteractive \cite{isensee2025nninteractive} & CNN & 3D & M & \cmark & \fbox{\cmark}/\cmark & \fbox{\cmark}/\cmark & \fbox{\cmark}/\textcolor{black!60}{\xmark} & \fbox{\cmark}/\cmark & \textcolor{black!60}{\xmark} \\
    Med-\textsc{Sam}2 \cite{ma2025medsam2} & \textsc{Sam}2 & 3D & M & \cmark & \fbox{\cmark}/\textcolor{black!60}{\xmark} & \textcolor{black!60}{\xmark}/\textcolor{black!60}{\xmark} & \textcolor{black!60}{\xmark}/\textcolor{black!60}{\xmark} & \fbox{\cmark}/\cmark & \cmark \\
    \hline
    \end{tabular}
\end{table*}

\begin{figure}[h]
    \centering
    \begin{subfigure}[b]{0.12\textwidth}
        \includegraphics[width=\textwidth]{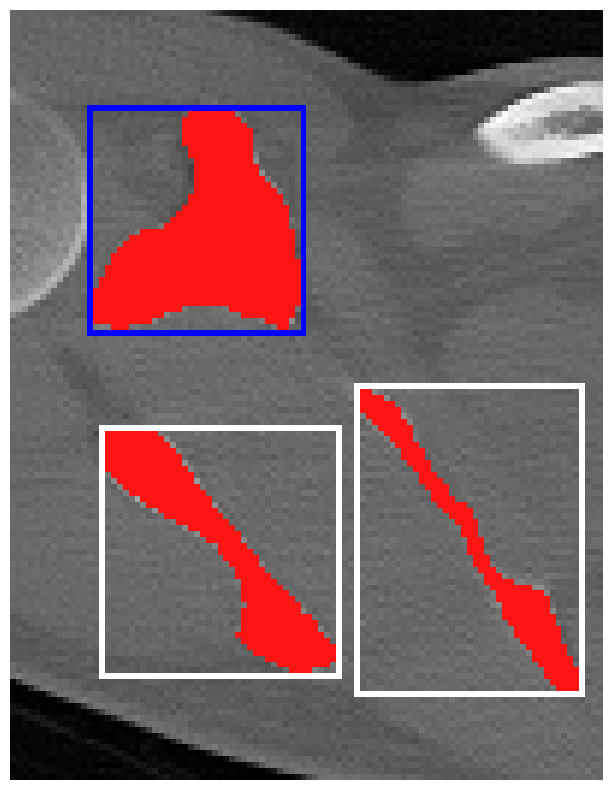}
        \caption{Bounding Box}
        \label{fig:ct_original}
    \end{subfigure}
    \hfill
    \begin{subfigure}[b]{0.12\textwidth}
        \includegraphics[width=\textwidth]{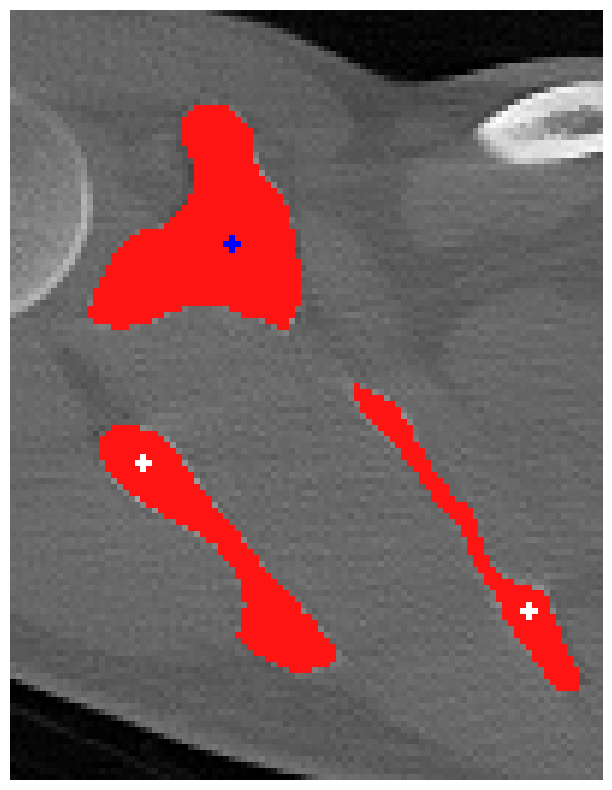}
        \caption{Center Point}
        \label{fig:ct_overlay}
    \end{subfigure}
    \hfill
    \begin{subfigure}[b]{0.2\textwidth}
        \includegraphics[width=\textwidth, trim={5cm 3.5cm 5cm 4cm},clip]{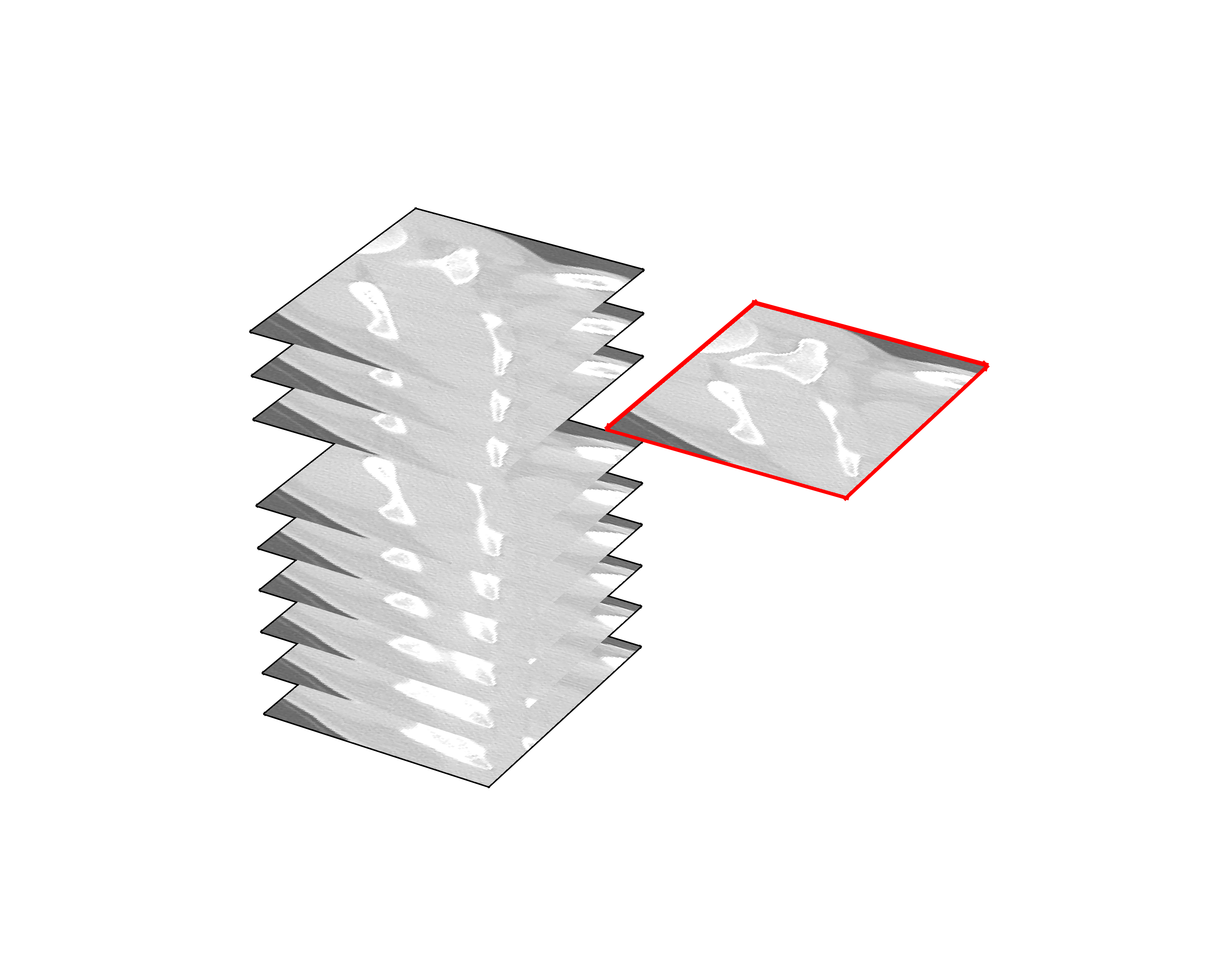}
        \caption{Slice Selection}
        \label{fig:ct_stack}
    \end{subfigure}
    \caption{Prompting strategies in 2D consist of prompt primitives, i.e., bounding box (a) and/or center point (b), and component selections, i.e., including prompts from either the largest component (blue prompt) or all components (white and blue prompts). The 3D prompting strategies extend this concept with slice selection (c).}
    \label{fig:prompting_strategies}
\end{figure}

\paragraph{Prompting Strategies in 2D} A \textit{2D prompting strategy} can be constructed with a \emph{primitive} and a \emph{component selection} criterion \cite{magg2025zeroshot} (Figure \ref{fig:prompting_strategies}). \emph{Primitives} are the building blocks of a prompt and in our work, the bounding box (referred to as \texttt{bbox} or box) and center point (referred to as \texttt{center} or point) are chosen, due to their demonstrated strong performance \cite{magg2025zeroshot} and the ability to compare reference and human prompts.
Following \cite{magg2025zeroshot}, the bounding box is defined as the tightest box enclosing the object, and the center point is defined as the pixel furthest away from the object boundary based on the Euclidean distance transform. This definition was used for the automatic prompt extraction and in the annotation guidelines for the observer study. The \emph{component selection} determines how many components of an anatomical structure are considered for the extraction of prompt primitives. Although anatomical structures may form a single 3D object, they can appear as multiple disconnected regions in individual 2D slices. Based on our dataset characteristics, up to $5$ components were considered for reference prompt extraction (referred to as $N_P$ prompts). Thus, for 2D prompting strategy, the default settings are: bounding box(\cblacksquare[0.2]{black}), center point(\cblackcircledot[0.25]{black}) or their combination(\cblacksquaredot[0.25]{black}), extracted for up to 5 components of the object of interest (Table \ref{tab:model_prompt_overview}).
\paragraph{Prompting Strategies in 3D} Models (except SegVol \cite{du2025segvol}) rely on pseudo-3D boxes defined by two coordinates representing a box in a 2D slice. Similarly, a 3D point can be represented as a 2D coordinate with a slice number. Thus, the main extension of the 2D framework to 3D is \emph{initial slice selection} (Figure \ref{fig:prompting_strategies}). 
Within this 3D context, a single prompt is defined as 2D coordinates localized within a single slice. Multiple 3D prompts are either several coordinates within one slice or individual coordinates distributed across multiple slices. In this work, as we utilized specific human-annotated slices, our strategy was limited to either a single selected slice or a combination of all selected slices ($N_S$ slices). Additionally, we investigated a prompting variant that incorporates the top and bottom slices of an object. By default, the same prompt primitives -- bounding box(\cblacksquare[0.2]{white}), center point(\cblackcircledot[0.25]{white}), or their combination(\cblacksquaredot[0.25]{white}) -- extracted from the largest component in a single slice were used, which represented the common configuration supported by all 3D models (Table \ref{tab:model_prompt_overview}).

\subsection{Dataset}\label{sec:dataset}

\begin{figure*}[h!]
    \centering
    \includegraphics[width=0.98\linewidth]{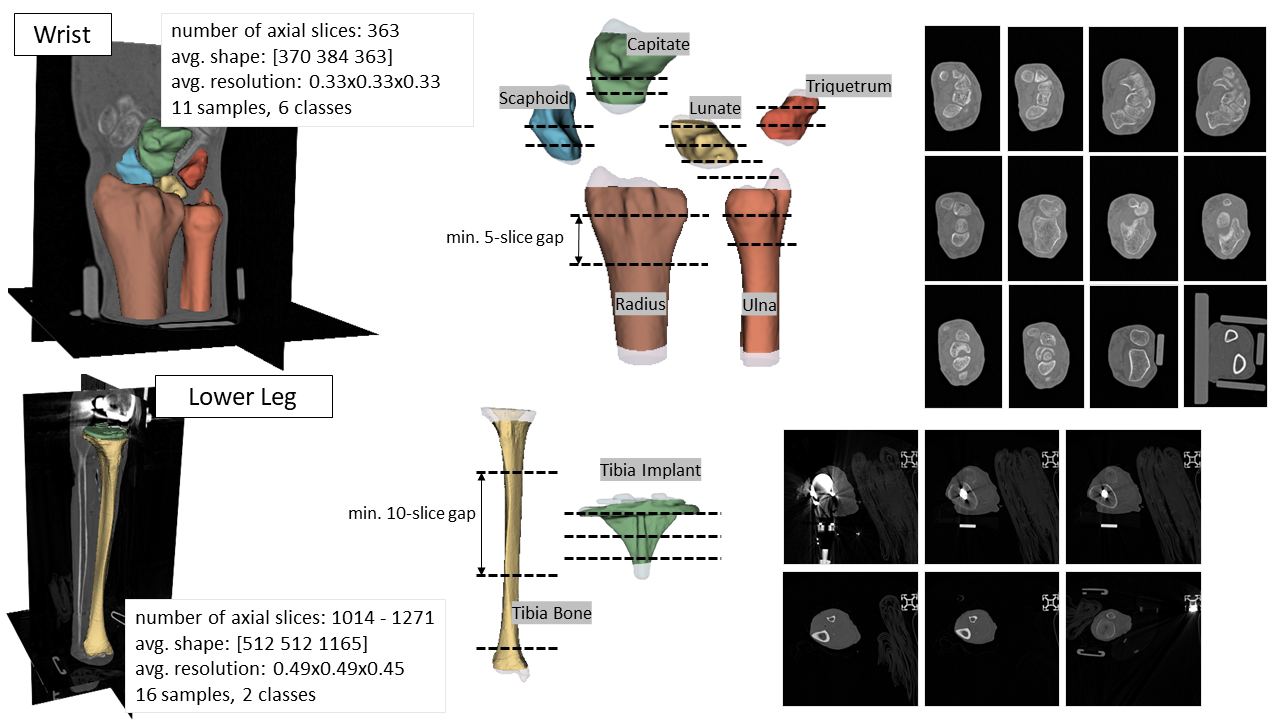}
    \hfill
    \includegraphics[width=0.98\linewidth]{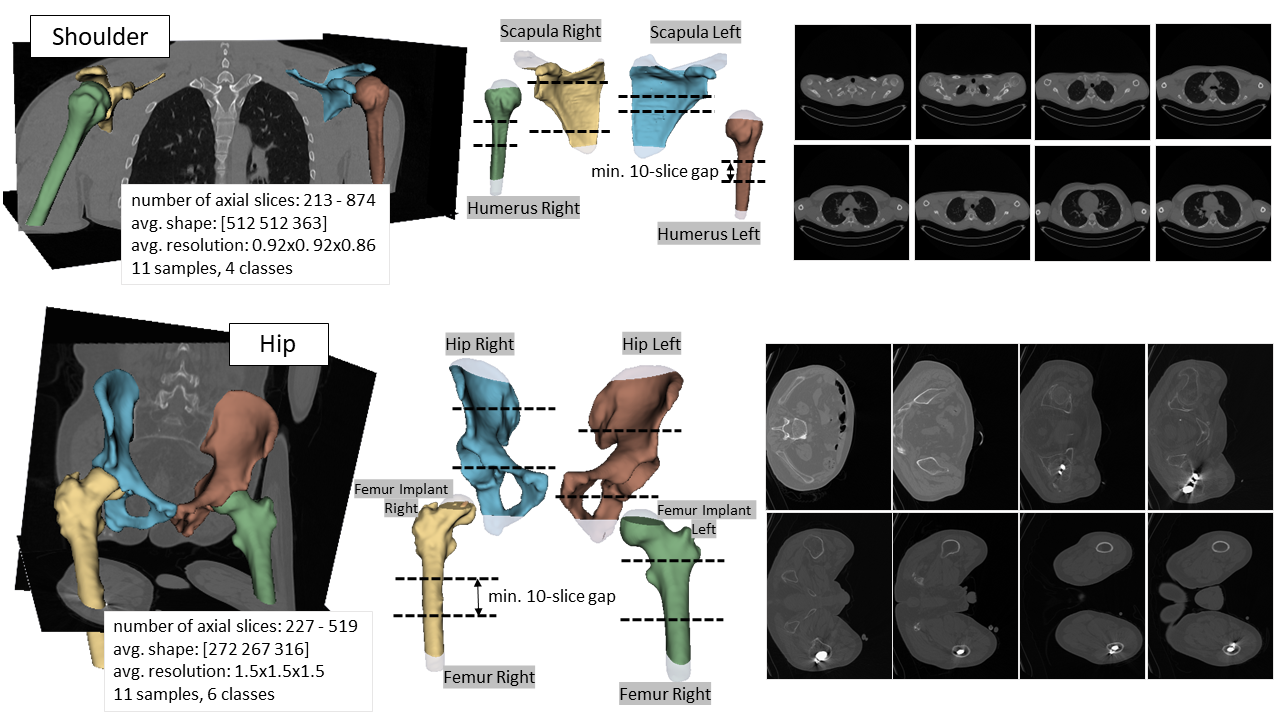}
    \caption{Dataset Overview: Our dataset consists of four subsets, i.e., Wrist, Lower Leg, Shoulder, and Hip\cite{wasserthal2023totalsegmentator}. A subset of 404 axial slices was extracted based on constraints ensuring data coverage, diversity and comparability.}
    \label{fig:dataset}
\end{figure*}

Since medical FMs are trained on publicly available datasets, including bone CT segmentation (\cite{wasserthal2023totalsegmentator, liu2021ctpelvic1k, verse2021}), an independent dataset is essential to fairly compare performance across models. 
A private dataset ensures a task-specific and independent evaluation, while public datasets enable to study reproducibility by the broader research community. To address both needs, we compiled a CT test dataset consisting of private CT scans from the department of Orthopaedic Surgery and Sports Medicine of the Amsterdam UMC, approved by the local Medical Ethics Committee (2025.0447), and selected CT samples of the TotalSegmentator test set \cite{wasserthal2023totalsegmentator}. Unfortunately, not all FMs included in our study specify their exact test dataset splits of the TotalSegmentator dataset (see Table \ref{tab:model_prompt_overview}). In total, our final dataset contains four skeletal regions, 49 CT scans and 18 class labels (Figure \ref{fig:dataset}). \newline

A subset of axial slices, the primary scanning direction, was selected from the full CT volumes to limit the annotation workload for participants in the observer study. The slice selection was performed once using random sampling for each class label (i.e., anatomical object), with constraints applied to ensure adequate data coverage, diversity, and comparability across data subsets (see \ref{sec:appendix_dataset}). The selection was kept consistent across all experiments and served as the initial slices for model prompting (i.e., with perfect and human prompting). In total, 404 axial CT slices have been selected, i.e., 132 for Wrist, 96 for Lower Leg, 88 for Shoulder and 88 for Hip.

\subsection{Observer Study}

An observer study was conducted on the platform \href{https://grand-challenge.org}{grand-challenge.org} with 20 medical students from Faculty of Medicine, University of Amsterdam. Participants provided informed consent prior to participation. 
Participants were instructed to place bounding boxes and center points on each bone structure visible in a given CT slice (with exception of vertebrae and rib bones to reduce annotation effort), following predefined annotation guidelines. These guidelines included precise definitions and multiple visual examples from different scans of the study dataset to ensure consistent interpretation. In addition to written guidelines, participants had access to a video showing the usage of the annotation platform. The annotation interface supported zooming and window/level adjustments, with default window settings tailored to each anatomical subset, and scrolling through the volume in all three planes (i.e., axial, sagittal, coronal), with the selected slice displayed as the default axial view.
To enable prompt-specific time tracking on grand-challenge.org, the placement of bounding boxes and center points was performed independently. Thus, each sample was annotated twice:  once per prompt type.
Before the main study, participants completed a training phase in which they annotated selected slices from one sample per data subset (i.e., per anatomical region; 18–34 slices in total) and received written feedback on their annotations. When deviations from the protocol were identified, participants were asked to correct their annotations and were provided with an additional round of feedback. This iterative process was repeated until the participant demonstrated a consistent understanding of the annotation protocol. After the training phase, all subsequent annotations were taken as provided, without additional correction or exclusion. Thus, no special handling of protocol deviations was applied.
The main study was conducted in the fixed annotation order: Wrist (180 slices), Lower Leg (180 slices), Shoulder (120 slices), and Hip (120 slices). Participants were randomly assigned to one of two groups to counterbalance ordering effects: one group always began with bounding box placement, the other with point placement.
To assess intra-rater variability, each project included duplicate slices together with the original samples. Participants were fully blinded to the duplication, meaning they were not informed that certain slices appeared twice nor in which order they occurred. The duplicated sample counts per subset were as follows: Wrist: 120 original + 60 duplicates, Lower Leg: 90 + 90 duplicates, Shoulder and Hip: 80 + 40 duplicates (for details see \ref{sec:appendix_dataset}).

\subsection{Evaluation design}

First, we quantified accuracy and consistency of the human prompts. Second, we compared the segmentation performance of the FMs prompted with perfect prompts, to make a model selection of the Pareto-optimal models. Then, we evaluated the segmentation accuracy and consistency of these FMs prompted with the human prompts and the performance difference between both prompt sources. Finally, the models' sensitivity to input prompt variations was determined. 

\subsubsection{Human prompt analysis}

To reduce annotation complexity, observer study prompts were categorized with broad categories (i.e., bone and implant) rather than the specific class labels required for prompting. Thus, a matching process assigned a class label to each observer study prompt by aligning the observer study prompts with their reference prompts (i.e., automatically extracted from the reference mask).
\textit{Human center points} were compared to reference points on a per-label, per-component basis. For each connected component in the reference mask, the Hungarian algorithm\footnote{\label{fn:scipy}See SciPy documentation: 
\url{https://docs.scipy.org/doc/scipy/reference/generated/scipy.optimize.linear_sum_assignment.html}}(linear sum assignment) with Euclidean distance as cost metric was used to ensure optimal one-to-one mapping. This approach minimizes total distances while allowing unmatched points, i.e., cases where the structures were either not labeled in the reference (e.g., the fibula in the Lower Leg or clavicle in the Shoulder dataset) or overlooked by the annotator. \textit{Human bounding boxes} were compared to the reference boxes on a per-label, per-object basis. Because a single bounding box may enclose multiple components, we matched boxes for each object (instead of component) using the Hungarian algorithm with Intersection over Union (IoU) as cost metric.\newline
Matched pairs were counted as true positives (TPs) and  unmatched reference prompts as false negatives (FNs). For completeness, unmatched human prompts were categories as false positives (FPs), and if both reference and human prompts were absent for a connected component or object, it was considered a true negative (TN). \newline

\textit{Detection performance} was measured by Recall ($TP/(TP+FN)$). 
For all TPs, \textit{localization error} was quantified for center points, i.e., human center points and center points derived from human bounding boxes, by calculating the Euclidean distance and the signed/absolute $x$ and $y$ coordinate offset ($\Delta x$, $\Delta y$, $|\Delta x|$, $|\Delta y|$) to the corresponding reference points. For bounding boxes, the Intersection over Union (IoU) and the signed/absolute difference in width and height ($\Delta w$, $\Delta h$, $|\Delta w|$, $|\Delta h|$) was computed with respect to the corresponding reference boxes.\newline
\textit{Annotation consistency} was evaluated at intra-rater and inter-rater levels using the same metrics as described above. Intra-rater consistency was assessed by comparing repeated annotations from the same annotator, while inter-rater consistency was assessed by pairwise comparison of annotations from two different annotators for the same object.\newline
Distances were calculated in both pixel coordinates and physical units (mm) based on the spacing of the reference masks (Figure \ref{fig:dataset}). The metrics for human prompt analysis were summarized as medians with interquartile ranges (IQR) to avoid scaling on outliers.\newline

For all annotators, \textit{annotation time} was recorded per sample. Due to platform functionality, annotation times at the level of individual annotations were not available. Therefore, the annotation time per annotation was estimated by averaging the total time spent per image over the number of annotations within each sample.

\subsubsection{Segmentation analysis} \label{sec:method_evaluation_segmentation}

\textit{Segmentation performance} was assessed by comparing masks generated from either reference or human prompts against the reference masks, which were obtained by manual segmentation, following \cite{magg2025zeroshot}. For human prompts, \textit{segmentation consistency} was determined using an intra-rater approach, where masks of the same sample generated from an annotator's first prompt set were compared to those from their second set. Finally, the \textit{performance gap between reference- and human-prompted results} was quantified by pairwise difference analysis per sample.\newline
For 2D models, the evaluation was performed on the predictions of the selected slices in a 2D manner (slice-wise). For 3D models, the generated volumetric predictions based on the selected slices as initial input were evaluated in a 3D manner (volumetric).\newline
Following previous work \cite{magg2025zeroshot} and MetricsReloaded  \cite{MaierHein2024MetricsReloaded}, the Dice Similarity Coefficient (DSC), the Normalized Surface Dice (NSD) (threshold is set to largest spatial resolution of $1.5$mm), and the $95\%$-percentile Hausdorff distance (HD95) were used as metrics for segmentation performance analysis, with the implementation of the DisTorch framework \cite{distorch}.
In line with common practice, summarized values of DSC, NSD and HD95 are reported as mean and standard deviation (std). 

\subsubsection{Pareto front}

A model $i$ with performance vector 
$ 
\mathbf{m}_i = (m_{i1}, m_{i2}, \dots, m_{id})
$
is \emph{Pareto-optimal} (non-dominated) if no other model $j$ dominates it. Model $j$ dominates model $i$ (denoted $\mathbf{m}_j \succ \mathbf{m}_i$) if:
\[
m_{jk} \;\succeq_k\; m_{ik} \quad \forall k \in {1, \dots, d} \quad \text{and} \quad \exists k' : m_{jk'} \;\succ_{k'}\; m_{ik'}.
\]
Here, $\succeq_k$ denotes the comparison direction for metric $k$ (i.e., superiority: $\geq$ if higher is better, $\leq$ if lower is better). In other words, no other model performs at least as good across all performance metrics and strictly better in at least one of them.
Then, the \emph{Pareto front} is defined as the set of all Pareto-optimal models:
\begin{equation*}
    \mathcal{P} = \left\{\, i \;\middle|\; \nexists j: \mathbf{m}_j \succ \mathbf{m}_i \right\}.
\end{equation*}
In our work, a model lies on the Pareto front if no other model achieved higher DSC, higher NSD, and lower HD95 simultaneously, with at least one of these comparisons being strictly inequal.

\subsubsection{Model selection}

Within each category -- defined by prediction dimensionality (2D vs. 3D), training data domain (medical vs. natural), and prompting strategy (bounding box, center point, or combination) -- the Pareto-optimal models with the smallest number of parameters were identified. The selection prioritized smaller models that demonstrate strong performance within their category, ensuring computational efficiency. These models were chosen for further analysis with human prompting.

\subsubsection{Model sensitivity to input prompt variations}

Following the analysis of prompt variability (intra-rater and inter-rater) and segmentation consistency, the relationship between these two factors was analyzed to assess model sensitivity to input prompt variability. 
Spearman's rank correlation coefficient ($\rho$) was calculated between the prompt variability (Euclidean distance or IoU) and the corresponding segmentation consistency (DSC, NSD, HD95). A low correlation coefficient indicates low sensitivity (i.e., increased robustness) to prompt variability, as it suggests the output masks remain similar regardless of variations in the input prompt.\newline
This analysis was first performed for the intra-rater prompt variability and segmentation consistency. In case models showed a lack of significant correlation for this setting, the analysis was repeated for an inter-rater setting on the same sample set to determine the transition between non statistically significant and statistically significant correlation. To optimize the computational overhead of exhaustive pairwise comparisons ($n=190$ per sample per model), an iterative search was implemented. First, the annotator pool was sorted by mean euclidean distance (i.e., prompt variation from one rater to all others), with the lowest-variability rater serving as lower bound and the highest-variability rater as upper bound. If statistically non-significant correlation was shown at the lower bound, the upper bound was tested. If statistically significant model sensitivity was detected at the upper bound, the pool median was tested. Then, the search proceeded by splitting the remaining intervals in halves: testing the lower partition to identify the statistical significance threshold, and the upper partition to verify statistical non significance. This recursive process pinpointed the threshold with only a fraction of the exhaustive calculations.

\subsubsection{Statistical significance tests}
The Wilcoxon signed-rank test was used for all pairwise comparisons, including the evaluation of performance differences between reference- and human-prompted segmentations, as well as the comparison of intra- versus inter-rater consistency. To account for multiple comparisons ($n$) within each test group, a Bonferroni correction was applied to the initial significance level ($\alpha = 0.05$). In the reported results, asterisks ($*$) denote statistical significance ($p < \alpha/n$), while a lack of $*$ indicates non-significant results.

\section{Results}\label{sec:results}

First, the human prompt accuracy and consistency were analyzed. Then, the segmentation performance was evaluated using reference prompts, including a model selection of the Pareto-optimal models. These models were further tested with the human prompts to determine segmentation performance and consistency, followed by an analysis of the performance differences of the two prompt sources. Finally, models' sensitivity to prompt variability was examined with intra- and inter-rater prompt variability and segmentation consistency.

\subsection{Human prompt variation}

\paragraph{Center points} The median Euclidean distance between the human and the reference center points was $1.50$mm (IQR: $0.7-3.0$mm) (Table \ref{tab:human_annotations_accuracy}). The median intra-rater Euclidean distance calculated from samples, that the annotators processed twice, was $0.98$mm (IQR: $0.5-1.9$mm) and the median inter-rater Euclidean distance was $1.37$mm (IQR: $0.7-2.6$mm) (Table \ref{tab:human_annotations_consistency}). 
\paragraph{Bounding Boxes} The median IoU between the human and the reference bounding boxes was $90.56$\% (IQR: $83.4-94.5\%$) (Table \ref{tab:human_annotations_accuracy}). The median intra-rater IoU on samples seen twice by the annotators was $93.04$\% (IQR: $88.5-96.1\%$, the median inter-rater IoU was $90.11$\% (IQR: $84.2-94.0\%$) (Table \ref{tab:human_annotations_consistency}).

\begin{table}[h!]
\centering
\caption{Annotation performance for human bounding box and center point variations, reported as median (IQR).}
\label{tab:human_annotations_accuracy}
\setlength{\tabcolsep}{3pt}
\renewcommand{\arraystretch}{1.2}
\centering
    \begin{tabular}{l|c|c}
    \hline
    Metrics & Bounding Box & Center Point \\
    \hline \hline
    \multicolumn{3}{c}{Annotation accuracy in \% $\uparrow$  } \\
    \hline
    IoU & 90.56{\footnotesize(83.4-94.5)} & - \\
    \hline
    \multicolumn{3}{c}{Localization error in mm $\downarrow$} \\
    \hline
    Eucl. distance &  0.49{\footnotesize(0.0-1.4)} & 1.50{\footnotesize(0.7-3.0)} \\
    $|\Delta x|$ & 0.00{\footnotesize(0.0-0.9)} & 0.98{\footnotesize(0.3-1.9)}  \\
    $|\Delta y|$ & 0.00{\footnotesize(0.0-0.9)} & 0.97{\footnotesize(0.3-1.9)}  \\
    $|\Delta w|$ & 1.30{\footnotesize(0.3-2.9)} & - \\
    $|\Delta h|$ & 1.46{\footnotesize(0.3-2.9)} & - \\
    $\Delta x$ & 0.00{\footnotesize(0.0-0.0)} & 0.33{\footnotesize(-0.3-1.5)} \\
    $\Delta y$ & 0.00{\footnotesize(0.0-0.3)} & 0.33{\footnotesize(0.0-1.5)} \\
    $\Delta w$ & 0.83{\footnotesize(0.0-2.0)} & - \\
    $\Delta h$ & 0.98{\footnotesize(0.0-2.5)} & - \\
    \hline
    \multicolumn{3}{c}{Detection performance in \% $\uparrow$  } \\
    \hline
    Recall & 96.1 & 95.5 \\
    \hline
    \end{tabular}
\end{table}

\paragraph{Intra- vs. Inter-rater annotation consistency} For center point and bounding box, comparing the intra- and inter-rater consistency revealed a statistically significant difference for all four datasets ($p$-values  $< 0.5/4=0.0125$), with intra-rater annotations demonstrated higher consistency compared to inter-rater annotations (Table \ref{tab:human_annotations_consistency}).

\begin{table}[h]
\centering
\caption{Intra-rater and inter-rater annotation consistency for human bounding box and center point, reported as median (IQR).}
\label{tab:human_annotations_consistency}
\setlength{\tabcolsep}{3pt}
\renewcommand{\arraystretch}{1.2}
\centering
\resizebox{\linewidth}{!}{
    \begin{tabular}{l|cc|cc}
    \hline
    Metrics & \multicolumn{2}{c|}{Bounding Box} & \multicolumn{2}{c}{Point} \\
    \multicolumn{1}{c|}{} & intra & inter & intra & inter \\
    \hline \hline

    \multicolumn{5}{c}{Agreement in \% $\uparrow$ } \\
    \hline
    IoU & 93.04{\footnotesize(88.5-96.1)} & 90.11{\footnotesize(84.2-94.0)} & - & - \\
    \hline
    \multicolumn{5}{c}{Variability in mm $\downarrow$} \\
    \hline
    Eucl. distance & 0.49{\footnotesize(0.0-1.4)} & 0.73{\footnotesize(0.3-1.5)} & 0.98{\footnotesize(0.5-1.9)} & 1.37{\footnotesize(0.7-2.6)} \\
    $|\Delta x|$ & 0.33{\footnotesize(0.0-0.9)} & 0.33{\footnotesize(0.0-1.0)} & 0.49{\footnotesize(0.0-1.5)} & 0.83{\footnotesize(0.0-1.6)}\\
    $|\Delta y|$ & 0.33{\footnotesize(0.0-0.8)} & 0.33{\footnotesize(0.0-1.0)} & 0.49{\footnotesize(0.0-1.5)} & 0.83{\footnotesize(0.0-1.6)} \\
    $|\Delta h|$ & 0.65{\footnotesize(0.0-1.5)} & 0.98{\footnotesize(0.3-2.6)} & - & - \\
    $|\Delta w|$ & 0.65{\footnotesize(0.0-1.5)} & 0.98{\footnotesize(0.3-2.6)} & - & - \\
    \hline
    \multicolumn{5}{c}{Systematic differences in mm $\downarrow$} \\
    \hline
    $\Delta x$ & 0.00{\footnotesize(-0.3-0.3)} & 0.00{\footnotesize(-0.3-0.3)} & 0.00{\footnotesize(-0.5-0.5)} & 0.00{\footnotesize(-0.3-0.3)} \\
    $\Delta y$ & 0.00{\footnotesize(-0.3-0.0)} & 0.00{\footnotesize(-0.3-0.3)} &  0.00{\footnotesize(-0.5-0.5)} & 0.00{\footnotesize(-0.8-0.8)} \\
    $\Delta w$ & 0.00{\footnotesize(-0.7-0.7)} & 0.00{\footnotesize(-1.0-1.0)} & - & - \\
    $\Delta h$ & 0.00{\footnotesize(-0.7-0.7)} & 0.00{\footnotesize(-1.0-1.0)} & - & - \\
    \hline
    \end{tabular}
}
\end{table}

\begin{figure*}[t]
    \centering
    \begin{minipage}{0.92\linewidth}
        \begin{subfigure}{0.99\linewidth}
        \centering
            \includegraphics[width=0.24\columnwidth]{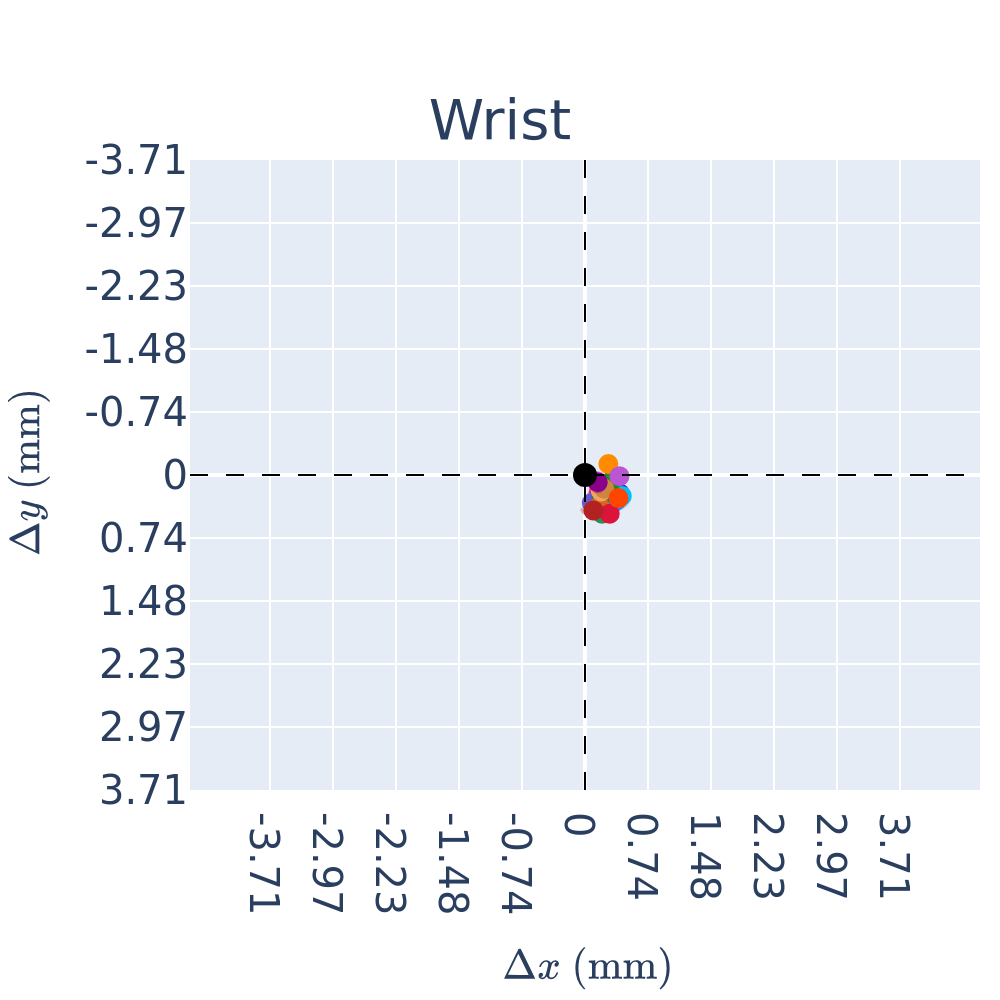}
            \includegraphics[width=0.24\columnwidth]{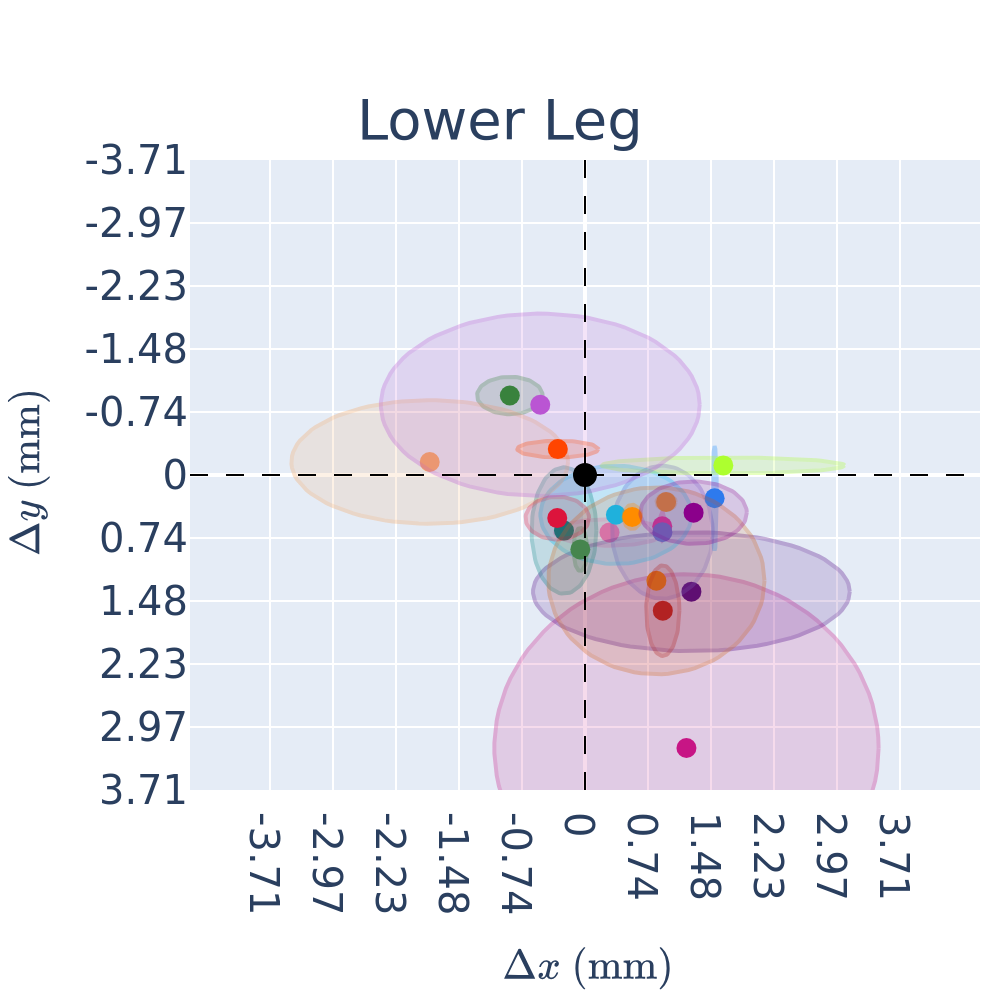}
            \includegraphics[width=0.24\columnwidth]{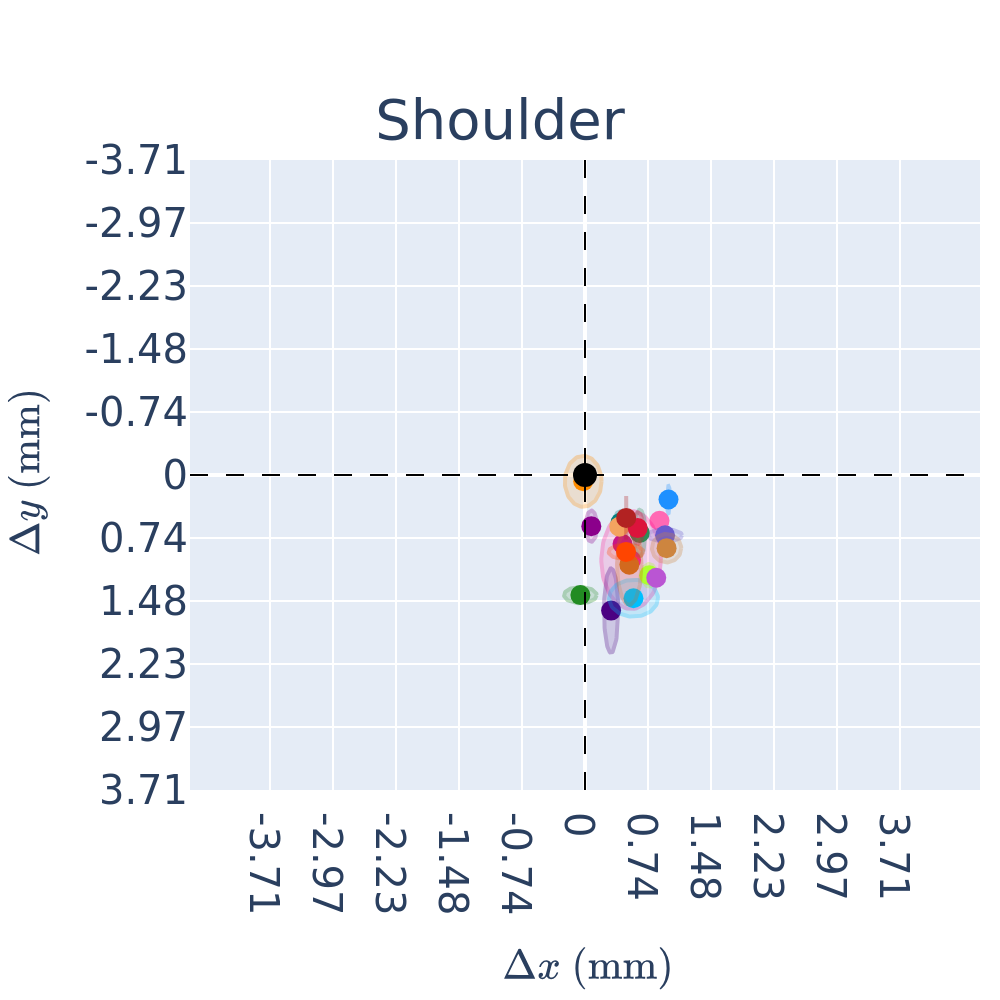}
            \includegraphics[width=0.24\columnwidth]{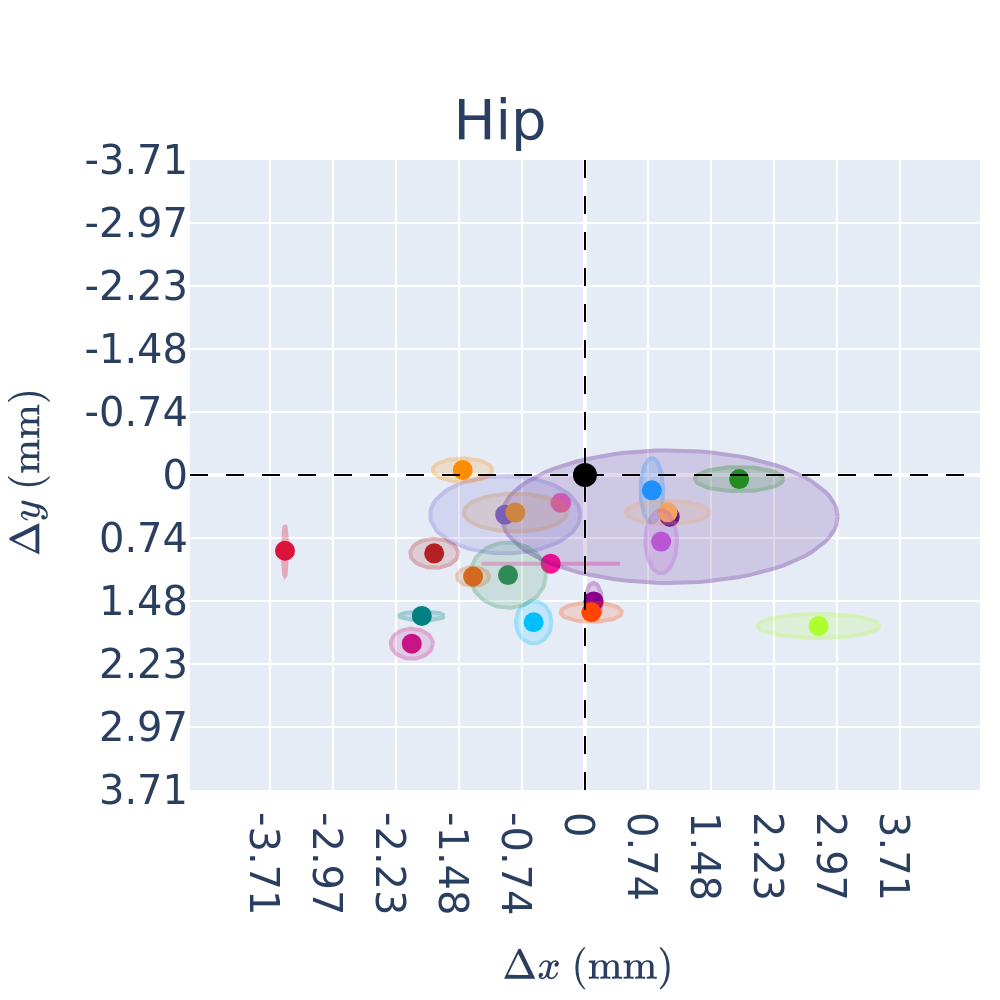}
            \caption{Center point}
            \label{fig:target_plot_point}
        \end{subfigure}
        \begin{subfigure}{0.99\linewidth}
        \centering
            \includegraphics[width=0.24\columnwidth]{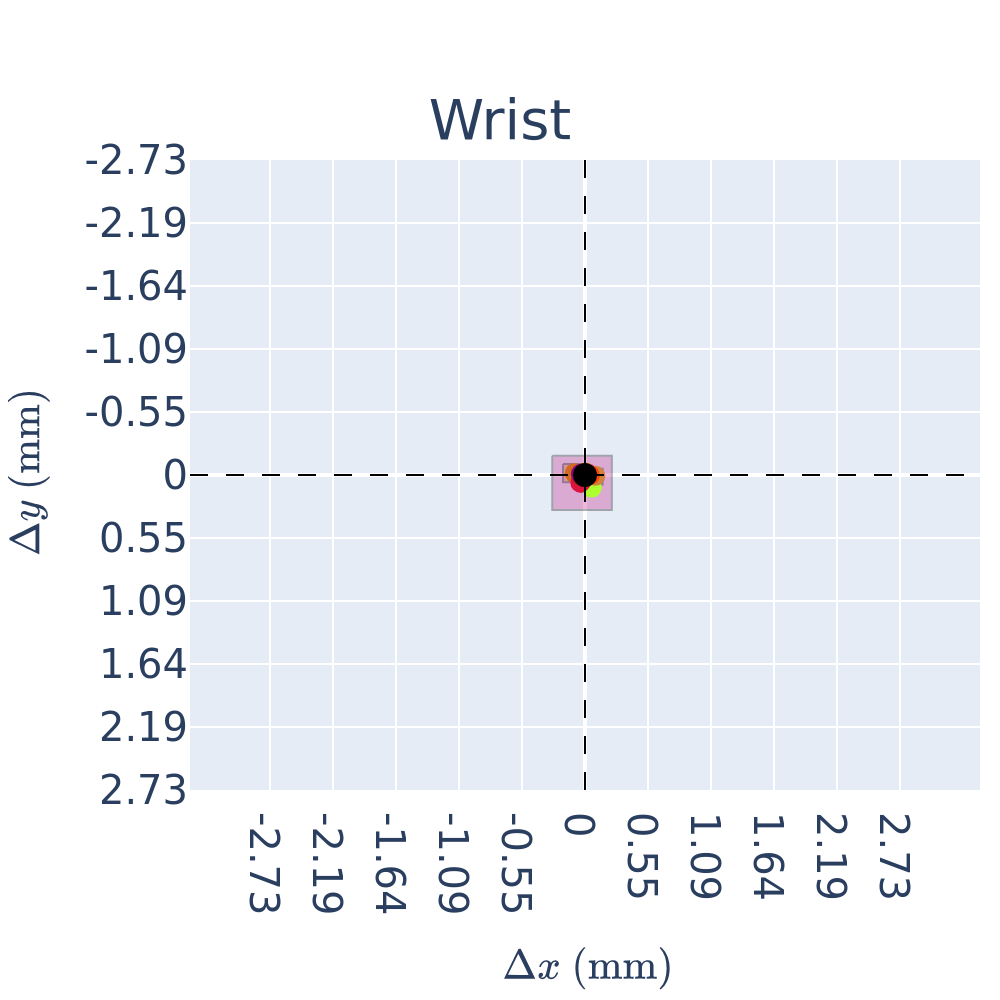}
            \includegraphics[width=0.24\columnwidth]{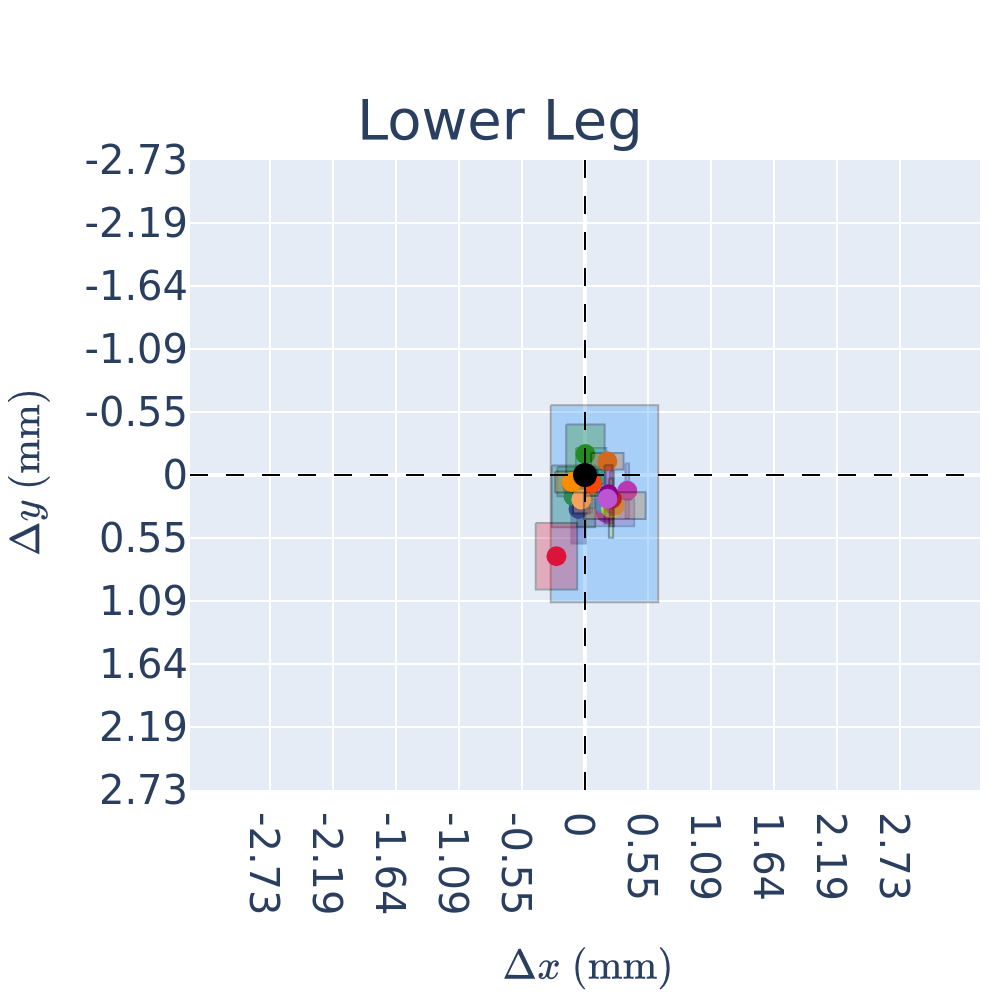}
            \includegraphics[width=0.24\columnwidth]{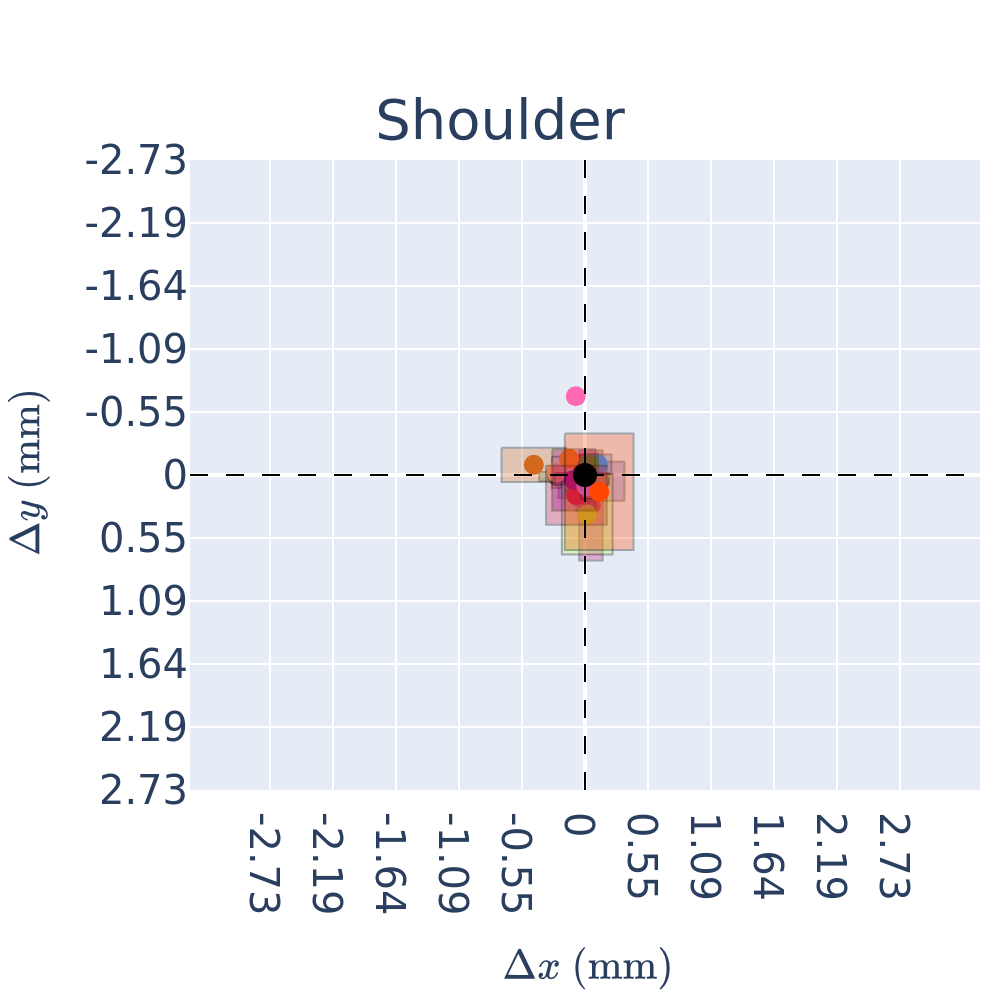}
            \includegraphics[width=0.24\columnwidth]{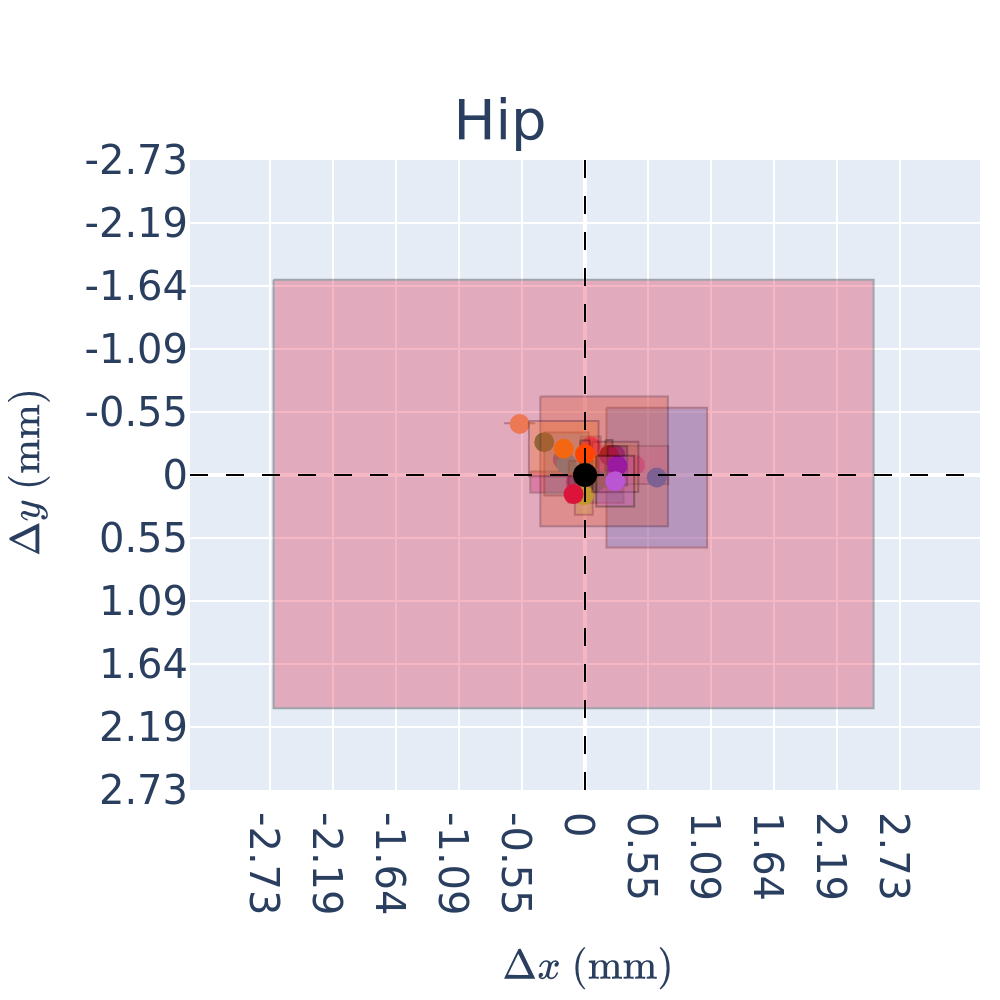}
            \caption{Bounding box}
            \label{fig:target_plot_box}
        \end{subfigure}
    \end{minipage}\hfill
    \begin{minipage}{0.08\linewidth}
        \centering
        \includegraphics[width=\linewidth, trim={905 100 0 150}, clip]{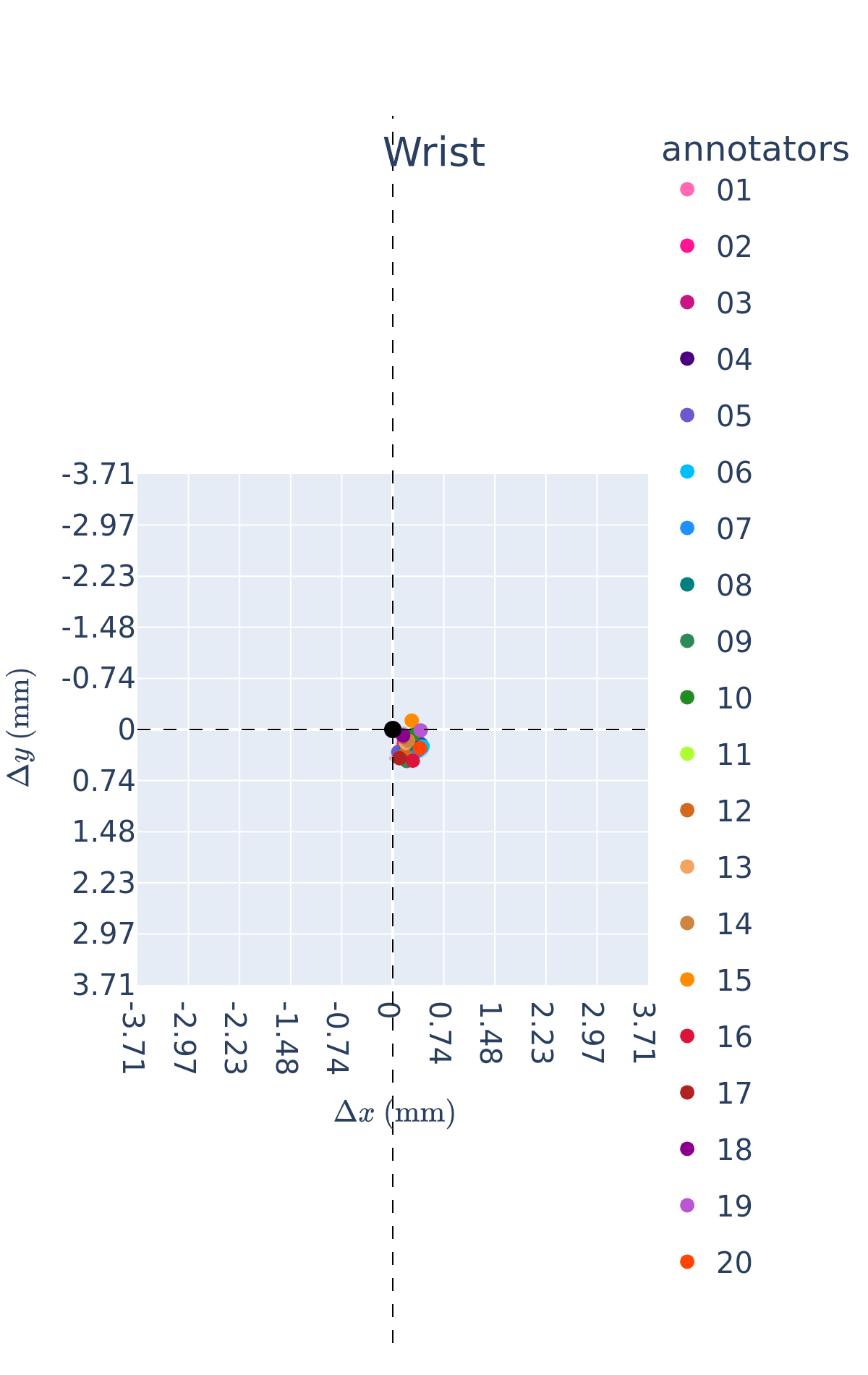}
    \end{minipage}
    \caption{Spatial distribution of prompt placement per annotator per dataset.\newline
    \scriptsize{Each point corresponds to one annotator. It represents the mean deviations (mm) in the x- ($\Delta x$) and y-directions ($\Delta y$) of the center point (i.e., human (a) or extracted from the bounding box (b)), relative to the reference prompt at the origin (0,0). The same-colored (more transparent) ellipse (a) and rectangles (b) represent each annotator's intra-rater consistency ((a): ($\Delta x$, $\Delta y$), (b): ($\Delta w$, $\Delta h$)). Wrist shows the least localization errors and highest consistency, while Lower Leg and Hip show high localization errors and low intra- and inter-rater consistency.}}
    \label{fig:target_plot}
\end{figure*}

\paragraph{Dataset-specific performance} For both human prompts, there were considerable differences across data subsets and class labels in terms of localization errors and intra-rater consistency (Figure \ref{fig:target_plot}). The annotations for the dataset Lower Leg and Hip showed high localization errors and low intra-rater consistency, mostly due to the class tibia bone for center points (Figure \ref{fig:ablation_human_points_scatter}) and tibia implant and hip for bounding box (Figure \ref{fig:ablation_human_box_scatter}), while annotations in the dataset Wrist showed overall the lowest localization errors and high intra-rater consistency.
Detailed results and visualizations per data subset and class labels are available in the \ref{app:human_generated_prompts_eval}.

\begin{figure}[h!]
    \centering
    \includegraphics[width=1\columnwidth]{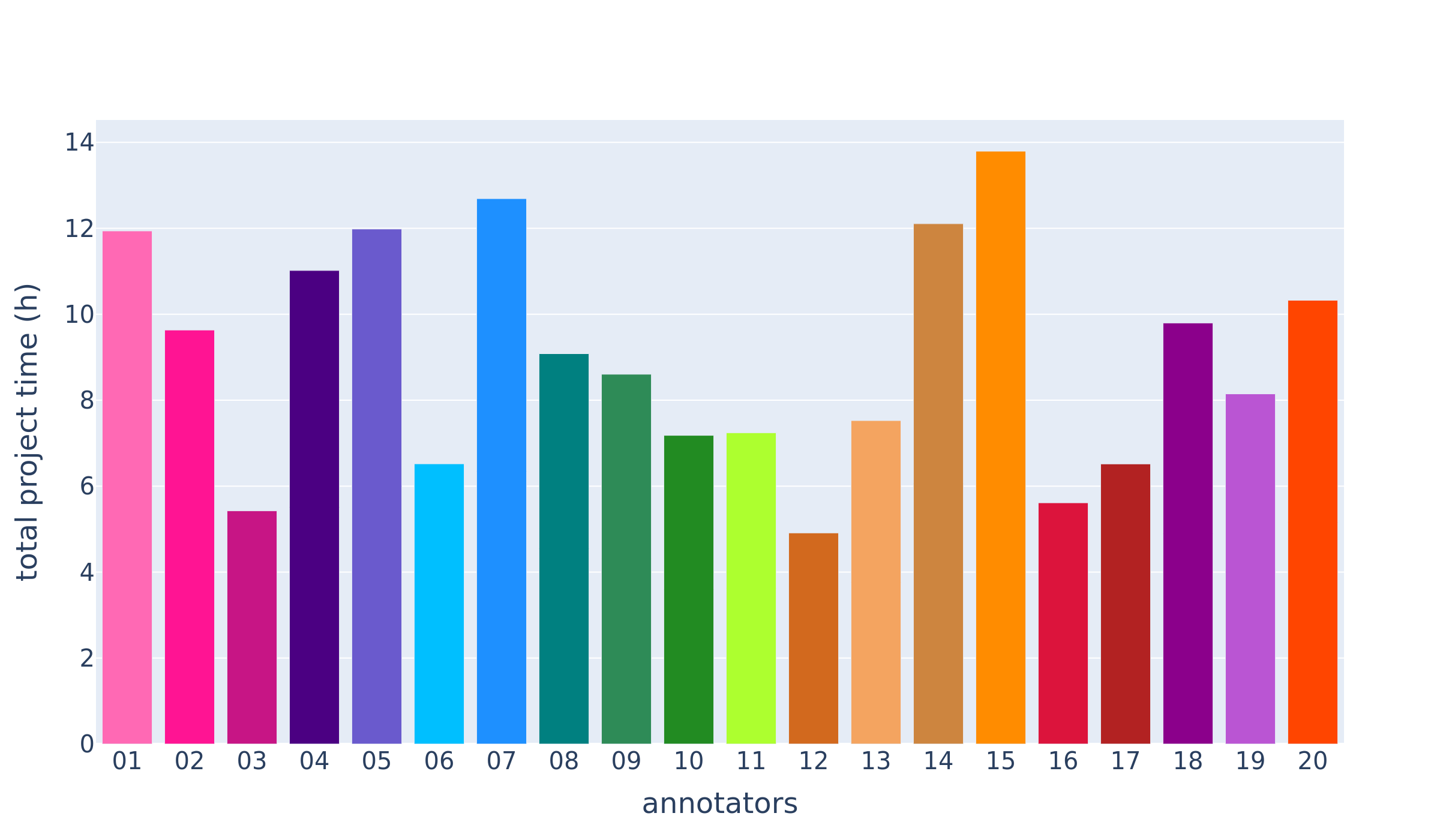}
    \caption{Accumulated annotation time per annotator for all projects.}
    \label{fig:time_analysis}
\end{figure}

\paragraph{Annotation Time}
The average annotation time was $4.22$ sec for a center point and $11.37$ sec for a bounding box. 
The annotators required between $5$ and $14$ hours to complete the project (excluding training phase), with a median of $8$ hours and $48$ min (IQR: $7$ hours to $11$ hours and $18$ minutes) (Figure \ref{fig:time_analysis}).

\subsection{Segmentation performance with reference prompts} \label{sec:results_pareto_front}

\begin{figure*}[t]
    \centering
    \begin{subfigure}{1\columnwidth}
        \centering
        \includegraphics[width=\linewidth]{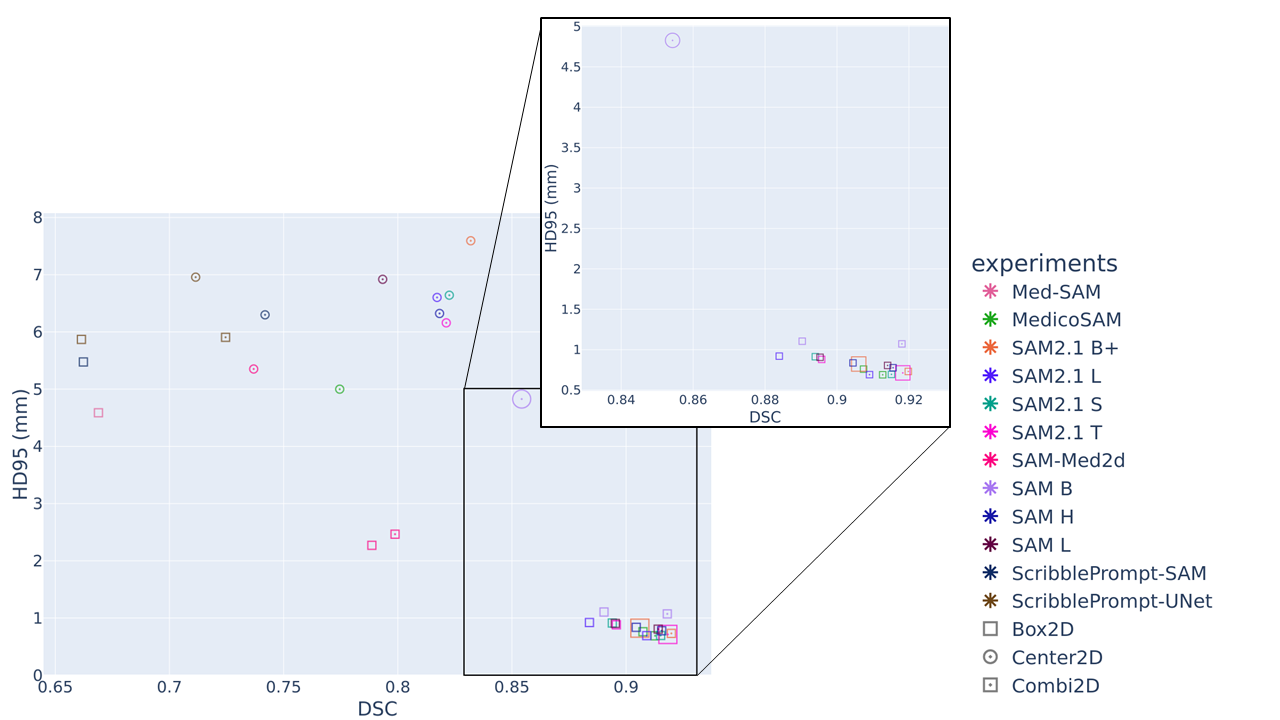}
        \caption{All $12$ 2d models evaluated slice-wise.}
        \label{subfig:dsc_hd95_automatic_prompts_2d}
    \end{subfigure}
    \hfill
    \begin{subfigure}{1\columnwidth}
        \centering
        \includegraphics[width=\linewidth]{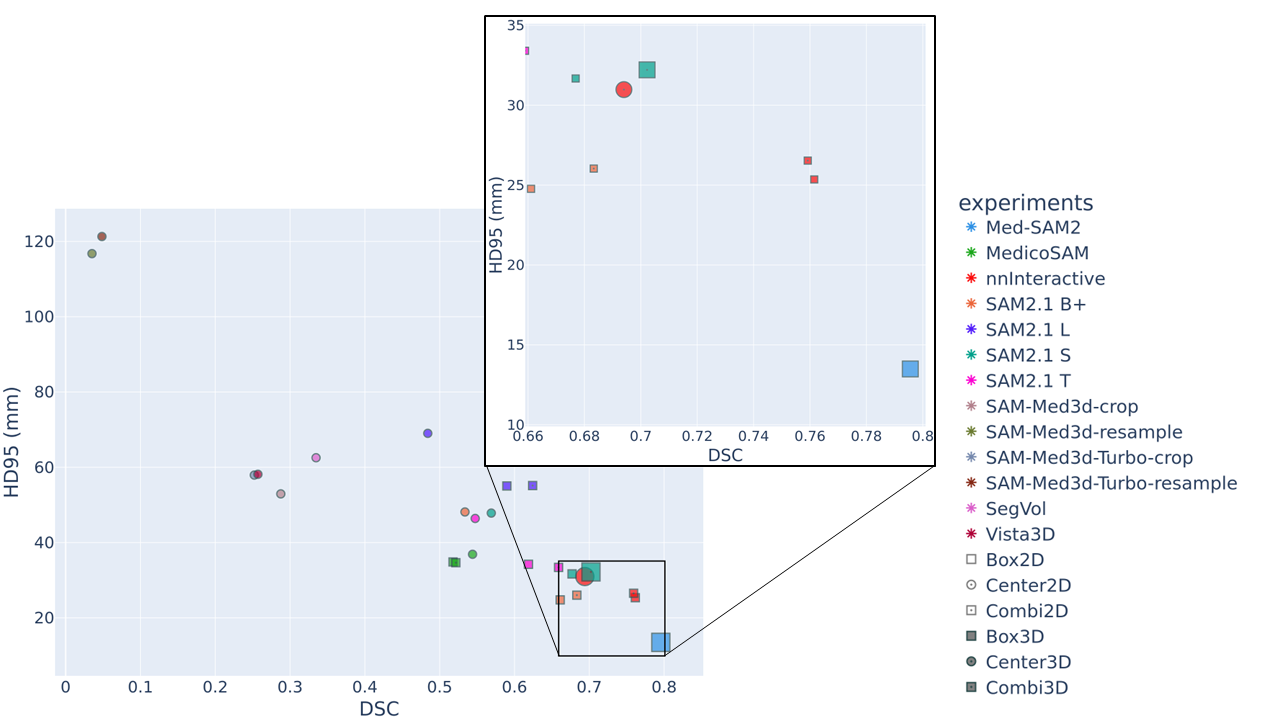}
        \caption{All $13$ 3D models evaluated volumetric.}
        \label{subfig:dsc_hd95_automatic_prompts_3d}
    \end{subfigure}
    \caption{DSC vs. HD95 (mm) performance of all models (color-encoded) and three prompt types (symbol-encoded) for perfect prompts. Larger symbols highlight the smallest Pareto-optimal models.}
    \label{fig:dsc_hd95_automatic_prompts}
\end{figure*}

\paragraph{2D models} For reference prompts, the combined prompting strategy worked the best, followed by the bounding box and then center point (Table \ref{tab:pareto_front}, Figure \ref{subfig:dsc_hd95_automatic_prompts_2d}, Table \ref{tab:appendix_results_all_models}). The overall best model was SAM2.1 T with combination prompting ($91.83\%$ DSC, $98.38\%$ NSD, $0.71$mm HD95).
\paragraph{3D models} The bounding box and combined prompting strategies achieved higher performance than center-point prompts (Table \ref{tab:pareto_front}, Figure \ref{subfig:dsc_hd95_automatic_prompts_3d}, Table \ref{tab:appendix_results_all_models}). The overall best model was Med-SAM2 with bounding box prompting ($79.56$\% DSC, $80.25$\% NSD, $13.49$mm HD95).

\begin{table*}[t]
\centering
\caption{Segmentation performance with reference prompts of Pareto-optimal models per prompt type (i.e., bounding box, center point, combination) and category (2D vs. 3D; medical vs. natural).\newline 
{\protect \scriptsize The Pareto-optimal models with the least parameters per category are highlighted in bold (selected for further analysis with human prompts). Gray-shaded cells and prompt symbols next to the model names indicate the smallest Pareto-optimal models per prompt type. No Pareto-optimal results are omitted in this Table and can be found in Table \ref{tab:appendix_results_all_models}.}}
\label{tab:pareto_front}
\setlength{\tabcolsep}{3pt}
\renewcommand{\arraystretch}{1.3}
\resizebox{\linewidth}{!}{
\begin{tabular}{llc:ccc:ccc:ccc}
    \hline
    & \textbf{Model} & & \multicolumn{3}{:c:}{\textbf{Bounding Box} 2D \cblacksquare[0.2]{black} or 3D \cblacksquare[0.2]{white}} & \multicolumn{3}{c}{\textbf{Center Point} \cblackcircledot[0.25]{black} (2D) or \cblackcircledot[0.25]{white} (3D)} & \multicolumn{3}{:c}{\textbf{Combination} \cblacksquaredot[0.25]{black} (2D) or \cwhitesquaredot[0.25]{black} (3D)} \\ 
    & & Size & DSC $\uparrow$ & NSD $\uparrow$ & HD95 $\downarrow$ & DSC $\uparrow$ & NSD $\uparrow$ & HD95 $\downarrow$ & DSC $\uparrow$ & NSD $\uparrow$ & HD95 $\downarrow$ \\
    &  & M & \% & \% & mm & \% & \% & mm & \% & \% & mm \\    
    \hline \hline \rule{0pt}{2.6ex}
    
    & \multicolumn{11}{c}{\textbf{2D Models}} \\
    \hline
    \multirow{3}{*}{\rotatebox[origin=c]{25}{\footnotesize{medical}}} & \textbf{MedicoSAM2D} & 94 & \textbf{90.74}{\footnotesize±7.7} & \textbf{97.36}{\footnotesize±3.6} & \textbf{0.76}{\footnotesize±0.9} & \textbf{77.46}{\footnotesize±19.3} & \textbf{83.23}{\footnotesize±18.4} & \textbf{5.00}{\footnotesize±5.9} & \textbf{91.27}{\footnotesize±7.4} & \textbf{97.74}{\footnotesize±3.3} & \textbf{0.69}{\footnotesize±0.8} \\
    & SAM-Med2d & 271 & - & - & - & 73.69{\footnotesize±17.0} & 84.48{\footnotesize±14.9} & 5.35{\footnotesize±5.0} & - & - & - \\
    & \textbf{ScribblePrompt-SAM} & 94 & - & - & - & \textbf{74.19}{\footnotesize±14.6} & \textbf{84.22}{\footnotesize±12.6} & \textbf{6.30}{\footnotesize±5.3} & - & - & - \\
    \hdashline
    \multirow{5}{*}{\rotatebox[origin=c]{25}{\footnotesize{natural}}} & \textbf{SAM B} \cblackcircledot[0.25]{black} & 94 & - & - & - & \fcolorbox{gray!50}{gray!50}{\textbf{85.43}{\footnotesize±14.4}} & \fcolorbox{gray!50}{gray!50}{\textbf{90.82}{\footnotesize±13.0}} & \fcolorbox{gray!50}{gray!50}{\textbf{4.83}{\footnotesize±6.3}} & - & - & - \\
    & \textbf{SAM2.1 B+} \cblacksquare[0.25]{black} & 81 & \fcolorbox{gray!50}{gray!50}{\textbf{90.60}{\footnotesize±8.1}} & \fcolorbox{gray!50}{gray!50}{\textbf{97.84}{\footnotesize±3.5}} & \fcolorbox{gray!50}{gray!50}{\textbf{0.82}{\footnotesize±1.0}} & - & - & - & 91.98{\footnotesize±7.2} & 98.21{\footnotesize±3.6} & 0.73{\footnotesize±1.1} \\
    & SAM2.1 L & 224 & - & - & - & - & - & - & 90.90{\footnotesize±6.9} & 98.36{\footnotesize±3.2} & 0.69{\footnotesize±1.0} \\
    & SAM2.1 S & 46 & - & - & - & - & - & - & 91.51{\footnotesize±7.0} & 98.40{\footnotesize±3.3} & 0.69{\footnotesize±0.9} \\
    & \textbf{SAM2.1 T} \cblacksquaredot[0.25]{black} & 39 & - & - & - & - & - & - & \fcolorbox{gray!50}{gray!50}{\textbf{91.83}{\footnotesize±6.9}} & \fcolorbox{gray!50}{gray!50}{\textbf{98.38}{\footnotesize±3.2}} & \fcolorbox{gray!50}{gray!50}{\textbf{0.71}{\footnotesize±1.0}} \\
    
    \hline
    & \multicolumn{11}{c}{\textbf{3D Models}} \\
    \hline
    \multirow{2}{*}{\rotatebox[origin=c]{25}{\footnotesize{medical}}} & \textbf{Med-SAM2} \cwhitesquare[0.25]{black} & 39 & \fcolorbox{gray!50}{gray!50}{\textbf{79.56}{\footnotesize±11.1}} & \fcolorbox{gray!50}{gray!50}{\textbf{80.25}{\footnotesize±10.5}} & \fcolorbox{gray!50}{gray!50}{\textbf{13.49}{\footnotesize±11.1}} & - & - & - & - & - & - \\
    & \textbf{nnInteractive} \cwhitecircledot[0.25]{black} \cwhitesquaredot[0.25]{black} & 102 & - & - & - & \fcolorbox{gray!50}{gray!50}{\textbf{69.40}{\footnotesize±11.2}} & \fcolorbox{gray!50}{gray!50}{\textbf{68.23}{\footnotesize±12.0}} & \fcolorbox{gray!50}{gray!50}{\textbf{30.98}{\footnotesize±9.4}} & \fcolorbox{gray!50}{gray!50}{\textbf{75.92}{\footnotesize±9.4}} & \fcolorbox{gray!50}{gray!50}{\textbf{76.60}{\footnotesize±9.6}} & \fcolorbox{gray!50}{gray!50}{\textbf{26.53}{\footnotesize±10.3}} \\
    \hdashline
    \multirow{3}{*}{\rotatebox[origin=c]{25}{\footnotesize{natural}}} & SAM2.1 B+ & 81 & 66.11{\footnotesize±10.1} & 66.59{\footnotesize±10.0} & 24.77{\footnotesize±18.1} & - & - & - & 68.33{\footnotesize±9.4} & 67.86{\footnotesize±10.2} & 26.04{\footnotesize±18.2} \\
    & \textbf{SAM2.1 S} & 46 & \textbf{67.69}{\footnotesize±10.2} & \textbf{68.48}{\footnotesize±10.0} & \textbf{31.67}{\footnotesize±21.6} & 56.90{\footnotesize±19.1} & 53.96{\footnotesize±20.2} & 47.84{\footnotesize±31.2} & \textbf{70.22}{\footnotesize±10.1} & \textbf{69.88}{\footnotesize±10.7} & \textbf{32.21}{\footnotesize±22.0} \\
    & \textbf{SAM2.1 T} & 39 & - & - & - & \textbf{54.74}{\footnotesize±15.9} & \textbf{52.92}{\footnotesize±16.9} & \textbf{46.40}{\footnotesize±28.5} & - & - & - \\
    \hline
\end{tabular}
}
\end{table*}

\paragraph{Selected Models} Considering prediction dimensionality (2D vs. 3D), training data domain (medical vs. natural), and prompting strategy (bounding box, center point, combination), the smallest Pareto-optimal models for prompting with reference prompts were collected in Table \ref{tab:pareto_front}.
Focusing only on dimensionality, ignoring the training data domain, the Pareto-optimal models with the least parameter per prompt type were: SAM2.1 B+ (2D bounding box), SAM B (2D center point), SAM2.1 T (2D center point), Med-SAM2 (3D bounding box), nnInteractive (3D center point, 3D combination). These models were marked with gray cells in Table \ref{tab:pareto_front} and large symbols in Figure \ref{fig:dsc_hd95_automatic_prompts}.
\paragraph{Ablation Studies} Comparing SAM2 with SAM2.1 and investigating variations of the 3D prompting strategies for automated extracted prompts showed the following trends:
\begin{itemize}[noitemsep, topsep=0pt]
    \item There was no statistically significant difference between SAM2 and SAM2.1, except for the tiny (T) models (\ref{sec:sam2_vs_sam21}).
    \item Limiting the propagation in SAM2.1 and Med-SAM2 prevented over-segmentation at the top and bottom of an object which improved performance (\ref{sec:limited_vs_unlimited}).
    \item Medical FMs benefit from multiple initial slices more than SAM2.1 models (\ref{sec:single_vs_multiple_initial_slices}). With multiple initial slices, nnInteractive exceeded the performance of Med-SAM2, which was the Pareto-optimal model for the default settings (i.e, with single initial slice).
    \item There was only a marginal difference (mostly statistically non-significant) between using a single or multiple prompts for 3D prompting (\ref{sec:single_vs_multiple_prompts}).
\end{itemize}

\subsection{Segmentation performance with human prompts}

\paragraph{2D models} SAM and SAM2.1 maintained their superior performance compared to medical FMs, mirroring the trends seen with reference prompts (Table \ref{tab:segmentation_human_prompts}). The overall best model was again SAM2.1 T with combination prompting ($89.65\%$ DSC, $97.71\%$ NSD, $1.06$mm HD95). 
\paragraph{3D models} All 3D medical FMs consistently outperformed SAM2.1 for all prompt types (Table \ref{tab:segmentation_human_prompts}). The overall best model was Med-SAM2 with bounding box prompting
($77.05\%$ DSC, $79.47\%$ NSD, $14.36$mm HD95).
\paragraph{Segmentation consistency} Intra-rater consistency is high for all FMs (Table \ref{tab:segmentation_human_prompts}). Notably, consistency was most pronounced in the top-performing models. 

\begin{table*}[ht]
\centering
\caption{Segmentation performance and intra-rater segmentation consistency with human prompts -- grouped by prompt type (bounding box, center point, combination).
\scriptsize{The best values per prompt type are highlighted in bold. The best performing models also showed the highest consistency. The performance difference to perfect prompts is shown in Table \ref{tab:automatic_vs_human_performance}.}}
\label{tab:segmentation_human_prompts}
\setlength{\tabcolsep}{3pt}
\renewcommand{\arraystretch}{1.3}
    \resizebox{0.99\linewidth}{!}{
    \begin{tabular}{llc:ccc:ccc:ccc}
        \hline
        & \textbf{Model} & & \multicolumn{3}{:c:}{\textbf{Bounding Box} 2D \cblacksquare[0.2]{black} or 3D \cblacksquare[0.2]{white}} & \multicolumn{3}{c}{\textbf{Center Point} \cblackcircledot[0.25]{black} (2D) or \cblackcircledot[0.25]{white} (3D)} & \multicolumn{3}{:c}{\textbf{Combination} \cblacksquaredot[0.25]{black} (2D) or \cwhitesquaredot[0.25]{black} (3D)} \\ 
        & & Size & DSC $\uparrow$ & NSD $\uparrow$ & HD95 $\downarrow$ & DSC $\uparrow$ & NSD $\uparrow$ & HD95 $\downarrow$ & DSC $\uparrow$ & NSD $\uparrow$ & HD95 $\downarrow$ \\
        &  & M & \% & \% & mm & \% & \% & mm & \% & \% & mm \\    
        \hline \hline \rule{0pt}{2.6ex}

        & \multicolumn{11}{c}{\textbf{Segmentation performance}} \\
        
        & \multicolumn{11}{c}{\textbf{2D Models}} \\
        \hline
        \multirow{2}{*}{\rotatebox[origin=c]{25}{\footnotesize{medical}}}& MedicoSAM2D & 94 & 86.12{\footnotesize±13.6} & 95.40{\footnotesize±5.8} & 1.26{\footnotesize±1.7} & 75.95{\footnotesize±20.7} & 83.54{\footnotesize±16.6} & 5.13{\footnotesize±5.8} & 86.53{\footnotesize±13.0} & 95.09{\footnotesize±7.0} & 1.38{\footnotesize±2.1} \\
        & ScribblePrompt-SAM & 94 & - & - & - & 72.50{\footnotesize±18.0} & 84.26{\footnotesize±12.9} & 6.39{\footnotesize±5.3} & - & - & - \\
        \hdashline
        \multirow{3}{*}{\rotatebox[origin=c]{25}{\footnotesize{natural}}} & SAM B & 94 & - & - & - & \textbf{83.69}{\footnotesize±17.5} & \textbf{90.99}{\footnotesize±11.6} & \textbf{4.85}{\footnotesize±6.2} & - & - & - \\
        & SAM2.1 B+ & 81 & \textbf{87.86}{\footnotesize±12.8} & \textbf{96.80}{\footnotesize±5.0} & \textbf{1.15}{\footnotesize±1.6} & - & - & - & - & - & - \\
        & SAM2.1 T & 39 & - & - & - & - & - & - & \textbf{89.65}{\footnotesize±10.8} & \textbf{97.41}{\footnotesize±4.7} & \textbf{1.06}{\footnotesize±1.7} \\
        \hline
        & \multicolumn{11}{c}{\textbf{3D Models}} \\
        \hline
        \multirow{2}{*}{\rotatebox[origin=c]{25}{\footnotesize{medical}}} & Med-SAM2 & 39 & \textbf{76.80}{\footnotesize±13.5} & \textbf{79.27}{\footnotesize±11.2} & \textbf{14.46}{\footnotesize±11.8} & - & - & - & - & - & - \\
        & nnInteractive & 102 & - & - & - & \textbf{68.12}{\footnotesize±12.6} & \textbf{68.63}{\footnotesize±11.5} & \textbf{30.10}{\footnotesize±8.8} & \textbf{75.59}{\footnotesize±10.6} & \textbf{77.29}{\footnotesize±9.1} & \textbf{25.65}{\footnotesize±9.5} \\
        \hdashline
        \multirow{2}{*}{\rotatebox[origin=c]{25}{\footnotesize{natural}}} & SAM2.1 S &46 & 65.93{\footnotesize±11.6} & 67.83{\footnotesize±10.2} & 32.71{\footnotesize±21.6} & - & - & - & 68.80{\footnotesize±11.2} & 69.19{\footnotesize±10.9} & 33.88{\footnotesize±22.4} \\
        & SAM2.1 T & 39 & - & - & - & 53.72{\footnotesize±16.3} & 52.93{\footnotesize±16.5} & 46.84{\footnotesize±27.8} & - & - & - \\ 
        
         \hline \hline \rule{0pt}{2.6ex}
         & \multicolumn{11}{c}{\textbf{Segmentation consistency}} \\
        
        & \multicolumn{11}{c}{\textbf{2D Models}} \\
        \hline
        \multirow{2}{*}{\rotatebox[origin=c]{25}{\footnotesize{medical}}} & MedicoSAM2D & 94 & 95.35{\footnotesize±8.1} & 99.20{\footnotesize±1.8} & 0.38{\footnotesize±0.8} & 96.27{\footnotesize±10.8} & 98.29{\footnotesize±6.3} & 0.99{\footnotesize±3.9} & 95.15{\footnotesize±9.7} & 98.48{\footnotesize±3.9} & 0.64{\footnotesize±1.7} \\
        & ScribblePrompt-SAM & 94 & - & - & - & 93.13{\footnotesize±13.1} & 96.45{\footnotesize±9.3} & 1.68{\footnotesize±5.0} & - & - & - \\
        \hdashline
        \multirow{3}{*}{\rotatebox[origin=c]{25}{\footnotesize{natural}}} & SAM B & 94 & - & - & - & \textbf{97.17}{\footnotesize±9.4} & \textbf{99.07}{\footnotesize±3.5} & \textbf{0.49}{\footnotesize±1.9} & - & - & - \\
        & SAM2.1 B+ & 81 & \textbf{97.13}{\footnotesize±8.1} & \textbf{99.40}{\footnotesize±1.8} & \textbf{0.26 }{\scriptsize±0.8} & - & - & - & - & - & - \\
        & SAM2.1 T & 39 & - & - & - & - & - & - & \textbf{97.71}{\footnotesize±9.2} & \textbf{99.38}{\footnotesize±2.1} & \textbf{0.31}{\footnotesize±1.2} \\
        \hline
        & \multicolumn{11}{c}{\textbf{3D Models}} \\
        \hline
        \multirow{2}{*}{\rotatebox[origin=c]{25}{\footnotesize{medical}}} & Med-SAM2 & 39 & \textbf{88.13}{\footnotesize±20.0} & \textbf{90.79}{\footnotesize±16.2} & \textbf{7.58}{\footnotesize±15.1} & - & - & - & - & - & - \\
        & nnInteractive & 102 & - & - & - & \textbf{84.89}{\footnotesize±20.6} & \textbf{86.71}{\footnotesize±18.0} & \textbf{7.32}{\footnotesize±12.7} & \textbf{88.44}{\footnotesize±17.7} & \textbf{88.75}{\footnotesize±15.2} & \textbf{8.05}{\footnotesize±12.9} \\
        \hdashline
        \multirow{2}{*}{\rotatebox[origin=c]{25}{\footnotesize{natural}}} & SAM2.1 S & 46 & 85.45{\footnotesize±20.9} & 87.63{\footnotesize±17.8} & 16.94{\footnotesize±31.3} & - & - & - & 87.46{\footnotesize±19.6} & 88.28{\footnotesize±17.6} & 16.64{\footnotesize±30.5} \\
        & SAM2.1 T & 39 & - & - & - & 79.63{\footnotesize±28.5} & 80.52{\footnotesize±27.2} & 23.94{\footnotesize±40.8} & - & - & - \\
        \hline
    \end{tabular}
    }
\end{table*}

\subsection{Performance gap between reference- and human-prompted results}
2D models exhibited a statistically significant decline in performance when transitioning to human prompts ($2.07\%$ DSC, $0.87\%$ NSD, $-0.25$mm HD95; $p$-values $<0.05/6=0.0083$), while 3D models showed a smaller but still statistically significant performance drop compared to their reference-prompted counterparts ($1.06\%$ DSC, $0.47\%$ NSD, $-0.39$mm HD95; with $p$-values $<0.05/6=0.0083$) (Table \ref{tab:automatic_vs_human_performance}).

\subsection{Model sensitivity to input prompt variations}
\begin{figure*}[ht]
    \centering
    \begin{minipage}{0.83\linewidth}
        \begin{subfigure}{0.49\linewidth}
        \centering
            \includegraphics[width=\columnwidth]{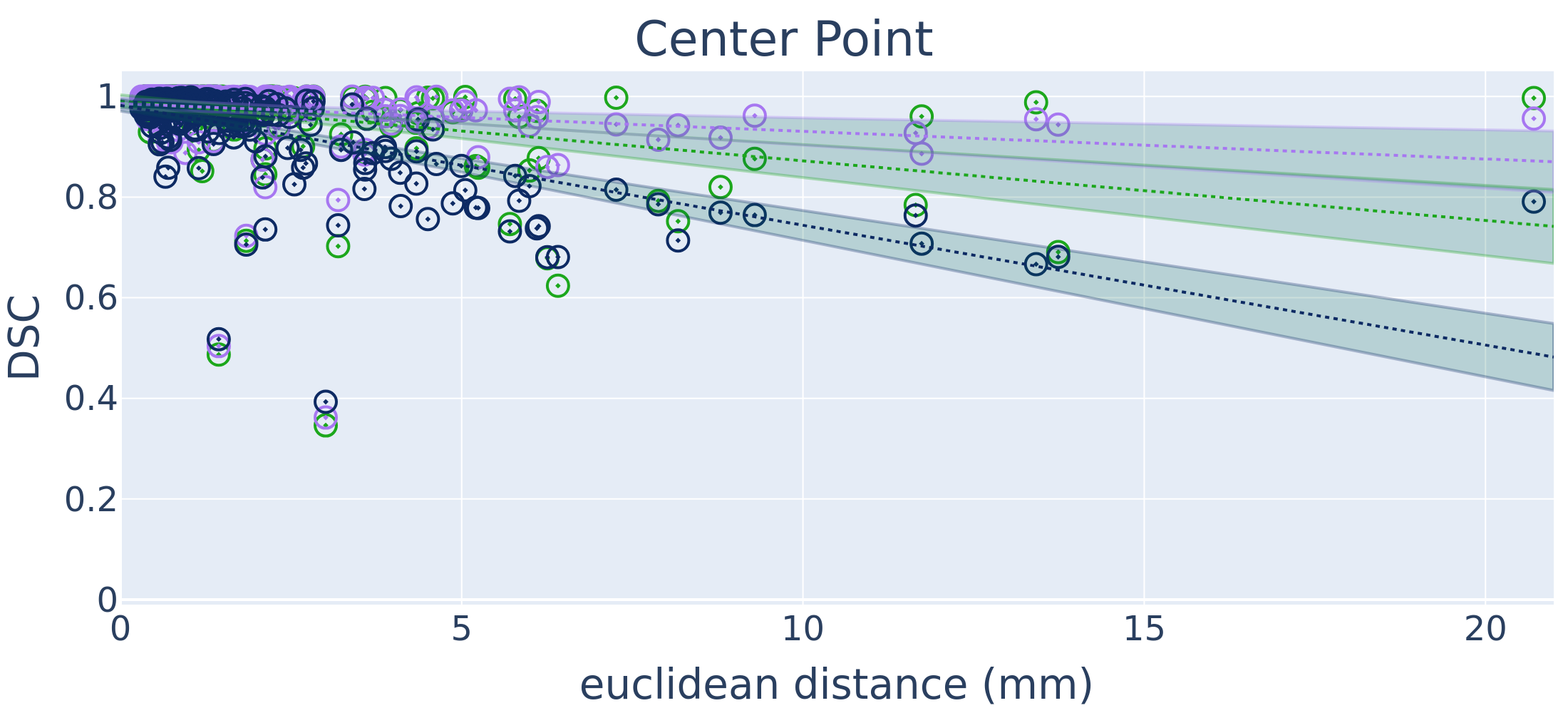}
            \includegraphics[width=\columnwidth]{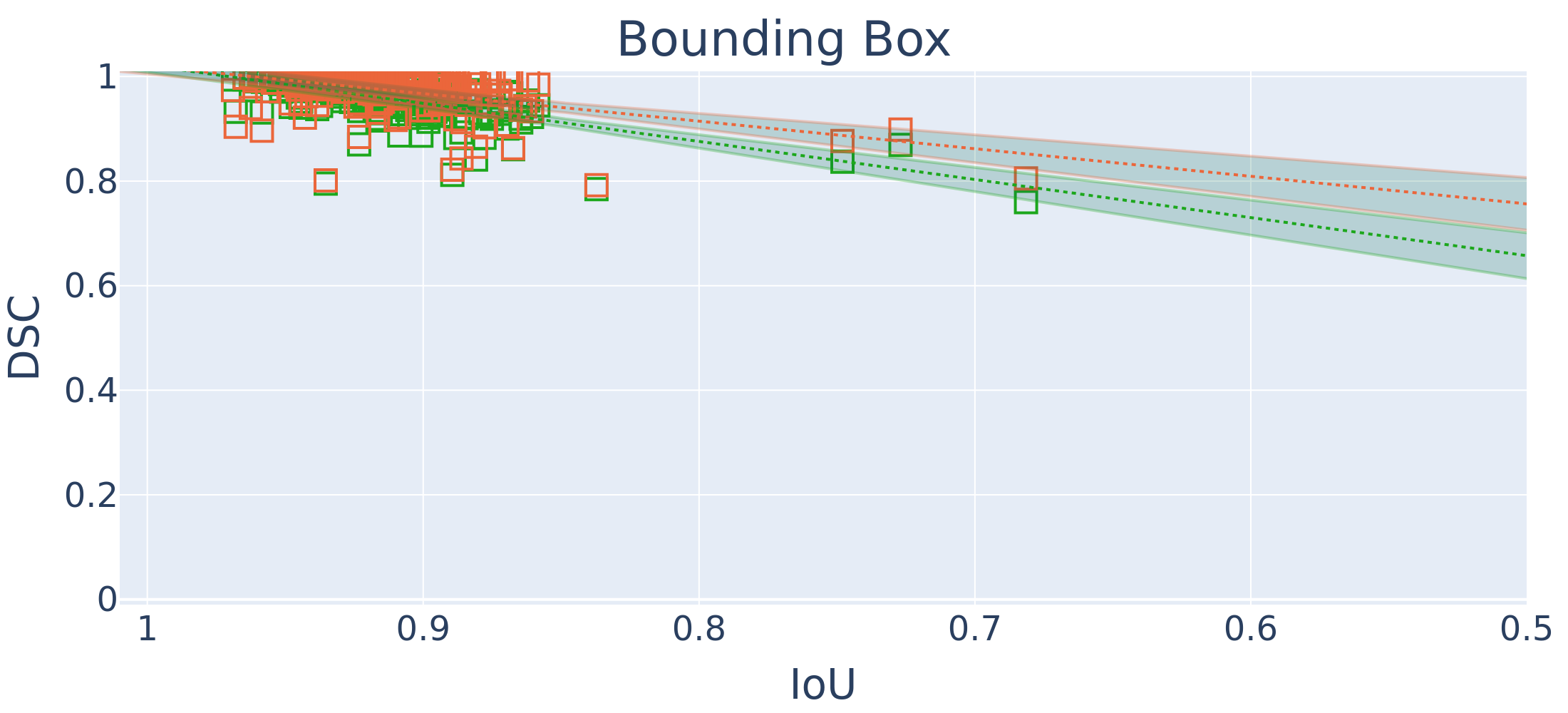}
            \includegraphics[width=\columnwidth]{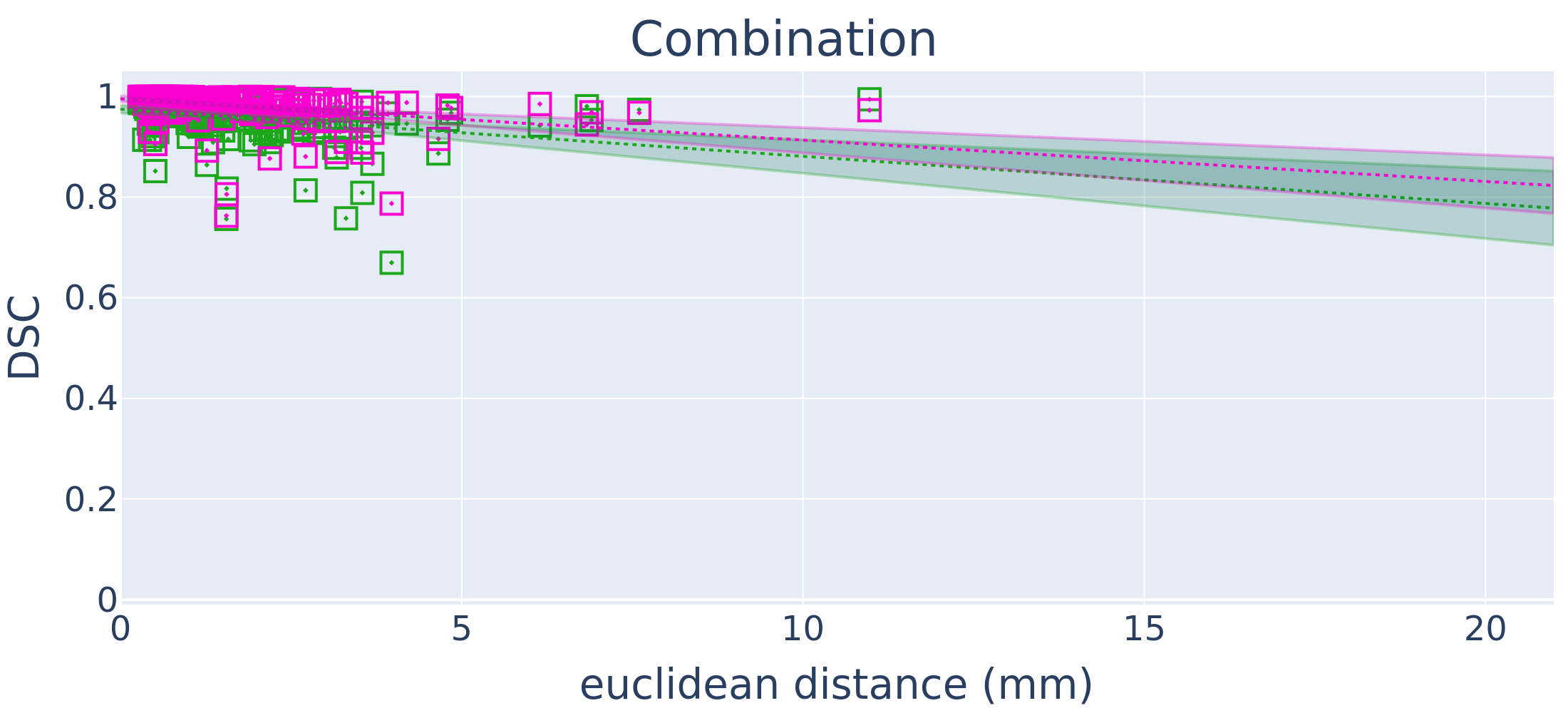}
            \caption{2D Models}
            \label{fig:robustness_2d_models}
        \end{subfigure}
        \hfill
        \begin{subfigure}{0.49\linewidth}
        \centering
            \includegraphics[width=\columnwidth]{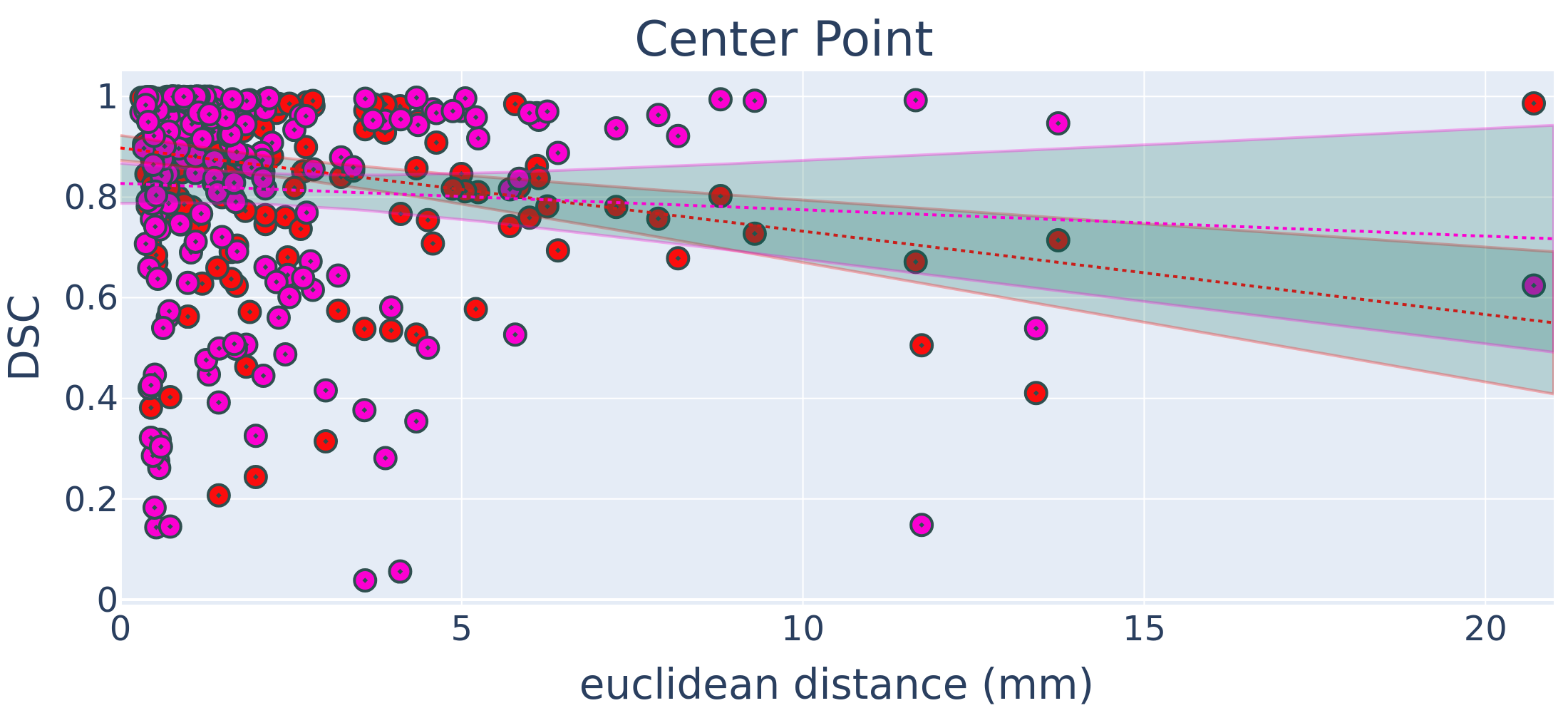}
            \includegraphics[width=\columnwidth]{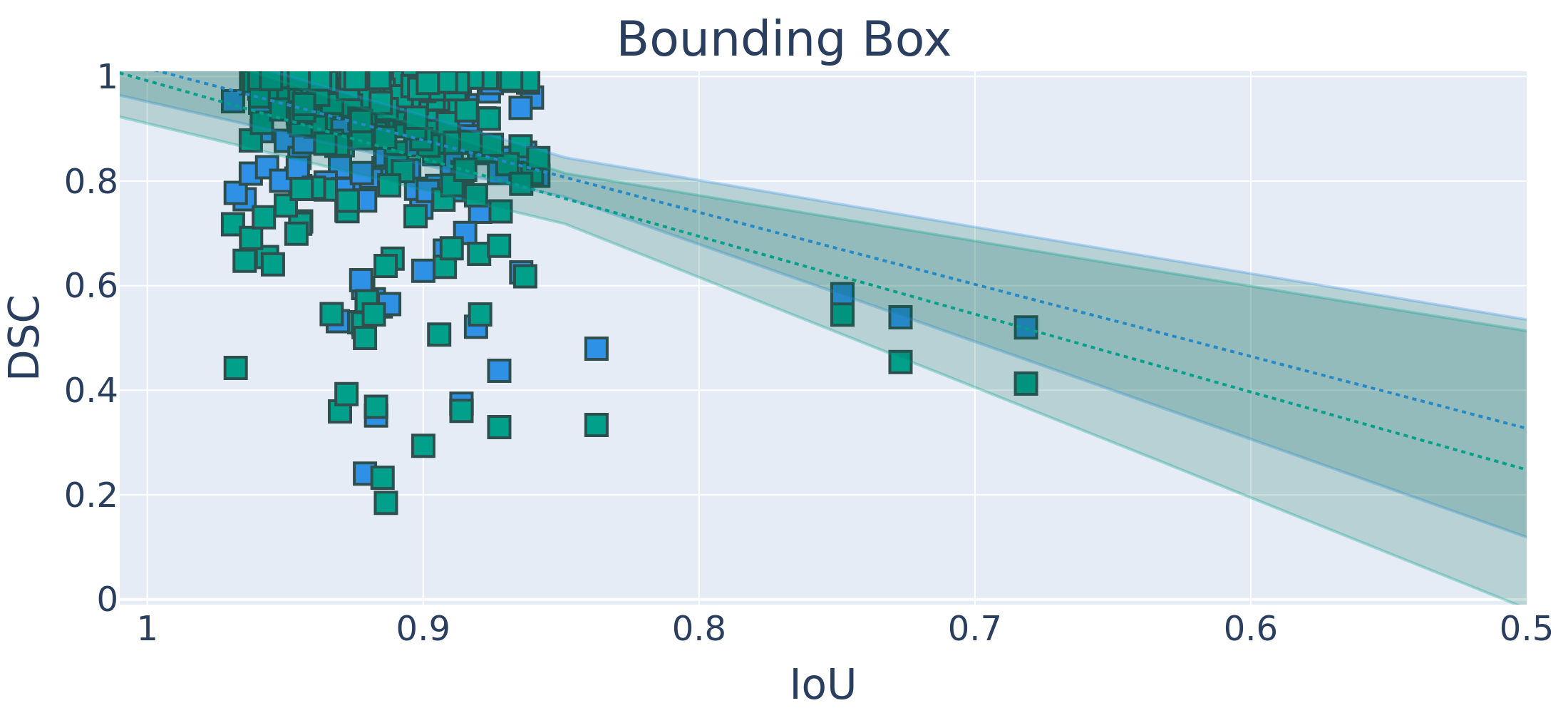}
            \includegraphics[width=\columnwidth]{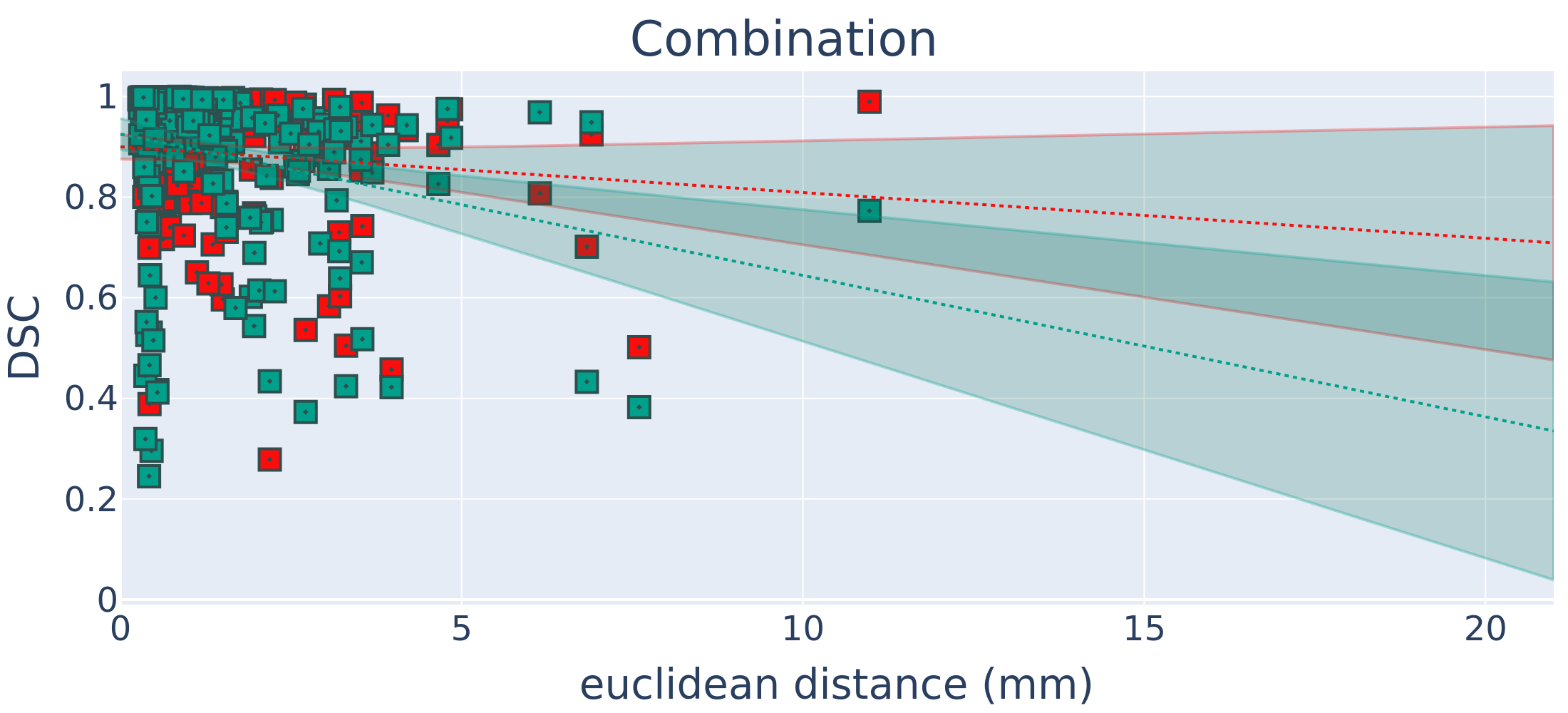}
            \caption{3D Models}
            \label{fig:robustness_3d_models}
        \end{subfigure}
    \end{minipage}\hfill
    \begin{minipage}{0.15\linewidth}
        \centering
        \includegraphics[width=\linewidth, trim={1600 180 0 180}, clip]{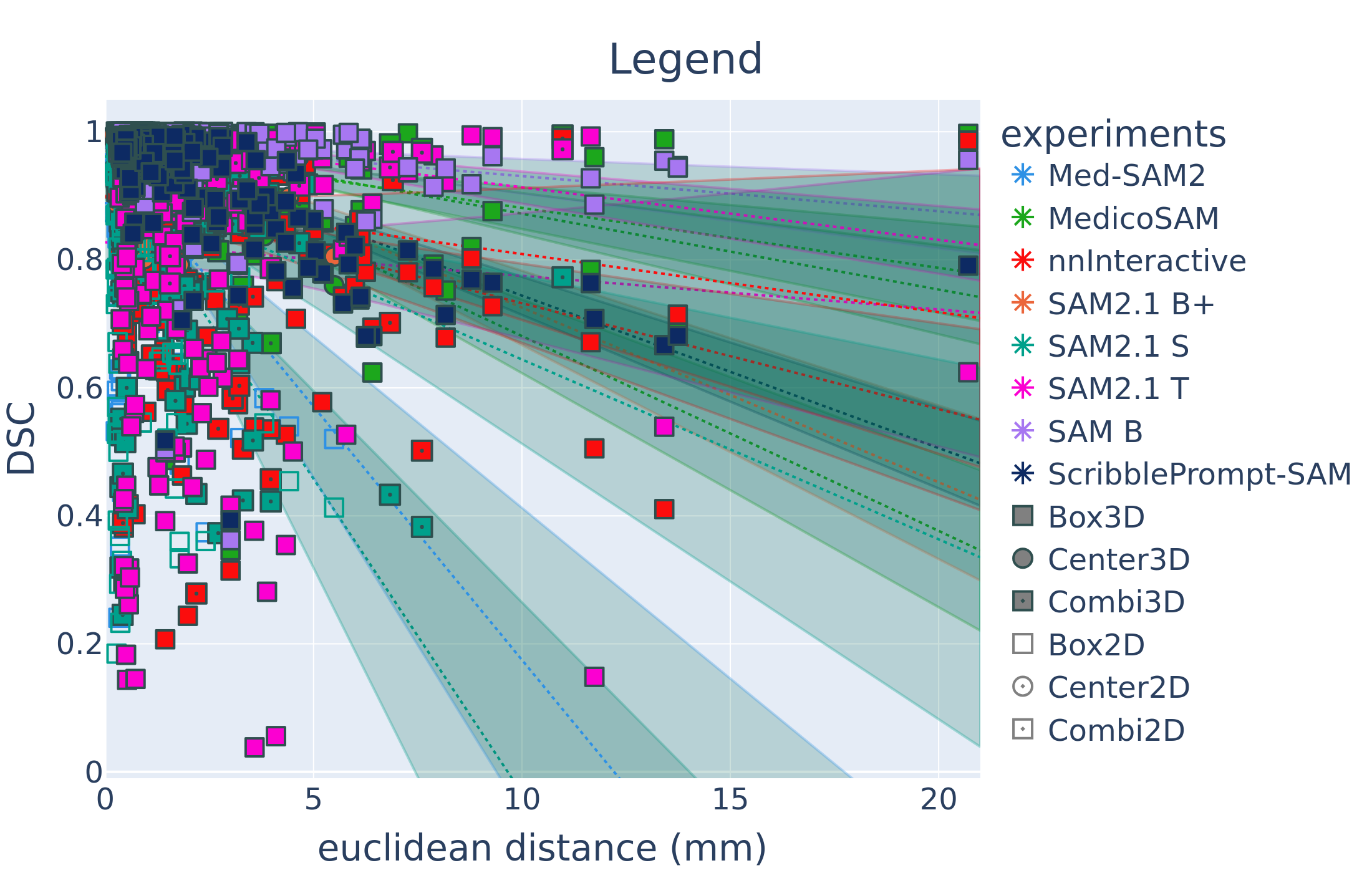}
    \end{minipage}
    \caption{Model sensitivity to input variations visualized as intra-rater annotation variability (euclidean distance or IoU) vs. segmentation consistency (DSC).
    \scriptsize{Each point represent the mean prompt variability and mean segmentation consistency for one sample. Dotted lines represent ordinary least squares (OLS) linear regression trends. Shaded areas denote the $95\% $confidence intervals ($\alpha = 0.05$).}}
    \label{fig:prompt_robustness}
\end{figure*}

For all models and prompt types, the correlation coefficient showed decreasing intra-rater segmentation consistency with increasing prompt variability (Table \ref{tab:segmentation_robustness}, Figure \ref{fig:prompt_robustness}). All 2D models and 3D models box-prompted showed statistically significant correlation ($p$-values$<0.05/(13\times3)=0.0013$) for intra-rater variability. Only nnInteractive combination-prompted and SAM2.1 T point-prompted were robust for intra-rater variability. These two models were analyzed for the inter-rater annotation variability and segmentation consistency in an iterative search pattern based on a sorted annotator pool (Table \ref{tab:inter_rater_ranking}), to identify the threshold for model sensitivity. The results showed that SAM2.1 T point-prompted was sensitive to the lowest inter-rater variability and nnInteractive combination-prompted was sensitive for the sixth lowest inter-rater variability (Table \ref{tab:segmentation_robustness_interrater}) ($p$-values $<0.05/(20\times3)=0.00083$).

\begin{table*}[h!]
\caption{Spearman’s rank correlation coefficients for each metric ($\rho_\text{DSC}$, $\rho_\text{NSD}$, $\rho_\text{HD95}$) between intra-rater annotation variability and segmentation consistency. \newline
\scriptsize{Asterisks ($*$) denote statistical significance. Positive values for HD95 and negative values for DSC and NSD indicate that increased prompt variability significantly reduces segmentation consistency.}}
\label{tab:segmentation_robustness}
\setlength{\tabcolsep}{3pt}
\renewcommand{\arraystretch}{1.3}
\resizebox{0.99\linewidth}{!}{
    \begin{tabular}{llc:ccc:ccc:cccc}
    \hline
    & \textbf{Model} & & \multicolumn{3}{:c:}{\textbf{Bounding Box} 2D \cblacksquare[0.2]{black} or 3D \cblacksquare[0.2]{white}} & \multicolumn{3}{c}{\textbf{Center Point} \cblackcircledot[0.25]{black} (2D) or \cblackcircledot[0.25]{white} (3D)} & \multicolumn{3}{:c}{\textbf{Combination} \cblacksquaredot[0.25]{black} (2D) or \cwhitesquaredot[0.25]{black} (3D)} \\ 
    & & Size & $\rho_\text{DSC}$ $\uparrow$ & $\rho_\text{NSD}$ $\uparrow$ & $\rho_\text{HD95}$ $\downarrow$ & $\rho_\text{DSC}$ $\uparrow$ & $\rho_\text{NSD}$ $\uparrow$ & $\rho_\text{HD95}$ $\downarrow$ & $\rho_\text{DSC}$ $\uparrow$ & $\rho_\text{NSD}$ $\uparrow$ & $\rho_\text{HD95}$ $\downarrow$ \\
    &  & M & \% & \% & mm & \% & \% & mm & \% & \% & mm \\  
    \hline

    & \multicolumn{11}{c}{\textbf{2D Models}} \\
    \hline
    \multirow{2}{*}{\rotatebox[origin=c]{25}{\footnotesize{medical}}} & MedicoSAM2D & 94 & -0.36* & -0.48* & 0.49* & -0.58* & -0.54* & 0.48* & -0.45* & -0.54* & 0.50* \\
    & ScribblePrompt-SAM & 94 & - & - & - & -0.57* & -0.55* & 0.59* & - & - & - \\
    \hdashline
    \multirow{3}{*}{\rotatebox[origin=c]{25}{\footnotesize{natural}}} & SAM B & 94 & - & - & - & -0.53* & -0.52* & 0.46* & - & - & - \\
    & SAM2.1 B+ & 81 & -0.33* & -0.38* & 0.41* & - & - & - & - & - & - \\
    & SAM2.1 T & 39 & - & - & - & - & - & - & -0.60* & -0.59* & 0.55* \\

    \hline
    & \multicolumn{11}{c}{\textbf{3D Models}} \\
    \hline
    \multirow{2}{*}{\rotatebox[origin=c]{25}{\footnotesize{medical}}} & Med-SAM2 & 39 & -0.38* & -0.41* & 0.50* & - & - & - & - & - & - \\
    & nnInteractive & 102 & - & - & - & -0.32* & -0.30* & 0.35* & \textbf{-0.09} & \textbf{-0.11} & \textbf{0.16} \\
    \hdashline
    \multirow{2}{*}{\rotatebox[origin=c]{25}{\footnotesize{natural}}} & SAM2.1 S & 46 & -0.23* & -0.29* & 0.41* & - & - & - & -0.31* & -0.32* & 0.41* \\
    & SAM2.1 T & 39 & - & - & - & \textbf{-0.05} & \textbf{-0.02} & \textbf{0.12} & - & - & - \\    
    \hline
    \end{tabular}
}
\end{table*}

\begin{table}[h!]
\caption{Spearman’s rank correlation coefficients for each metric ($\rho_\text{DSC}$, $\rho_\text{NSD}$, $\rho_\text{HD95}$) between inter-rater annotation variability and segmentation consistency. \newline 
\scriptsize{Annotator rows are ordered by euclidean distance (mm), starting with the lowest-variance rater. Due to iterative search, not all inter-rater variabilities are tested (Table \ref{tab:inter_rater_ranking}). The first column indicates the order of tests. Asterisks ($*$) denote statistical significance. Positive values for HD95 and negative values for DSC and NSD indicate that increased prompt variability significantly reduces segmentation consistency.}}
\label{tab:segmentation_robustness_interrater}
\setlength{\tabcolsep}{3pt}
\renewcommand{\arraystretch}{1.3}
\resizebox{0.99\linewidth}{!}{
    \begin{tabular}{lc:cccc}
    \hline
    & Annotator & Eucl. distance & $\rho_\text{DSC}$ $\uparrow$ & $\rho_\text{NSD}$ $\uparrow$ & $\rho_\text{HD95}$ $\downarrow$ \\
    & & (mm) & (\%) & (\%) & (mm) \\
    \hline

    \multicolumn{6}{c}{\textbf{nnInteractive} \cwhitesquaredot[0.25]{black}} \\
    1& Annotator02 & 1.67{\footnotesize±2.8} & -0.19 & -0.15 & 0.14  \\
    5& Annotator05 & 1.85{\footnotesize±3.2} & -0.21 & -0.21 & 0.18 \\
    6& Annotator14 & 1.86{\footnotesize±3.2} & -0.22 & 0.18 & -0.22 \\
    7& Annotator20 & 1.86{\footnotesize±3.2} & -0.26* & 0.20 & -0.26* \\
    4& \textbf{Annotator01} & \textbf{1.87}{\footnotesize±3.3} & -0.27* & -0.27* & 0.23* \\
    3& Annotator09 & 1.96{\footnotesize±3.1} & -0.25* & -0.27* & 0.24* \\
    2& Annotator07 & 2.40{\footnotesize±3.6} & -0.37* & -0.38* & 0.35* \\
    \hline
    \multicolumn{6}{c}{\textbf{SAM2.1 T } \cblackcircledot[0.25]{white}} \\
    1& \textbf{Annotator15} & \textbf{2.51}{\footnotesize±5.5} & -0.23* & 0.30* & -0.29* \\
    \hline
    \end{tabular}
}
\end{table}

\begin{figure*}[h]
    \centering
    \begin{subfigure}{\columnwidth}
        \includegraphics[width=\columnwidth]{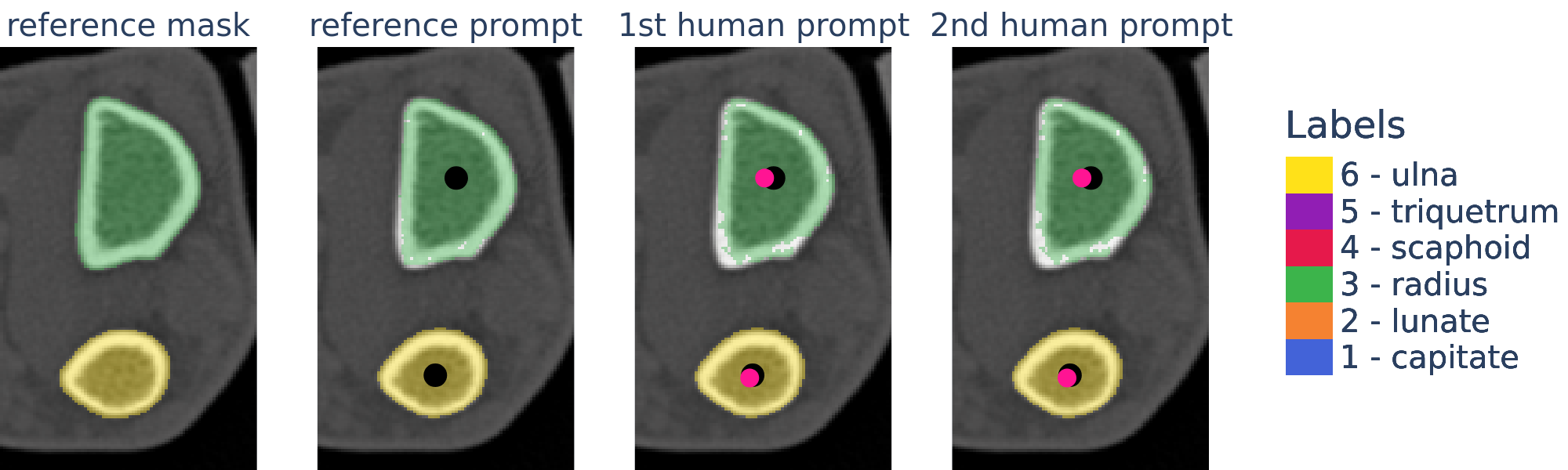}\newline
        \includegraphics[width=0.19\columnwidth, trim={150 0 180 0}, clip]{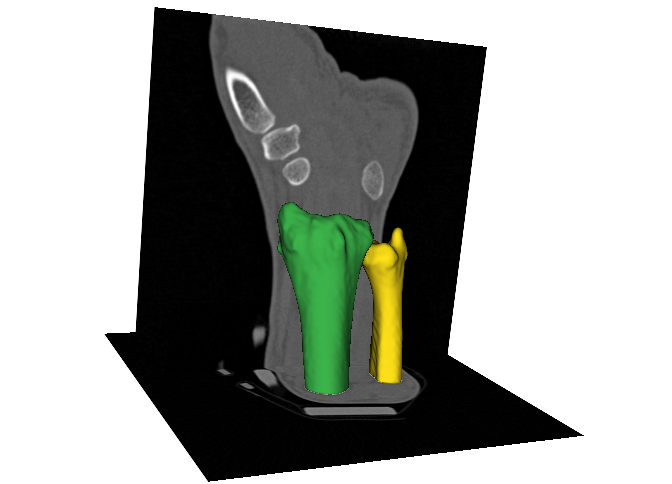}
        \includegraphics[width=0.19\columnwidth, trim={150 0 180 0}, clip]{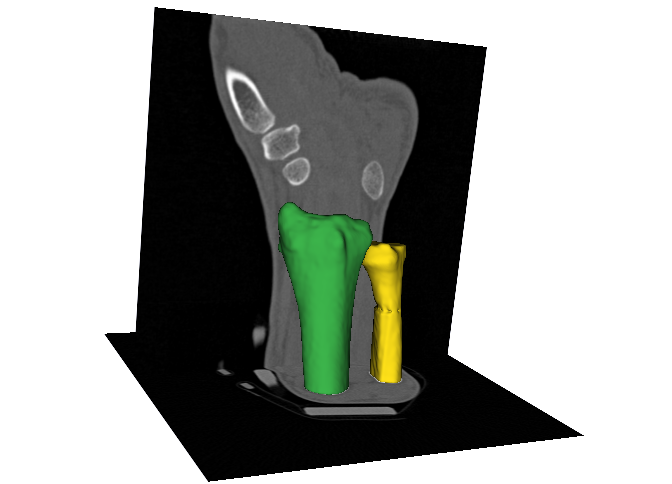}
        \includegraphics[width=0.19\columnwidth, trim={150 0 180 0}, clip]{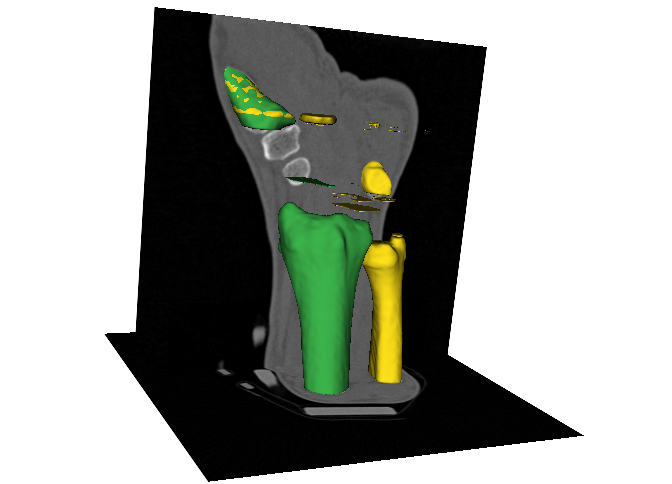}
        \includegraphics[width=0.19\columnwidth, trim={150 0 180 0}, clip]{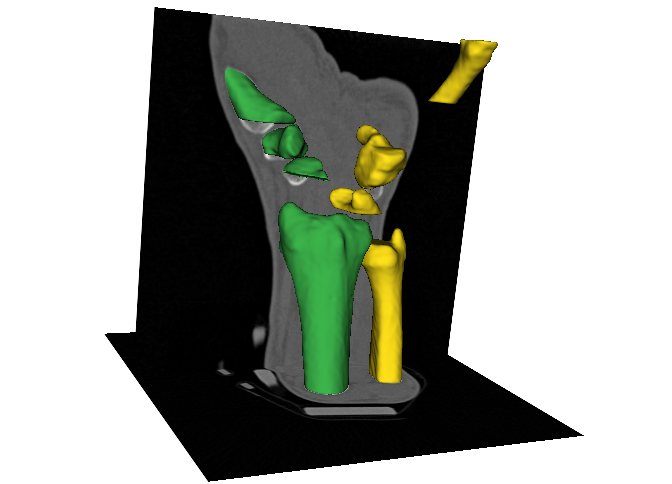}
        \caption{SAM2.1 T point-prompted; Wrist sample showing ulna and radius; Despite small intra-rater variation, the resulting 3D prediction shows large differences (72.5\% DSC, 69.0\% NSD, 24.8mm HD95).}
    \end{subfigure}\hfill
    \begin{subfigure}{\columnwidth}
        \includegraphics[width=\columnwidth]{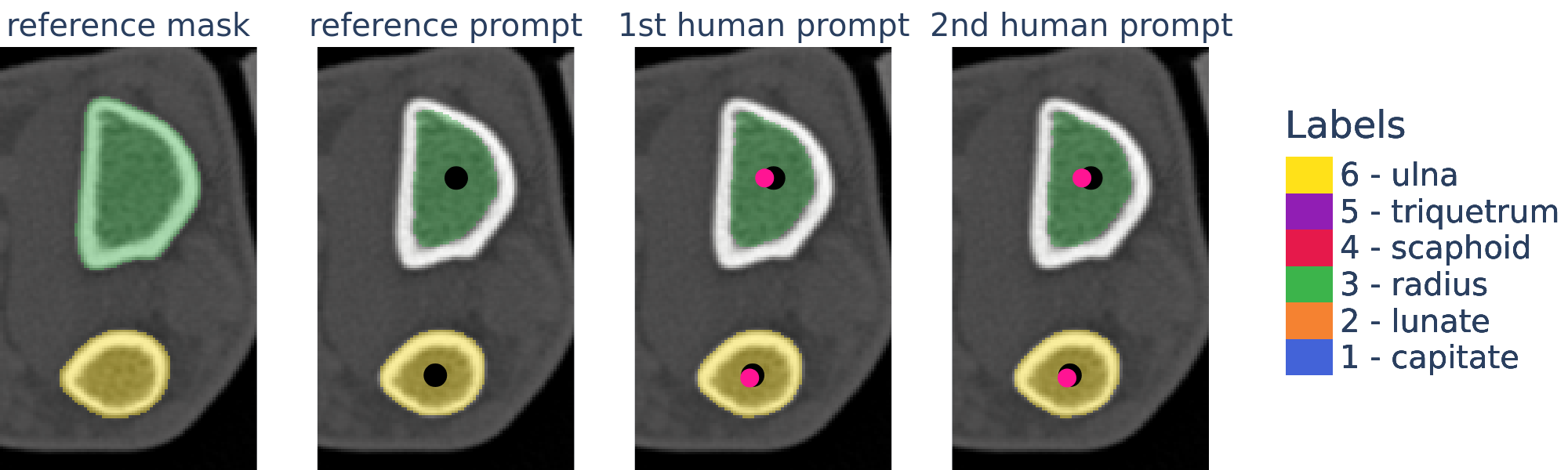}\newline
        \includegraphics[width=0.19\columnwidth, trim={150 0 180 0}, clip]{w7_1_gt_0.png}
        \includegraphics[width=0.19\columnwidth, trim={150 0 180 0}, clip]{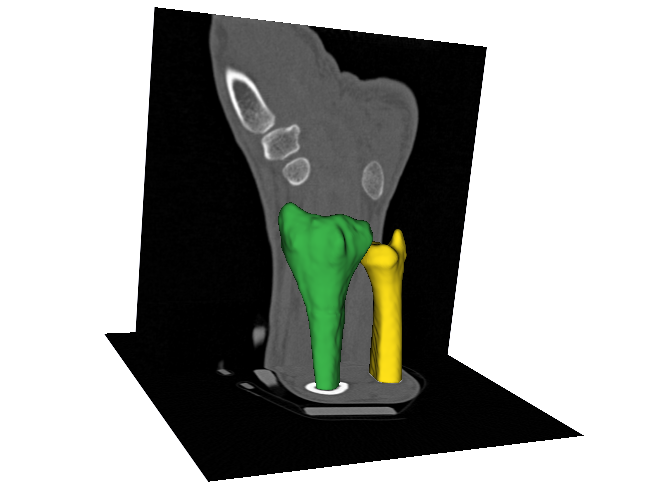}
        \includegraphics[width=0.19\columnwidth, trim={150 0 180 0}, clip]{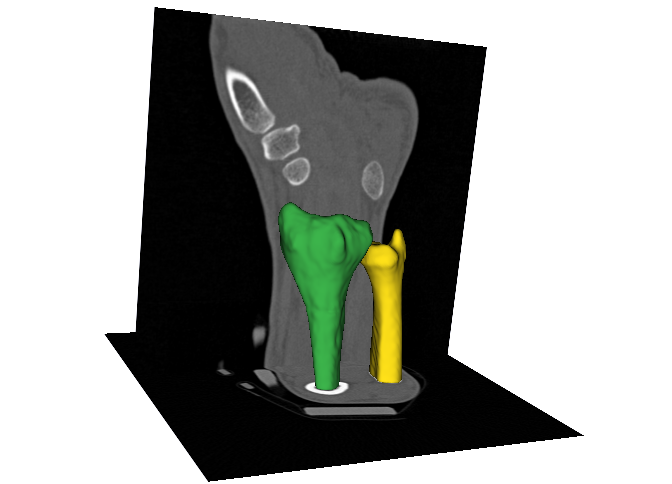}
        \includegraphics[width=0.19\columnwidth, trim={150 0 180 0}, clip]{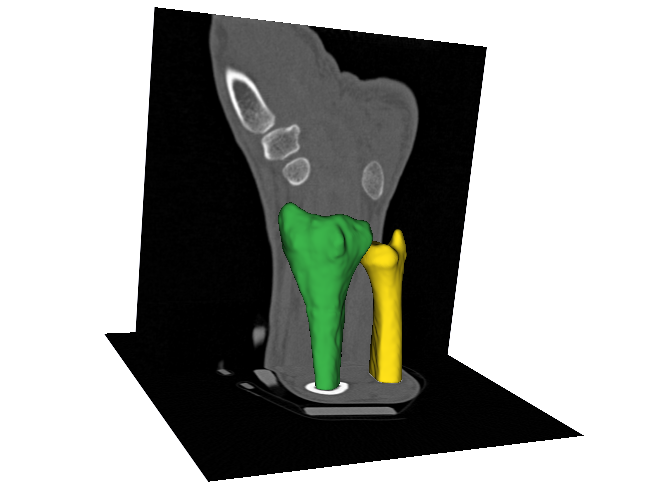}
        \caption{nnInteractive point-prompted; Wrist sample showing ulna and radius; Small intra-rater variation with small differences in resulting 3D prediction (98.7\% DSC, 100.0\% NSD, 0.3mm HD95).}
    \end{subfigure}\newline
    \begin{subfigure}{\columnwidth}
        \includegraphics[width=1\columnwidth]{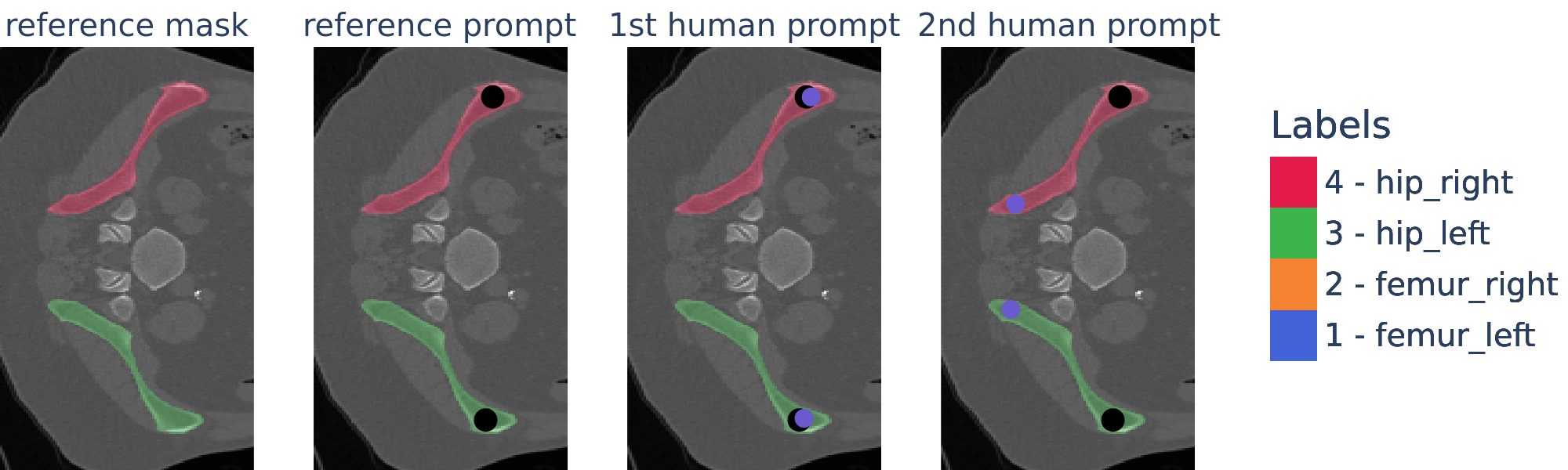}
        \includegraphics[width=1\columnwidth, trim={0 0 0 60}, clip]{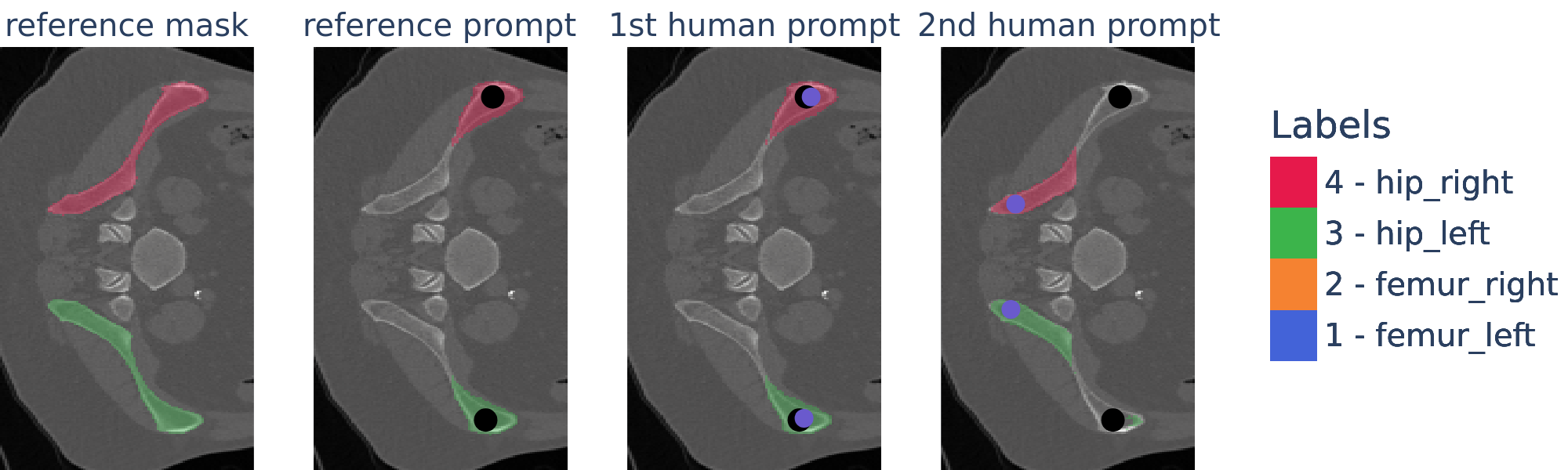}
        \caption{MedicoSAM2D (first row), ScribblePrompt-SAM (second row); Hip sample showing left/right hip; Varying model sensitivity for input prompt variations.}
    \end{subfigure}\hfill
    \begin{subfigure}{\columnwidth}
        \includegraphics[width=\columnwidth]{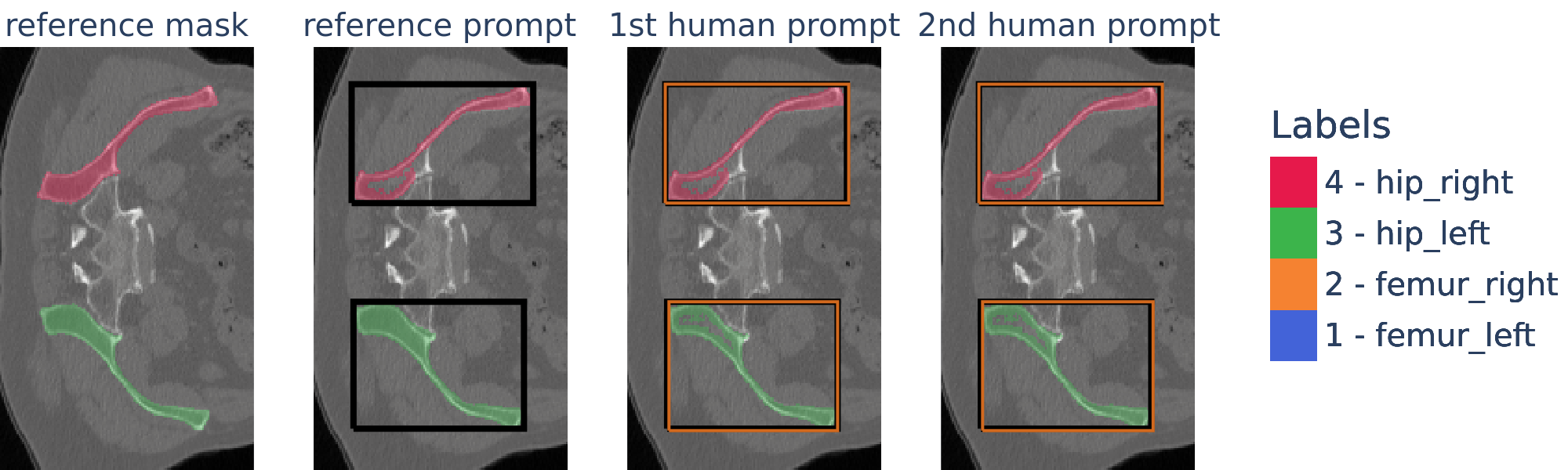}\newline
        \includegraphics[width=0.19\columnwidth, trim={150 0 180 0}, clip]{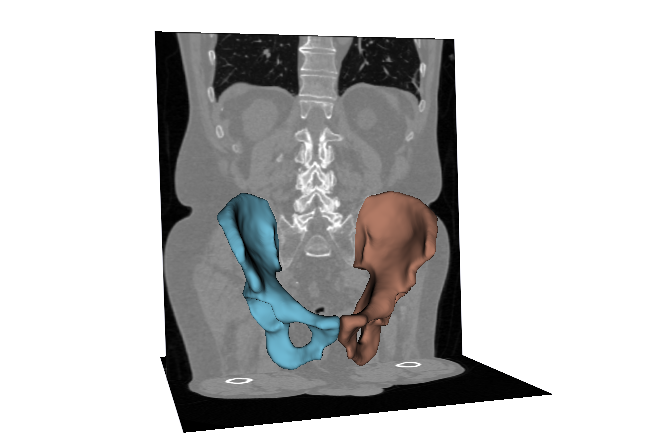}
        \includegraphics[width=0.19\columnwidth, trim={150 0 180 0}, clip]{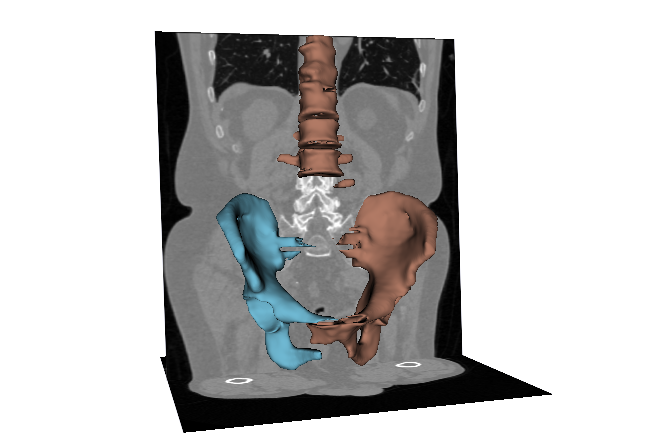}
        \includegraphics[width=0.19\columnwidth, trim={150 0 180 0}, clip]{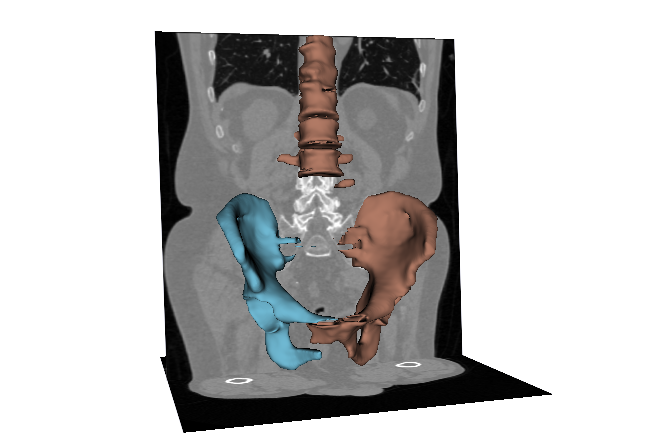}
        \includegraphics[width=0.19\columnwidth, trim={150 0 180 0}, clip]{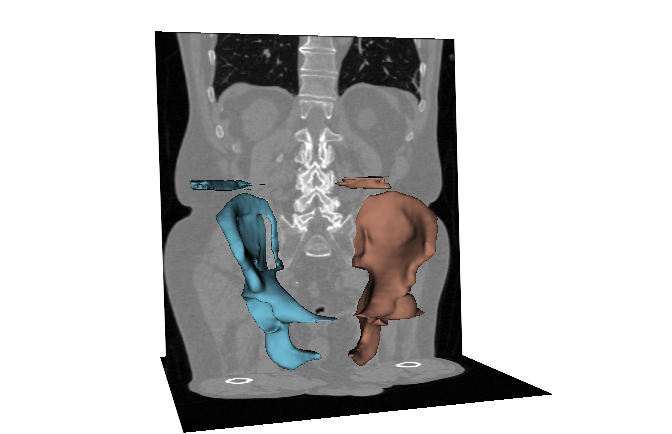}
        \caption{SAM2.1 S box-prompted; Hip sample showing left/right hip; Despite small intra-rater variation, the resulting 3D prediction shows large differences (64.1\% DSC, 62.9\% NSD, 48.6mm HD95) }
    \end{subfigure}
    \caption{Visual examples for model sensitivity to input prompt variations: reference mask, predicted mask with reference prompt, predicted mask with 1st set of human prompt and with 2nd set of the same annotator.
    \scriptsize{The reference prompt is drawn as black point or box. The human prompt is drawn as colored point or box.}}
    \label{fig:examples}
\end{figure*}

\section{Discussion}\label{discussion}

Our study quantified intra- and inter-rater variability in human prompts and analyzed their impact on segmentation consistency of Pareto-optimal FMs for MSK CT application, across four anatomical regions. The main findings in analyzing the model sensitivity to input prompt variations were: 1.) All 2D models showed sensitivity to prompt variations. 2.) 3D models SAM2.1 T point-prompted and nnInteractive combination-prompted showed robustness for intra-rater variations, but not for all inter-rater variations. 3.) Performance estimates of ``ideal'' prompting (i.e., reference prompts) do not translate to a human-driven setting. 

\subsection{Human prompt analysis}

There were considerable differences across data subsets and class labels for human prompts (Tables \ref{tab:ablation_human_point_performance}, \ref{tab:ablation_human_box_performance}), but some consistent findings emerged. Circular structures (e.g., humerus, wrist bones) showed high rater consistency. Point placement was more prone to deviation in elongated, thin, or annular bone shapes (e.g., scapula, femur with metal implant, see Figures \ref{fig:ablation_human_points_scatter}, \ref{fig:ablation_prompt_eval_point_examples}). For bounding boxes, consistency decreased in structures with complex topologies and multiple components (e.g., scapula, metal implants, see Figures \ref{fig:ablation_human_box_scatter}, \ref{fig:ablation_prompt_eval_boxes_examples}).
Overall, bounding box prompts demonstrated higher accuracy and consistency than point prompts, likely because defining a precise point for complex geometries is less intuitive for annotators than defining spatial boundaries. For instance, the hip appears as ``hourglass'' shape, making the center point definition less evident (Figure \ref{fig:examples}).

\subsection{Segmentation analysis}

In both 2D and 3D models, there were considerable performance variations for reference prompts across model types and prompting strategies(Figure \ref{fig:dsc_hd95_automatic_prompts}, Table \ref{tab:appendix_results_all_models}). For 2D medical FMs, MedicoSAM showed high performance compared to its alternatives Med-SAM and SAM-Med2D, which is likely due to its training on a complex objective (in contrast to Med-SAM) while keeping the SAM architecture without adapters (in contrast to SAM-Med2D) \cite{archit2025medicosam}. However, going to 3D, its propagation is outperformed by native 3D models such as Med-SAM2 and nnInteractive. While MedicoSAM3D projects the prediction of adjacent slices, Med-SAM2 leverages the memory bank mechanism of SAM2 and nnInteractive integrates user prompts as an additional input channel for 3D feature extraction, which is less prone to error propagation by design.\newline
Several 3D medical FMs perform significantly worse than others. For SAM-Med3D, resampling the entire image without cropping to 128×128×128 leads to notable loss of performance, likely due to image distortion and misalignment of the object of interest relative to the training data distribution. Even with cropping, performance remains below competitive levels. SegVol and Vista3D prompted with center points also demonstrate suboptimal results, which is likely due to the underlying training data, favoring abdominal and thoracic organ segmentation over bone and metal implant segmentation.

A direct comparison between 2D and 3D models is limited by fundamental differences in evaluation and prompting strategies. 3D performance was measured across entire volumes, where error propagation in more distal slices can lower overall metrics, whereas 2D models were evaluated on single slices without such penalties. In addition, 2D methods utilized prompts per component (i.e., multiple prompts per object), while 3D models were often restricted to a single prompt per object, especially for bounding box prompting. Due to these differences, we treated 2D and 3D models as two different categories of models in our analysis.

While performance findings remain consistent for both prompt sources (i.e., perfect and human), the performance drops for human prompts. This suggests that reference prompts are more ``ideal'' for optimizing model output, indicating that standard benchmarks might overestimate achievable performance in practical, human-driven settings. 

Visual inspection of segmentation results revealed three common mistakes (Figure \ref{fig:examples}), which also explain poor performance metrics: 1.) Anatomical Ambiguity: Due to the different Houndsfield Unit (HU) for cortical bone and trabecular bone, models struggled to differentiate between these structures and the combined total bone volume. This issue is caused by prompt ambiguity, where a point or box may not clearly define whether the user intends to segment the entire bone or just a specific layer. 2.) Oversegmentation: Both 2D and 3D architectures sometimes failed to identify clear anatomical boundaries. For 2D models, this typically resulted in the prediction extending beyond the bone contour within a single slice. For 3D models, these boundary failures were magnified by the additional spatial dimension, allowing errors to propagate and grow into neighboring structures, even across joint spaces. This propagation error suggests that the models lack a robust volumetric ``stop'' signal. 3.) Undersegmentation: In regions with fading or fluctuating intensity values, models sometimes stopped the predictions too early. 

\subsection{Model sensitivity to input prompt variations}

An inverse relationship was observed between prompt variability and segmentation consistency; as input prompt variability increases, segmentation consistency declines. While the high values for segmentation consistency suggest that the resulting masks remain mostly similar, the models nonetheless show sensitivity where even minor changes in the input prompt can trigger large changes in the output segmentation (Figure \ref{fig:prompt_robustness}, Figure \ref{fig:examples}). Consequently, sensitivity to prompt fluctuations should be considered a critical performance metric for the development and real-world evaluation of FMs, particularly in domains where user input can inherently vary.

Intra-rater annotation variability was consistently lower than even the most stable inter-rater setting for both point prompts (intra-rater Euclidean distance of $2.00 \pm 5.3$ mm vs. lowest pairwise Euclidean distance of $2.51 \pm 5.5$ mm) and combined prompts (intra-rater Euclidean distance of $1.41 \pm 2.4$ mm vs. lowest inter-rater Euclidean distance of $1.68 \pm 2.8$ mm). Therefore, if a model demonstrated sensitivity to the variations within a single annotator, it likely exhibits similar or greater sensitivity to the larger fluctuations between annotators. For models that did not show statistically significant correlation for the intra-rater setting, the correlation for inter-rater settings was tested as well. 
While all 2D models and box-prompted 3D models exhibited sensitivity to prompt fluctuations already for the intra-rater setting, nnInteractive combination-prompted and SAM 2.1 T point-prompted showed a lack of statistically significant correlations, suggesting model robustness to intra-rater input prompt variations. Testing them further with inter-rater settings, SAM2.1 T point-prompted showed sensitivity at the first inter-rater iterations ($2.51\pm5.5$), while nnInteractive combination-prompted showed sensitivity at the sixth inter-rater level with $1.87\pm3.3$. Thus, no tested model is robust against large fluctuations between annotators, but nnInteractive shows the least sensitivity. It is critical to emphasize that model sensitivity should not be viewed as an isolated performance metric. It must be evaluated in combination with absolute performance and segmentation consistency to ensure a more complete evaluation. Considering that, nnInteractive combination-prompted presented itself as the best option of all tested models.

\subsection{Limitations \& Future Work}

\paragraph{Dataset} The TotalSegmentator dataset was used to train some of the investigated FMs. Not all FMs reported a detailed train–test split (Table \ref{tab:model_prompt_overview}). However, by introducing the new classes femur implant left and right, the evaluated task extended beyond the original training labels and posed a new task unseen by the FMs, even if the selected test samples were included in previous training.
\paragraph{Axial slices} We limited our study to axial slices to limit the workload for annotators. Sagittal and coronal slices, which are often underexplored, could serve as a meaningful alternative or complementary source of information.
\paragraph{Observer Study} The annotators in our observer study are medical students rather than trained radiologists, primarily due to availability. However, the results indicate that extensive medical training may not be required for the investigated tasks, although this may not generalize to more complex clinical applications such as tumor identification. 
\paragraph{Non-iterative prompting} Our study was conducted in a static setting without iterative refinement or segmentation correction. While interactive workflows are important for real-world deployment, they increase the complexity of the evaluation, as the individual contributions of the interaction step and their effect on model sensitivity would be more difficult to isolate and quantify. Future evaluation studies should be conducted to analyze interactive refinement efficiency, which may mitigate commonly observed segmentation mistakes. For example, the impact of severe oversegmentation and volumetric leakage could be mitigated by the strategic use of negative prompts to define exclusion zones. Similarly, anatomical ambiguity could be overcome by several carefully placed positive prompts until the the desired anatomical boundary is reached. However, a disadvantage of iterative refinement is the additionally required user interaction and time, where the ultimate goal is often to fully automate the segmentation process without user interaction.
\paragraph{Geometric prompting} Aside from geometric prompting, also text prompts become more popular and are for example integrated in the recently released SAM3 framework \cite{carion2025sam3}. Text prompts remove user interaction and therefore geometric variations and could potentially be automatized for specific medical tasks if always the same structures should be segmented.

\section{Conclusion}\label{conclusion}
The observed performance drop when transitioning from idealized reference prompts to human inputs and the sensitivity to human prompt fluctuations across models, shows that prompt placement matters. Our findings suggest that segmentation performances derived from ``ideal'' (i.e., reference prompts) may not accurately reflect performance in human-driven settings. Consequently, model sensitivity to prompt variability should be established as a complementary performance metric for the development and real-world evaluation of promptable FMs. This would help bridging the gap between theoretical potential and practical application.

\section*{Acknowledgments}
We thank all the students, who participated in the observer study and made the collection of human prompts possible. We also want to thank Dieuwertje Luitse, for her input to study questionnaires sent to the students at the beginning and end of their study participation to collect additional information about the study participants and their study experience. We thank Thomas Koopman and the team from grand-challenge for their great help with setting up the observer study.

\bibliographystyle{elsarticle-num} 
\bibliography{bibliography}


\newpage
\clearpage

\appendix
\onecolumn

\section{Dataset}\label{sec:appendix_dataset}

\paragraph{Private Datasets} The three data subsets Wrist, Lower Leg, Shoulder were acquired at the Amsterdam UMC with a Brilliance 64-channel CT Scanner (Philips Healthcare, Best, The Netherlands) or a Siemens SOMATOM Force. The reference segmentation masks were generated in a two-step annotation process: First, an in-house 3D annotation software \cite{dobbe2014articulus} was used to generate preliminary mask with a threshold-based region-growing segmentation algorithm. Then, these preliminary masks were manually corrected and refined with ITK-SNAP \cite{itksnap}. 

\paragraph{Public Dataset} The fourth data subset Hip is derived from the publicly reported test set of TotalSegmentator \cite{wasserthal2023totalsegmentator}, a labeled CT dataset created by the Research and Analysis Department at University Hospital Basel. Following the official test split, we selected 11 CT scans, manually ensuring that 6 of them contained at least one hip implant. The reference segmentation mask was generated by merging the original reference mask with a manually created annotation in ITK-SNAP \cite{itksnap} of the hip implant (stem and cup together). The existing segmentation masks for the left and right hips, as well as the left and right femurs, were left unchanged; no corrections for over- or under-segmentation were applied.

\paragraph{Slice Selection} To reduce the workload in the observer study, axial slices were extracted from the 3D CT volumes taking into account data coverage, diversity and comparability between data subsets. To avoid slices with little to no relevant anatomical information, the top and bottom 10\% of each object were excluded from the slice selection. By default, two slices per class were extracted from the remaining object volume, maintaining at least a 10-slice interval (see Figure \ref{fig:dataset}). However, since the data subsets differ in their characteristics (e.g., number of classes and slices), the default setting was adjusted accordingly. For Wrist, a 5-slice gap was used because six classes were distributed across an average of 363 axial slices, making a 10-slice gap too large to maintain. To ensure a comparable number of slices across datasets and to account for the large volume size (over 1000 slices), three slices per class were selected for Lower Leg. For Hip, only the original classes were used for slice selection to ensure an equal number of slices per sample, as the two newly added labels do not appear in every CT scan.

\paragraph{Samples seen twice by annotators} 
A dataset-specific duplication strategy was applied. For the Wrist, Shoulder, and Hip datasets, a balanced approach was used by selecting one of the two selected slices per class label a second time (i.e., 50\% slices used twice). In contrast, all samples in the Lower Leg dataset were used a second time due to several dataset-specific characteristics: The number of classes per slice is limited (at most two reference classes), which reduces annotation time per sample; The majority of selected slices only contains one class, whereas slices in Wrist, Shoulder and Hip commonly display multiple classes; The extraction of three slices per class label precludes an even duplication split, unlike the other datasets.

\section{Model Implementation} \label{sec:appendix_models}

SAM, SAM2, Med-SAM,  Med-SAM2, SAM-Med2D, ScribblePrompt, SegVol, Vista3D, MedicoSAM2D, and nnInteractive were used as described by their GitHub repositories, including the provided tutorials and example scripts for data pre-processing\footnote{SAM: commit 6fdee8f, SAM2: commit 2b90b9f, Med-SAM: commit 2b7c64c, Med-SAM2: commit 332f30d, SAM-Med2D: commit bfd2b93, ScribblePrompt: commit 182449, SegVol:  4ee0a47, Vista3D: commit 8bb7572, MedicoSAM: 9d73c29, nnInteractive: 47c4626}.\newline

MedicoSAM3D \cite{archit2025medicosam} has three hyperparameter for prompt propagation: the IoU threshold, projection mode, and box extension factor, which controls the expansion of the box after projection. Optimal performance requires tuning these hyperparameter for each data subset. To establish a single standardized inference protocol for our entire dataset, we performed a grid-based hyperparameter search on four representative samples -- one from each subset, the same samples that participants from the observer study used for their training phase. The search space included IoU thresholds from 0.7 to 0.9 (step 0.1), projection modes box, points, points and masks, single point, and box extensions from 0 to 0.25 (step 0.05). The final settings, selected by majority vote from all experiments, were \texttt{iou\_threshold = 0.7}, \texttt{projection = single\_point}, and \texttt{box\_extension = 0.0}.\newline

The latest version of SAM-Med3D\footnote{SAM-Med3D commit: e8d2e0a} does not support sliding-window inference with built-in prompt propagation, in contrast to methods such as SegVol \cite{du2025segvol} or Vista3D \cite{he2024vista3d}. In its current implementation, inference operates on independent (128,128,128) window crops, each of which requires a newly provided prompt. Because the method does not implement an overlapping sliding window where prompts are automatically derived from the previously generated mask, the user needs to provide prompts for every crop. As our use case requires fully automatic inference after the initial prompt, this evaluation strategy cannot be applied.
To perform inference with SAM-Med3D, we implemented two alternatives without modifying the model framework: The first naive approach is to crop a (128,128,128) window around the initial prompt, which may fail to fully capture objects that exceed this size; The second is to resample the entire image by resizing its longest side to 128 voxels. Although this ensuring that the entire object is captures, it can significantly distort the image and affect the performance.\newline

MedicalSAM2 (MedSAM-2) \cite{zhu2024medicalsam2} was not included in our analysis due to persistent assertion errors in the model architecture code preventing successful execution\footnote{MedicalSAM2: commit 18b0f5b}, and resolving these issues would have required extensive investigation beyond the scope of this study. CT-SAM3D \cite{guo2024ctsam3d} was not included in our analysis because preliminary tests produced empty prediction masks. We hypothesize that the fixed 64×64×64 patch size in combination with the absence of a sliding-window inference or automatic prompt propagation (similar to SAM-Med3D) did not generalize well to our data.

\newpage

\section{Human prompt variation} \label{app:human_generated_prompts_eval}

\subsection{Accuracy of human prompts} \label{app:accuracy_point}

\paragraph{Center point}
Table \ref{tab:ablation_human_point_performance} collects detailed results on the median Euclidean distance (mm). Figure \ref{fig:ablation_human_points_scatter} visualizes the spatial distribution of the center point deviations ($\Delta x$, $\Delta y$) and the intra-rater consistency ($\Delta x$, $\Delta y$) per class label.

\begin{table*}[h]
\centering

\caption{Euclidean distances (mm) of human center points compared to reference center points measured as median and IQR.}
\label{tab:ablation_human_point_performance}

\begin{subtable}{\textwidth}
    \centering
    \caption{Total Average \& Dataset Lower Leg and corresponding class labels}
    \resizebox{0.6\textwidth}{!}{
    \begin{tabular}{lccccc}
    \hline
    Annotator & Total & Lower Leg & Tibia Implant & Tibia bone \\
    \midrule
    all & 1.50 (0.7-3.0) & 1.76 (1.0-1.0) & 1.54 (0.7-0.7) & 2.04 (1.1-1.1) \\
    annotator01 & 1.50 (0.7-0.7) & 1.95 (1.0-1.0) & 1.54 (0.7-0.7) & 2.04 (1.1-1.1) \\
    annotator02 & 1.50 (0.7-0.7) & 1.38 (0.7-0.7) & 1.54 (1.0-1.0) & 1.38 (0.7-0.7) \\
    annotator03 & 1.50 (0.8-0.8) & 2.01 (1.0-1.0) & 1.42 (0.7-0.7) & 2.01 (1.1-1.1) \\
    annotator04 & 1.50 (0.7-0.7) & 1.76 (1.0-1.0) & 1.46 (0.5-0.5) & 1.95 (1.1-1.1) \\
    annotator05 & 1.38 (0.7-0.7) & 1.54 (0.7-0.7) & 1.24 (0.5-0.5) & 1.54 (0.7-0.7) \\
    annotator06 & 1.63 (0.8-0.8) & 2.04 (1.0-1.0) & 1.78 (0.8-0.8) & 2.13 (1.1-1.1) \\
    annotator07 & 1.50 (0.7-0.7) & 2.85 (1.1-1.1) & 1.65 (1.0-1.0) & 3.59 (1.7-1.7) \\
    annotator08 & 1.37 (0.7-0.7) & 1.54 (1.0-1.0) & 1.76 (1.0-1.0) & 1.54 (1.0-1.0) \\
    annotator09 & 1.50 (0.8-0.8) & 1.95 (1.1-1.1) & 1.78 (1.0-1.0) & 1.95 (1.1-1.1) \\
    annotator10 & 1.59 (0.7-0.7) & 1.95 (1.1-1.1) & 1.09 (0.7-0.7) & 2.01 (1.2-1.2) \\
    annotator11 & 1.74 (1.0-1.0) & 1.76 (1.1-1.1) & 1.95 (1.1-1.1) & 1.76 (1.1-1.1) \\
    annotator12 & 1.66 (1.0-1.0) & 1.76 (1.1-1.1) & 1.76 (1.1-1.1) & 1.95 (1.1-1.1) \\
    annotator13 & 1.50 (0.8-0.8) & 1.54 (1.0-1.0) & 1.38 (0.7-0.7) & 1.95 (1.1-1.1) \\
    annotator14 & 1.46 (0.7-0.7) & 1.54 (1.0-1.0) & 1.09 (0.7-0.7) & 1.76 (1.0-1.0) \\
    annotator15 & 1.38 (0.7-0.7) & 1.09 (0.7-0.7) & 1.09 (0.7-0.7) & 1.09 (0.7-0.7) \\
    annotator16 & 1.71 (0.9-0.9) & 2.07 (1.1-1.1) & 1.50 (1.1-1.1) & 2.18 (1.5-1.5) \\
    annotator17 & 1.46 (0.7-0.7) & 1.76 (1.0-1.0) & 1.38 (0.7-0.7) & 2.01 (1.1-1.1) \\
    annotator18 & 1.50 (0.7-0.7) & 1.95 (1.1-1.1) & 1.95 (1.1-1.1) & 1.95 (1.1-1.1) \\
    annotator19 & 1.52 (0.7-0.7) & 2.01 (1.0-1.0) & 1.09 (0.5-0.5) & 2.31 (1.1-1.1) \\
    annotator20 & 1.54 (0.8-0.8) & 2.01 (1.1-1.1) & 2.07 (1.0-1.0) & 2.01 (1.4-1.4) \\
    \hline
    \end{tabular}
    }
\end{subtable}

\medskip

\begin{subtable}{\textwidth}
    \centering
    \caption{Dataset Shoulder and corresponding class labels}
    \resizebox{0.6\textwidth}{!}{
    \begin{tabular}{lcccccc}
    \hline
    Annotator & Shoulder & Humerus R & Scapula R & Humerus L & Scapula L \\
    \hline
    all & 1.86 (1.0-1.0) & 1.38 (1.0-1.0) & 1.67 (1.0-1.0) & 1.36 (0.9-0.9) & 2.18 (1.2-1.2) \\
    annotator01 & 1.38 (1.0-1.0) & 1.38 (1.0-1.0) & 1.67 (1.0-1.0) & 1.36 (0.9-0.9) & 2.18 (1.2-1.2) \\
    annotator02 & 1.91 (1.0-1.0) & 1.29 (1.0-1.0) & 2.17 (1.2-1.2) & 1.56 (1.0-1.0) & 2.04 (1.2-1.2) \\
    annotator03 & 1.94 (1.2-1.2) & 1.89 (1.0-1.0) & 2.18 (1.6-1.6) & 1.38 (1.0-1.0) & 2.18 (1.2-1.2) \\
    annotator04 & 1.91 (1.0-1.0) & 1.69 (1.0-1.0) & 2.50 (1.4-1.4) & 1.22 (0.9-0.9) & 2.46 (1.4-1.4) \\
    annotator05 & 1.38 (1.0-1.0) & 1.21 (1.0-1.0) & 1.86 (1.0-1.0) & 1.38 (1.0-1.0) & 1.38 (1.0-1.0) \\
    annotator06 & 1.91 (1.0-1.0) & 1.38 (1.0-1.0) & 2.76 (1.5-1.5) & 1.29 (0.9-0.9) & 3.08 (1.3-1.3) \\
    annotator07 & 1.66 (1.0-1.0) & 1.69 (1.0-1.0) & 1.91 (1.2-1.2) & 1.36 (1.0-1.0) & 1.86 (1.0-1.0) \\
    annotator08 & 1.37 (1.0-1.0) & 1.28 (1.0-1.0) & 1.89 (1.0-1.0) & 1.18 (0.9-0.9) & 1.38 (1.0-1.0) \\
    annotator09 & 1.66 (1.0-1.0) & 1.38 (1.0-1.0) & 2.18 (1.0-1.0) & 1.18 (1.0-1.0) & 1.94 (1.3-1.3) \\
    annotator10 & 1.94 (1.0-1.0) & 0.98 (0.8-0.8) & 4.03 (2.7-2.7) & 0.98 (0.9-0.9) & 4.03 (1.9-1.9) \\
    annotator11 & 1.95 (1.2-1.2) & 1.86 (1.4-1.4) & 2.18 (1.9-1.9) & 1.29 (1.0-1.0) & 2.36 (1.4-1.4) \\
    annotator12 & 1.94 (1.2-1.2) & 1.89 (1.2-1.2) & 2.18 (1.7-1.7) & 1.52 (1.0-1.0) & 1.95 (1.4-1.4) \\
    annotator13 & 1.86 (1.0-1.0) & 1.37 (1.0-1.0) & 1.95 (1.2-1.2) & 1.38 (1.0-1.0) & 2.30 (1.4-1.4) \\
    annotator14 & 1.86 (1.2-1.2) & 1.69 (1.0-1.0) & 2.18 (1.7-1.7) & 1.37 (1.0-1.0) & 2.18 (1.2-1.2) \\
    annotator15 & 1.38 (1.0-1.0) & 1.21 (1.0-1.0) & 1.95 (1.2-1.2) & 1.19 (1.0-1.0) & 1.94 (1.0-1.0) \\
    annotator16 & 1.95 (1.2-1.2) & 1.94 (1.2-1.2) & 2.53 (1.7-1.7) & 1.30 (1.0-1.0) & 2.27 (1.9-1.9) \\
    annotator17 & 1.38 (1.0-1.0) & 1.23 (1.0-1.0) & 1.95 (1.2-1.2) & 1.18 (1.0-1.0) & 1.94 (1.2-1.2) \\
    annotator18 & 1.38 (1.0-1.0) & 1.26 (1.0-1.0) & 1.94 (1.1-1.1) & 1.19 (1.0-1.0) & 1.94 (1.4-1.4) \\
    annotator19 & 1.94 (1.2-1.2) & 1.38 (1.0-1.0) & 2.36 (1.4-1.4) & 1.22 (1.0-1.0) & 3.44 (1.9-1.9) \\
    annotator20 & 1.91 (1.0-1.0) & 1.69 (1.0-1.0) & 2.06 (1.4-1.4) & 1.38 (1.0-1.0) & 2.17 (1.4-1.4) \\
    \hline
    \end{tabular}
    }
\end{subtable}
\end{table*}

\begin{table*}[h]\ContinuedFloat
\begin{subtable}{\textwidth}
    \centering
    \caption{Dataset Wrist and corresponding class labels}
    \resizebox{0.95\textwidth}{!}{
    \begin{tabular}{lcccccccc}
    \hline
    Annotator & Wrist & Capitate & Lunate & Radius & Scaphoid & Triquetrum & Ulna \\
    \hline
    all & 0.73 (0.5-0.5) & 0.65 (0.5-0.5) & 0.95 (0.7-0.7) & 0.65 (0.3-0.3) & 0.73 (0.6-0.6) & 0.73 (0.5-0.5) & 0.46 (0.3-0.3) \\
    annotator01 & 0.73 (0.5-0.5) & 0.65 (0.5-0.5) & 0.95 (0.7-0.7) & 0.65 (0.3-0.3) & 0.73 (0.6-0.6) & 0.73 (0.5-0.5) & 0.46 (0.3-0.3) \\
    annotator02 & 0.73 (0.5-0.5) & 0.65 (0.3-0.3) & 1.17 (0.7-0.7) & 0.73 (0.5-0.5) & 0.73 (0.6-0.6) & 0.65 (0.3-0.3) & 0.65 (0.4-0.4) \\
    annotator03 & 0.73 (0.5-0.5) & 0.65 (0.5-0.5) & 1.46 (0.7-0.7) & 0.92 (0.5-0.5) & 0.73 (0.5-0.5) & 0.69 (0.5-0.5) & 0.46 (0.3-0.3) \\
    annotator04 & 0.73 (0.5-0.5) & 0.65 (0.4-0.4) & 1.17 (0.7-0.7) & 0.73 (0.3-0.3) & 0.92 (0.7-0.7) & 0.73 (0.5-0.5) & 0.46 (0.3-0.3) \\
    annotator05 & 0.65 (0.3-0.3) & 0.65 (0.3-0.3) & 0.73 (0.5-0.5) & 0.65 (0.4-0.4) & 0.65 (0.3-0.3) & 0.46 (0.3-0.3) & 0.46 (0.4-0.4) \\
    annotator06 & 0.73 (0.5-0.5) & 0.73 (0.3-0.3) & 1.03 (0.7-0.7) & 0.73 (0.3-0.3) & 0.73 (0.7-0.7) & 0.73 (0.5-0.5) & 0.65 (0.3-0.3) \\
    annotator07 & 0.73 (0.5-0.5) & 0.73 (0.5-0.5) & 0.92 (0.7-0.7) & 0.65 (0.3-0.3) & 0.98 (0.7-0.7) & 0.69 (0.5-0.5) & 0.65 (0.5-0.5) \\
    annotator08 & 0.73 (0.5-0.5) & 0.73 (0.5-0.5) & 0.73 (0.5-0.5) & 0.65 (0.3-0.3) & 0.73 (0.5-0.5) & 0.73 (0.5-0.5) & 0.65 (0.5-0.5) \\
    annotator09 & 0.73 (0.5-0.5) & 0.69 (0.3-0.3) & 1.03 (0.7-0.7) & 0.73 (0.7-0.7) & 0.92 (0.7-0.7) & 0.73 (0.5-0.5) & 0.46 (0.3-0.3) \\
    annotator10 & 0.73 (0.5-0.5) & 0.65 (0.3-0.3) & 1.03 (0.5-0.5) & 0.69 (0.4-0.4) & 0.98 (0.7-0.7) & 0.46 (0.3-0.3) & 0.46 (0.4-0.4) \\
    annotator11 & 0.73 (0.5-0.5) & 0.73 (0.3-0.3) & 1.17 (0.7-0.7) & 0.73 (0.5-0.5) & 0.95 (0.7-0.7) & 0.92 (0.5-0.5) & 0.69 (0.4-0.4) \\
    annotator12 & 0.73 (0.5-0.5) & 0.73 (0.5-0.5) & 1.17 (0.7-0.7) & 0.82 (0.5-0.5) & 0.92 (0.7-0.7) & 0.73 (0.5-0.5) & 0.73 (0.5-0.5) \\
    annotator13 & 0.73 (0.5-0.5) & 0.65 (0.3-0.3) & 0.92 (0.7-0.7) & 0.65 (0.4-0.4) & 0.73 (0.7-0.7) & 0.65 (0.4-0.4) & 0.46 (0.3-0.3) \\
    annotator14 & 0.73 (0.3-0.3) & 0.73 (0.3-0.3) & 0.92 (0.5-0.5) & 0.65 (0.3-0.3) & 0.73 (0.5-0.5) & 0.65 (0.5-0.5) & 0.46 (0.3-0.3) \\
    annotator15 & 0.73 (0.5-0.5) & 0.46 (0.3-0.3) & 1.26 (0.7-0.7) & 0.69 (0.5-0.5) & 0.73 (0.5-0.5) & 0.65 (0.3-0.3) & 0.73 (0.5-0.5) \\
    annotator16 & 0.73 (0.5-0.5) & 0.73 (0.5-0.5) & 0.98 (0.7-0.7) & 0.73 (0.5-0.5) & 0.98 (0.7-0.7) & 0.65 (0.3-0.3) & 0.65 (0.5-0.5) \\
    annotator17 & 0.65 (0.5-0.5) & 0.65 (0.3-0.3) & 1.00 (0.6-0.6) & 0.65 (0.3-0.3) & 0.73 (0.5-0.5) & 0.73 (0.4-0.4) & 0.65 (0.5-0.5) \\
    annotator18 & 0.73 (0.5-0.5) & 0.65 (0.3-0.3) & 0.92 (0.5-0.5) & 0.65 (0.3-0.3) & 0.73 (0.5-0.5) & 0.73 (0.5-0.5) & 0.73 (0.4-0.4) \\
    annotator19 & 0.65 (0.4-0.4) & 0.46 (0.3-0.3) & 1.10 (0.5-0.5) & 0.65 (0.5-0.5) & 0.98 (0.7-0.7) & 0.46 (0.3-0.3) & 0.46 (0.3-0.3) \\
    annotator20 & 0.73 (0.5-0.5) & 0.73 (0.3-0.3) & 1.17 (0.7-0.7) & 0.46 (0.5-0.5) & 0.98 (0.7-0.7) & 0.73 (0.5-0.5) & 0.56 (0.3-0.3) \\
    \hline
    \end{tabular}    
    }
\end{subtable}

\begin{subtable}{\textwidth}
    \centering
    \caption{Dataset Hip and corresponding class labels}
    \resizebox{0.95\textwidth}{!}{
    \begin{tabular}{lcccccccc}
    \hline
    Annotator & Hip & Femur L & Femur R & Hip L & Hip R & Femur Implant L & Femur Implant R \\
    \hline
    all & 3.35 (2.1-2.1) & 3.00 (2.1-2.1) & 3.35 (2.1-2.1) & 3.35 (2.1-2.1) & 3.00 (2.1-2.1) & 1.50 (1.5-1.5) & 2.12 (1.5-1.5) \\
    annotator01 & 3.00 (2.1-2.1) & 3.00 (2.1-2.1) & 3.35 (2.1-2.1) & 3.35 (2.1-2.1) & 3.00 (2.1-2.1) & 1.50 (1.5-1.5) & 2.12 (1.5-1.5) \\
    annotator02 & 3.35 (2.1-2.1) & 3.18 (2.1-2.1) & 5.41 (3.0-3.0) & 4.74 (3.3-3.3) & 4.37 (3.0-3.0) & 2.12 (1.5-1.5) & 2.12 (1.5-1.5) \\
    annotator03 & 3.35 (2.1-2.1) & 3.35 (1.7-1.7) & 4.50 (2.1-2.1) & 4.24 (2.1-2.1) & 4.74 (3.0-3.0) & 1.50 (1.5-1.5) & 2.12 (1.5-1.5) \\
    annotator04 & 3.35 (2.1-2.1) & 3.00 (2.1-2.1) & 4.24 (2.1-2.1) & 5.41 (3.0-3.0) & 4.74 (3.0-3.0) & 2.12 (1.5-1.5) & 2.12 (1.5-1.5) \\
    annotator05 & 3.00 (1.5-1.5) & 2.12 (2.1-2.1) & 3.18 (1.5-1.5) & 3.35 (2.1-2.1) & 3.00 (1.5-1.5) & 1.50 (0.0-0.0) & 2.12 (1.5-1.5) \\
    annotator06 & 3.35 (2.1-2.1) & 3.00 (1.5-1.5) & 4.74 (3.4-3.4) & 4.37 (2.1-2.1) & 3.35 (2.1-2.1) & 2.12 (1.5-1.5) & 2.12 (1.5-1.5) \\
    annotator07 & 3.18 (1.5-1.5) & 3.00 (1.5-1.5) & 4.50 (2.1-2.1) & 3.35 (2.1-2.1) & 3.35 (1.5-1.5) & 1.50 (1.5-1.5) & 2.12 (1.5-1.5) \\
    annotator08 & 3.00 (1.5-1.5) & 2.12 (1.5-1.5) & 3.35 (3.0-3.0) & 3.35 (2.1-2.1) & 2.12 (1.5-1.5) & 2.12 (2.1-2.1) & 2.12 (1.5-1.5) \\
    annotator09 & 3.35 (2.1-2.1) & 3.35 (2.1-2.1) & 4.74 (3.3-3.3) & 4.24 (3.0-3.0) & 4.24 (2.1-2.1) & 1.50 (1.5-1.5) & 3.35 (2.1-2.1) \\
    annotator10 & 4.50 (2.1-2.1) & 2.12 (1.5-1.5) & 4.50 (3.0-3.0) & 6.71 (3.4-3.4) & 7.50 (3.4-3.4) & 1.50 (1.5-1.5) & 2.12 (1.5-1.5) \\
    annotator11 & 3.35 (2.1-2.1) & 3.35 (1.5-1.5) & 4.74 (3.4-3.4) & 4.74 (3.3-3.3) & 3.35 (2.1-2.1) & 1.50 (0.0-0.0) & 2.12 (1.5-1.5) \\
    annotator12 & 3.35 (2.1-2.1) & 3.35 (2.1-2.1) & 5.41 (3.4-3.4) & 3.35 (2.1-2.1) & 4.24 (2.3-2.3) & 1.50 (1.5-1.5) & 1.50 (1.5-1.5) \\
    annotator13 & 3.35 (2.1-2.1) & 2.12 (2.1-2.1) & 4.24 (2.1-2.1) & 4.24 (3.0-3.0) & 3.35 (2.1-2.1) & 1.50 (1.5-1.5) & 2.12 (1.5-1.5) \\
    annotator14 & 3.35 (2.1-2.1) & 3.00 (1.5-1.5) & 6.35 (3.0-3.0) & 3.35 (2.1-2.1) & 3.35 (2.1-2.1) & 1.50 (1.5-1.5) & 2.12 (1.5-1.5) \\
    annotator15 & 3.35 (1.5-1.5) & 2.12 (1.5-1.5) & 4.95 (1.5-1.5) & 4.74 (3.4-3.4) & 4.50 (2.1-2.1) & 1.50 (0.0-0.0) & 1.50 (1.5-1.5) \\
    annotator16 & 3.35 (2.1-2.1) & 2.12 (1.5-1.5) & 4.24 (2.1-2.1) & 5.41 (3.4-3.4) & 4.50 (2.8-2.8) & 2.12 (1.5-1.5) & 2.12 (1.5-1.5) \\
    annotator17 & 3.00 (1.5-1.5) & 2.12 (1.5-1.5) & 3.35 (2.1-2.1) & 3.35 (1.5-1.5) & 3.00 (2.1-2.1) & 2.12 (1.5-1.5) & 2.12 (1.5-1.5) \\
    annotator18 & 3.35 (1.5-1.5) & 2.12 (1.5-1.5) & 4.24 (2.1-2.1) & 3.35 (2.1-2.1) & 4.50 (2.1-2.1) & 1.50 (1.5-1.5) & 2.12 (1.5-1.5) \\
    annotator19 & 3.35 (2.1-2.1) & 3.18 (1.5-1.5) & 4.24 (3.0-3.0) & 5.41 (3.0-3.0) & 4.74 (2.1-2.1) & 1.50 (1.5-1.5) & 1.50 (1.5-1.5) \\
    annotator20 & 3.35 (2.1-2.1) & 2.74 (2.1-2.1) & 3.35 (3.0-3.0) & 4.24 (2.1-2.1) & 3.35 (2.1-2.1) & 1.50 (1.5-1.5) & 2.12 (1.5-1.5) \\
    \hline
    \end{tabular}
    }
\end{subtable}
\end{table*}

\begin{figure*}[h]
    \centering
    \begin{subfigure}[t]{0.9\textwidth}
      \includegraphics[width=1\linewidth]{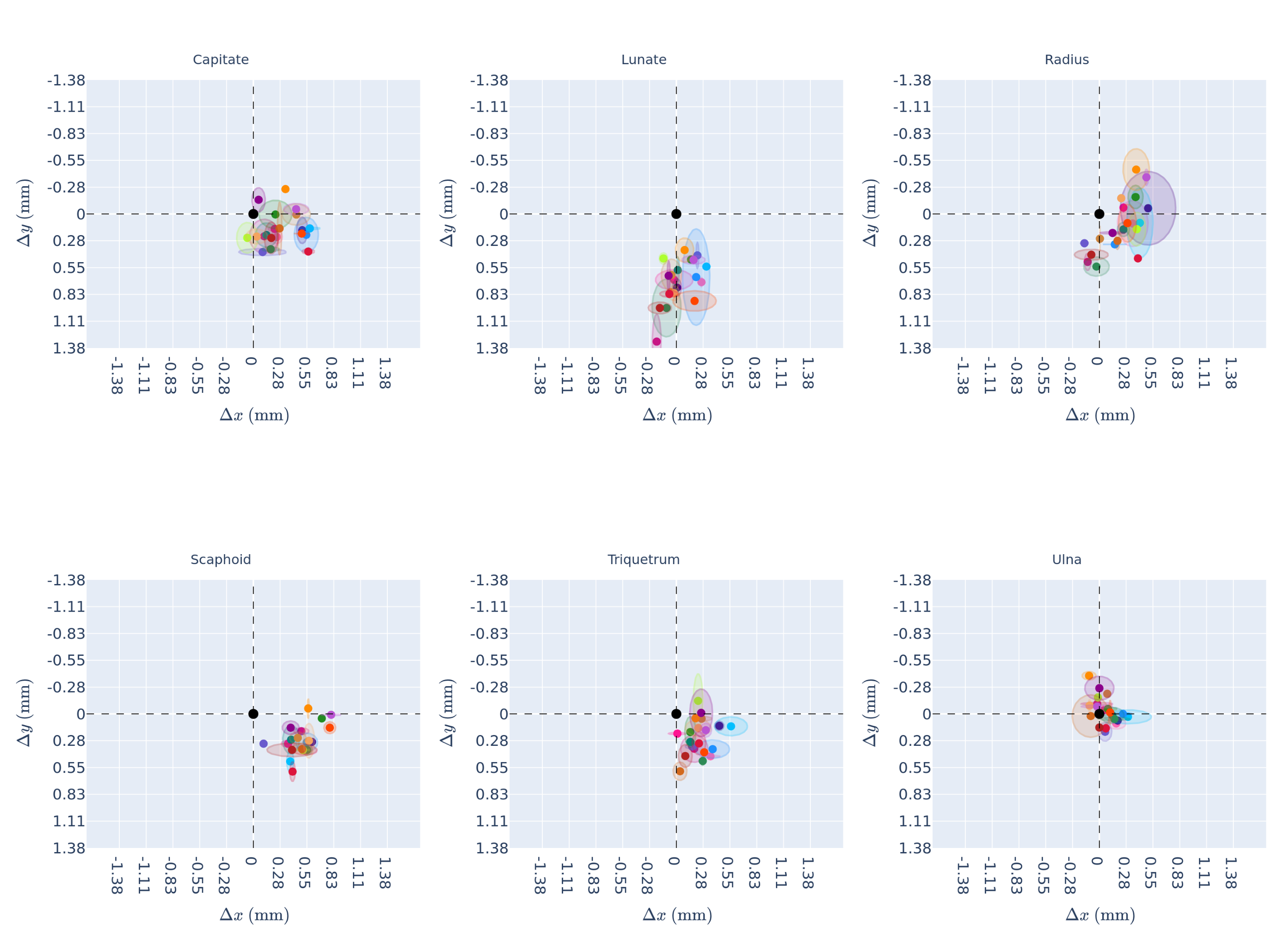}
      \caption{Wrist}
    \end{subfigure} 
    \begin{subfigure}[t]{0.9\textwidth}
      \includegraphics[width=1\linewidth]{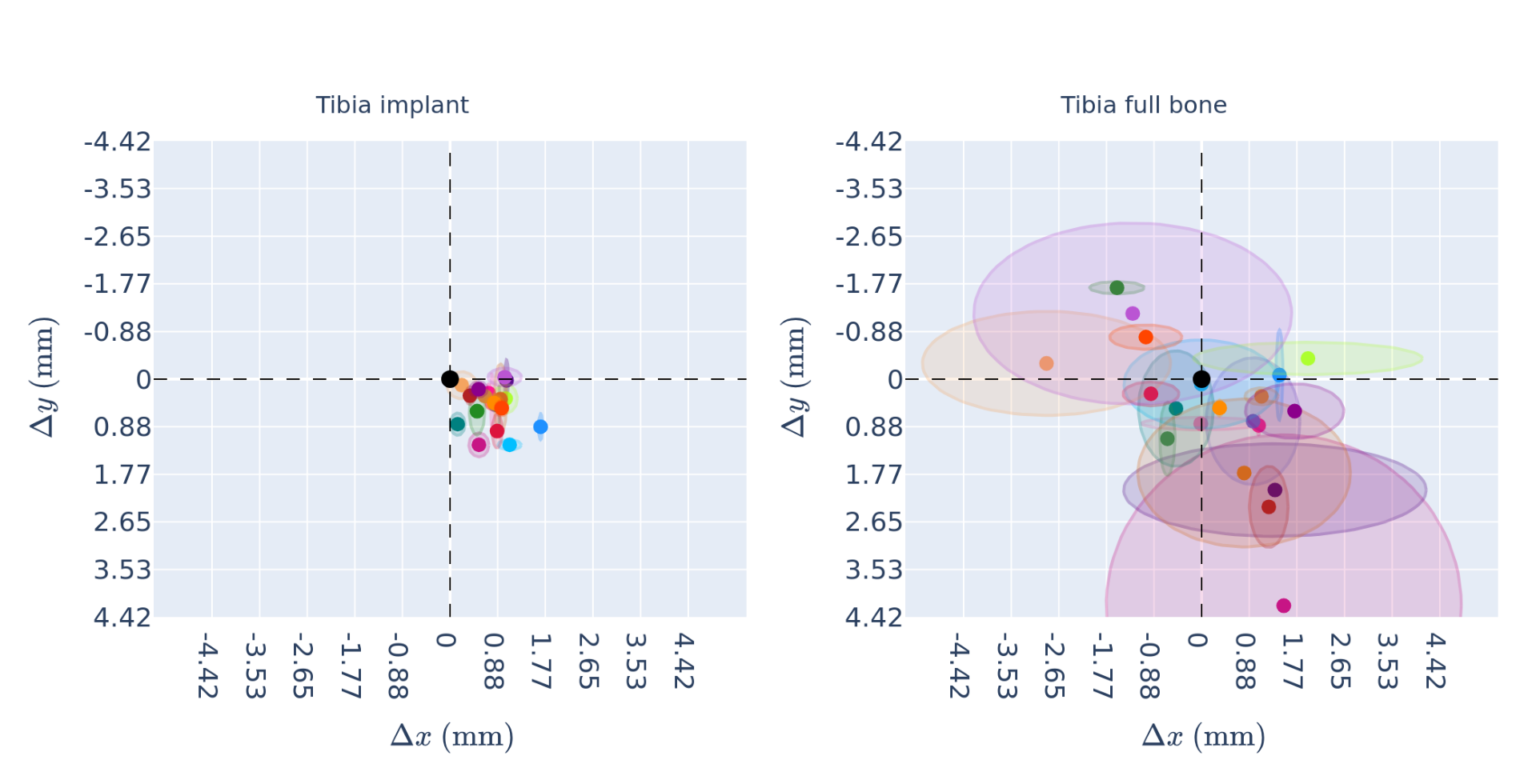}
      \caption{Lower Leg}
      \end{subfigure}
\end{figure*}

\begin{figure*}[h]\ContinuedFloat
    \centering
     \begin{subfigure}[t]{0.9\textwidth}
      \includegraphics[width=1\linewidth]{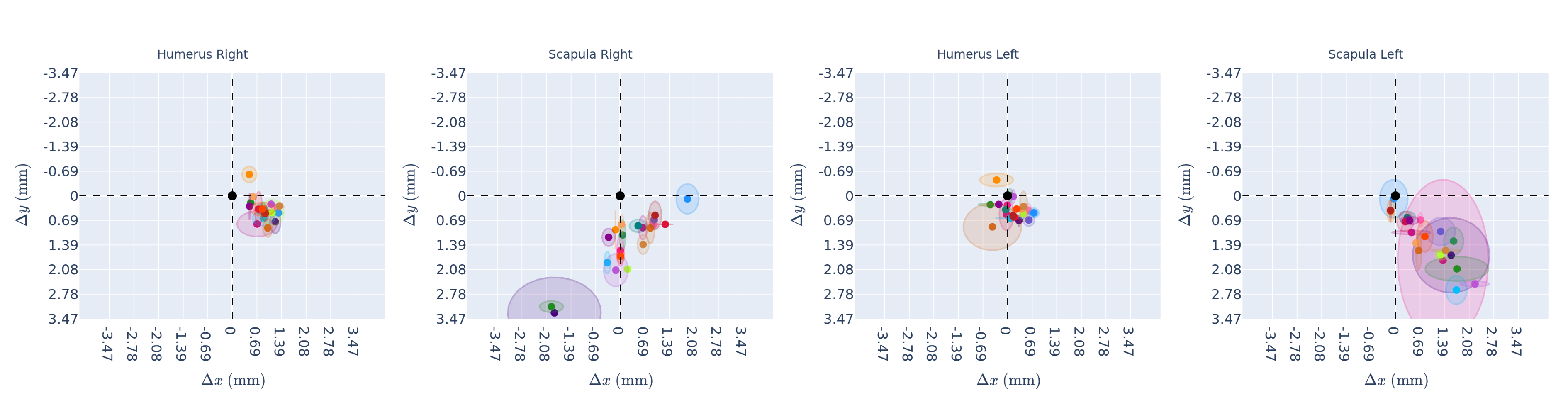}
      \caption{Shoulder}
    \end{subfigure} 
    \begin{subfigure}[t]{0.9\textwidth}
      \includegraphics[width=1\linewidth]{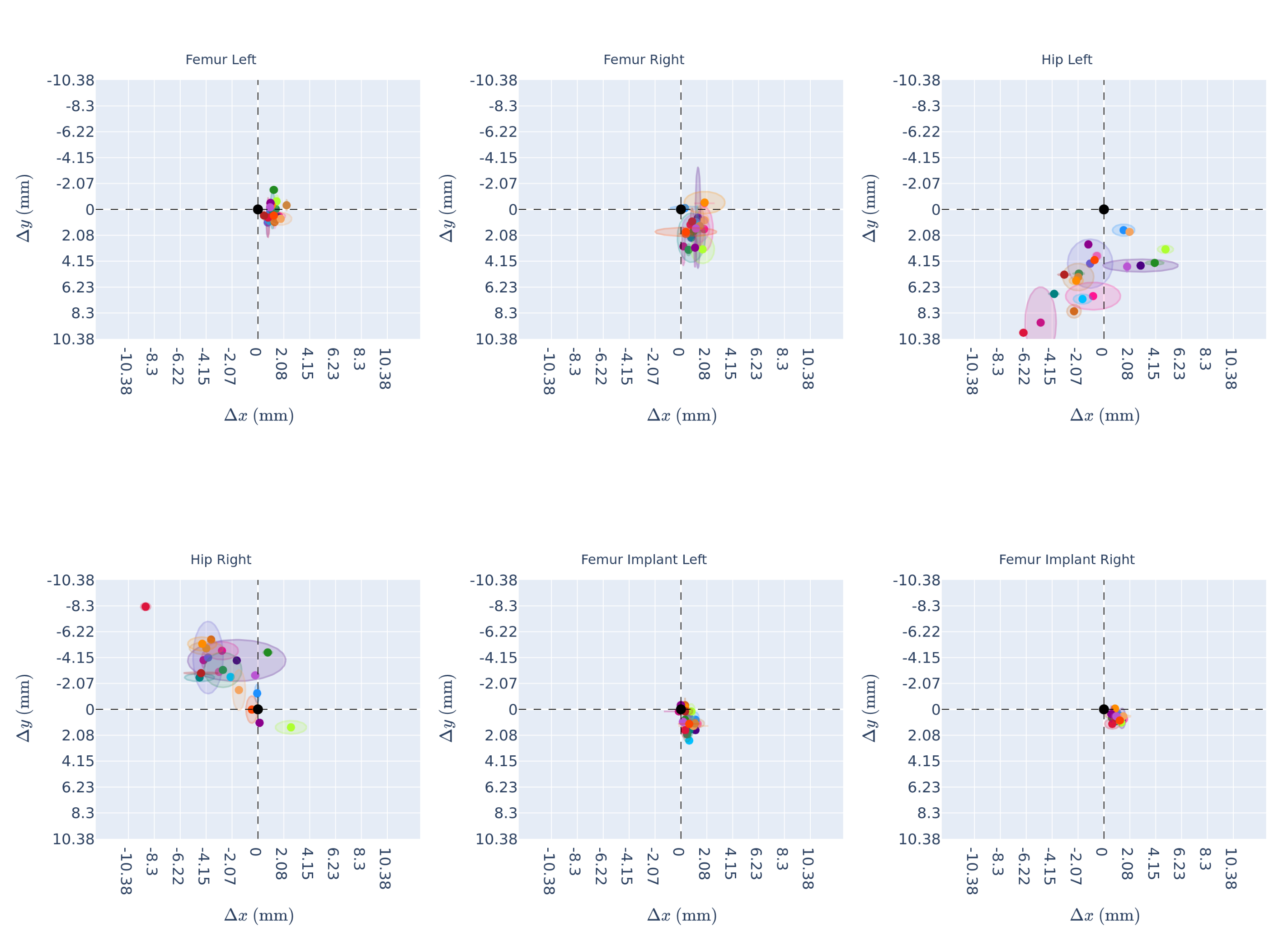}
      \caption{Hip}
    \end{subfigure} 
    \caption{Spatial distribution of mean $\Delta x$ and $\Delta y$ per annotator per class label. The same-colored (more transparent) ellipse represent each annotator's intra-rater consistency ($\Delta x$, $\Delta y$).}
    \label{fig:ablation_human_points_scatter}
\end{figure*}

\begin{figure*}[h]
    \centering
    \centering
    \begin{subfigure}[t]{0.48\textwidth}
        \includegraphics[width=1\linewidth]{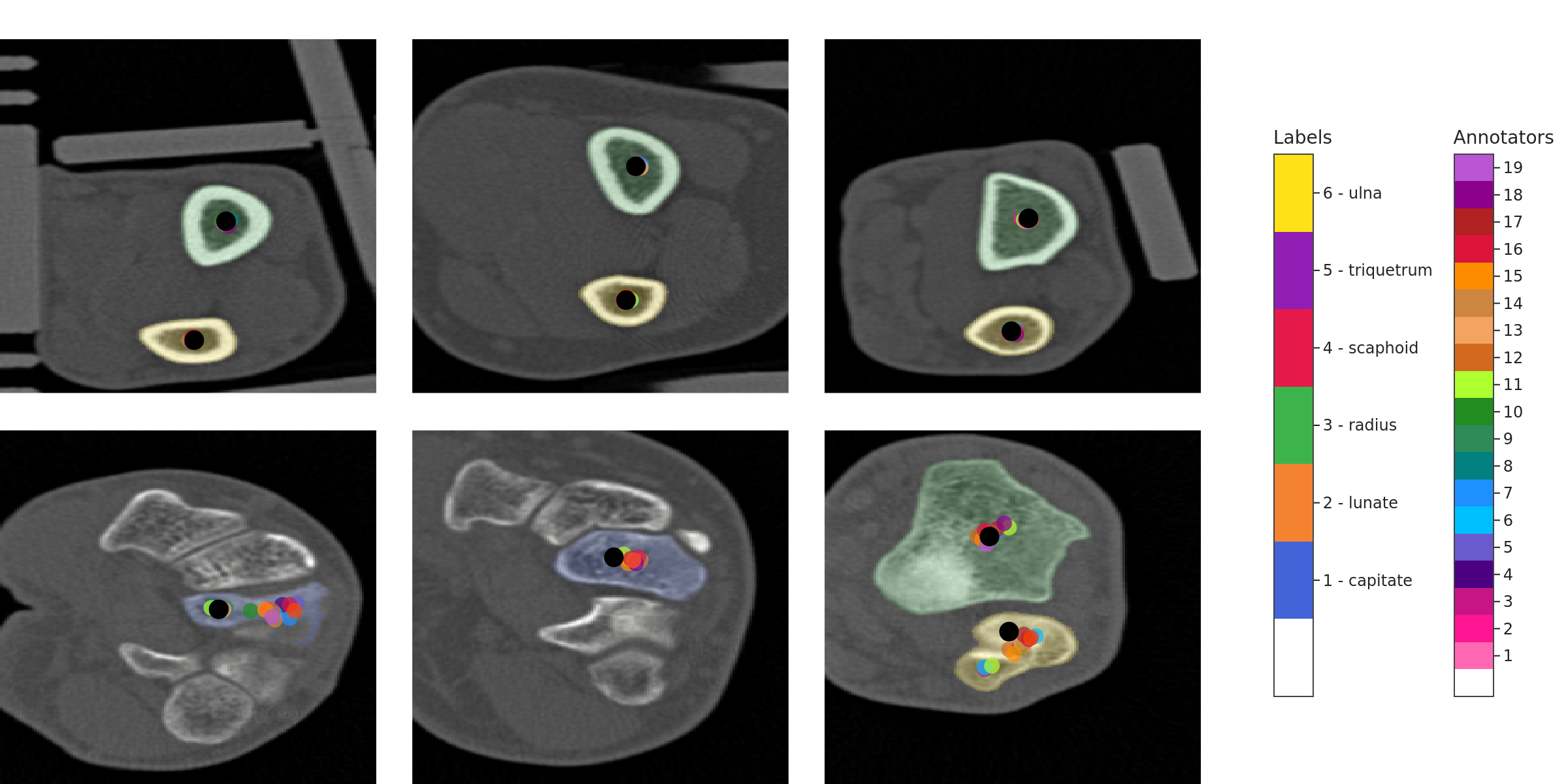}
        \subcaption{Wrist}
    \end{subfigure}\hfill
    \begin{subfigure}[t]{0.48\textwidth}
        \includegraphics[width=1\linewidth]{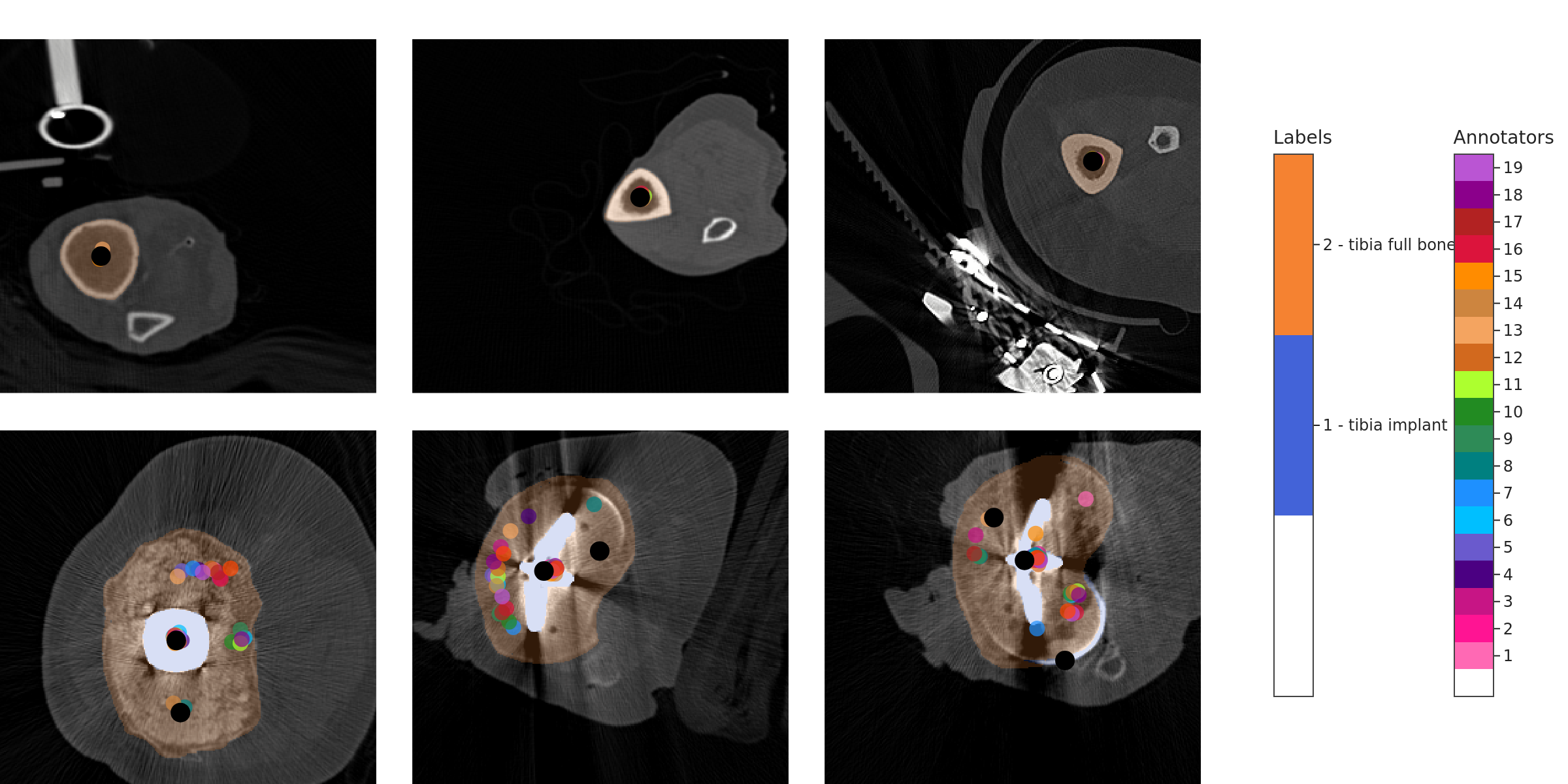}
        \subcaption{Lower Leg}
    \end{subfigure}
    \begin{subfigure}[t]{0.98\textwidth}
        \includegraphics[width=1\linewidth]{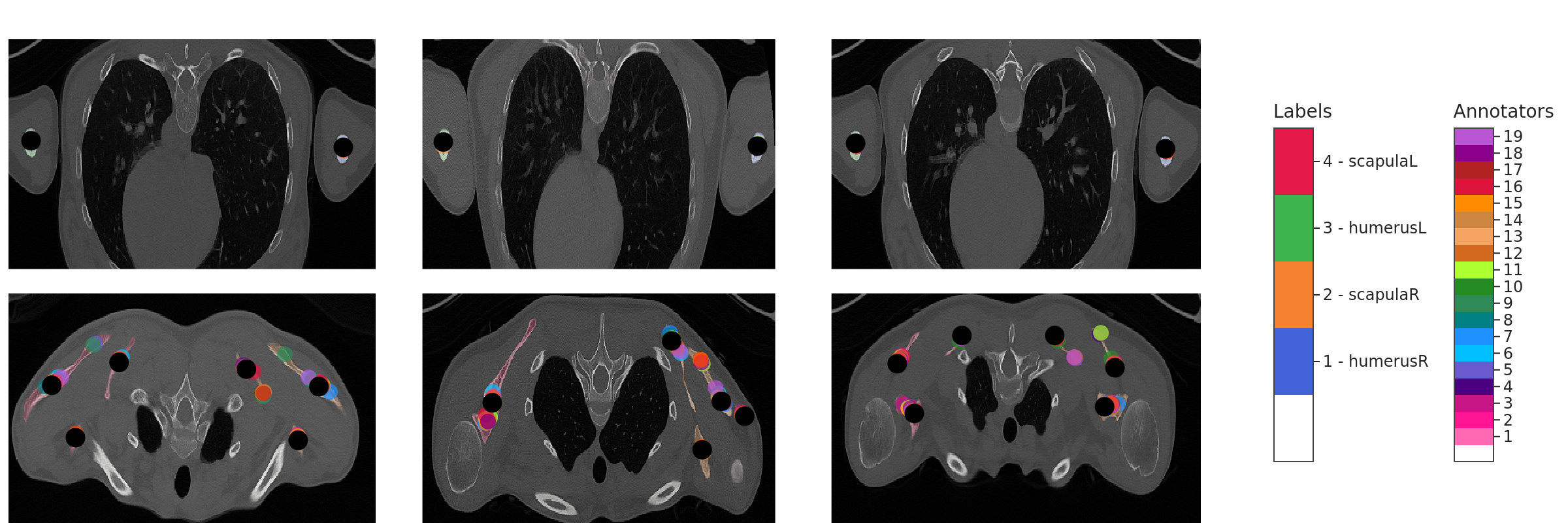}
        \subcaption{Shoulder}
    \end{subfigure}
    \begin{subfigure}[t]{0.98\textwidth}
        \includegraphics[width=1\linewidth]{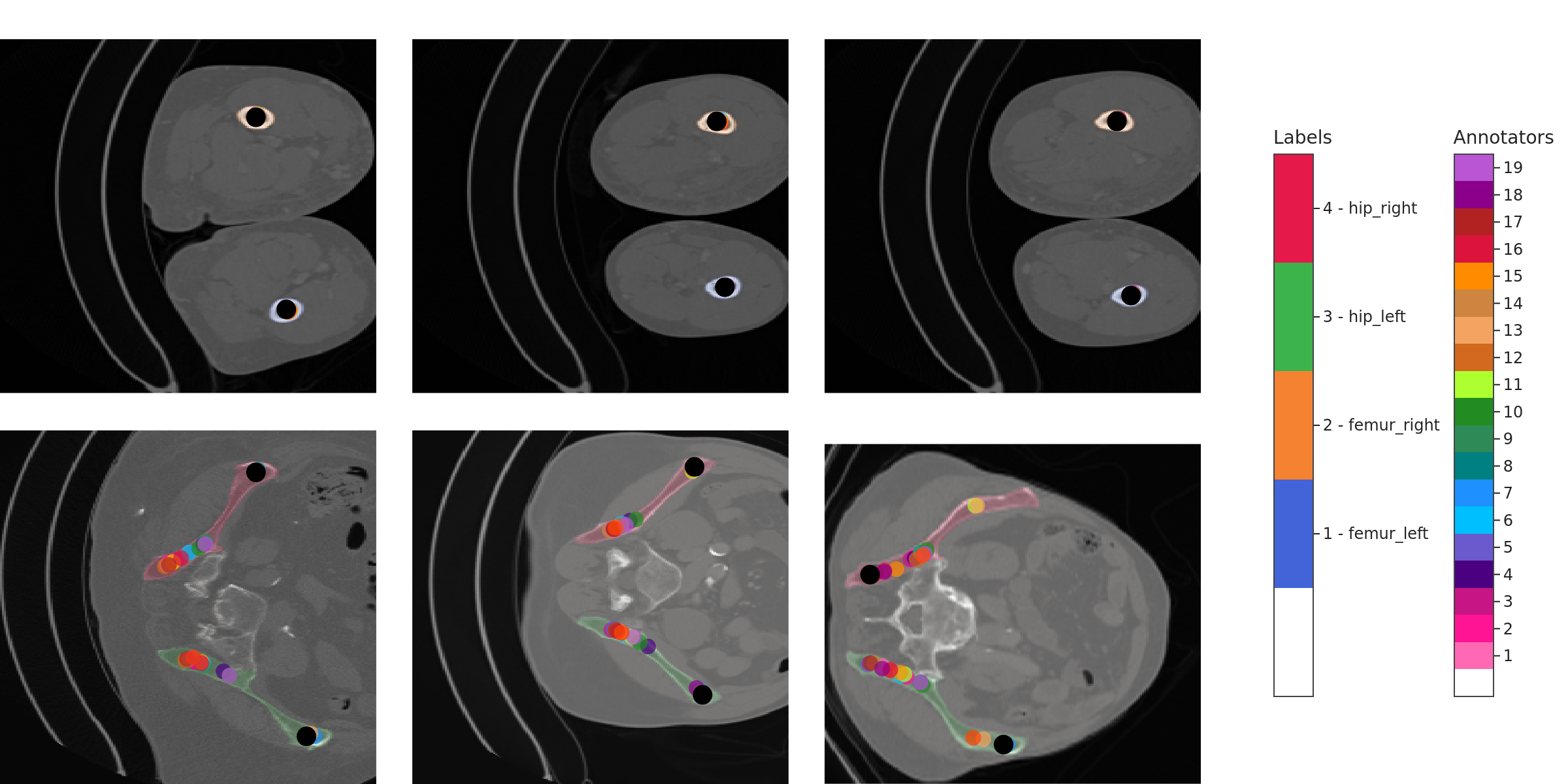}
        \subcaption{Hip}
    \end{subfigure}
    \caption{Examples for center point annotations: Center points with low euclidean distance (mm) (top row) and high values (bottom row) per data subset. Black dots are automatically extracted reference annotation, annotators' annotations are color-encoded.}
    \label{fig:ablation_prompt_eval_point_examples}
\end{figure*}

\newpage
\clearpage

\paragraph{Bounding Box}
Table \ref{tab:ablation_human_box_performance} collects detailed results on the median IoU (\%). Figure \ref{fig:ablation_human_box_scatter} visualizes the spatial distribution of the bounding boxes' center point deviations ($\Delta x$, $\Delta y$) and the intra-rater consistency ($\Delta w$, $\Delta h$) per class labels.

\begin{table*}[h]
\centering

\caption{IoU (\%) of human bounding boxes compared to referenc bounding boxes measured as median and IQR.}
\label{tab:ablation_human_box_performance}

\begin{subtable}{\textwidth}
    \centering
    \caption{Total Average \& Dataset Lower Leg and corresponding class labels}
    \resizebox{0.6\textwidth}{!}{
    \begin{tabular}{lccccc}
    \hline
    Annotator & Total & Lower Leg & Tibia Implant & Tibia bone \\
    \hline
    all & 90.56 (83.4-94.5) & 90.02 (80.3-80.3) & 84.82 (79.1-79.1) & 92.92 (89.5-89.5) \\
    annotator01 & 91.22 (85.5-85.5) & 90.10 (86.5-86.5) & 84.82 (79.1-79.1) & 92.92 (89.5-89.5) \\
    annotator02 & 91.22 (85.9-85.9) & 91.68 (81.2-81.2) & 75.00 (68.5-68.5) & 93.33 (91.8-91.8) \\
    annotator03 & 84.27 (73.8-73.8) & 79.96 (61.2-61.2) & 52.91 (45.4-45.4) & 84.06 (79.6-79.6) \\
    annotator04 & 93.32 (89.0-89.0) & 93.66 (90.2-90.2) & 89.29 (80.9-80.9) & 94.83 (91.9-91.9) \\
    annotator05 & 86.99 (79.1-79.1) & 84.33 (77.0-77.0) & 72.63 (56.3-56.3) & 87.80 (83.5-83.5) \\
    annotator06 & 92.62 (86.8-86.8) & 93.48 (88.6-88.6) & 86.89 (78.8-78.8) & 95.29 (92.8-92.8) \\
    annotator07 & 92.87 (87.2-87.2) & 91.12 (70.7-70.7) & 63.89 (49.8-49.8) & 93.86 (91.4-91.4) \\
    annotator08 & 93.15 (88.5-88.5) & 93.33 (88.7-88.7) & 84.67 (75.5-75.5) & 95.64 (92.8-92.8) \\
    annotator09 & 90.69 (85.0-85.0) & 92.03 (81.5-81.5) & 76.38 (66.4-66.4) & 94.60 (92.2-92.2) \\
    annotator10 & 93.29 (88.2-88.2) & 92.93 (87.2-87.2) & 81.59 (70.0-70.0) & 94.70 (91.7-91.7) \\
    annotator11 & 85.43 (76.0-76.0) & 82.85 (70.2-70.2) & 64.10 (50.4-50.4) & 88.11 (82.5-82.5) \\
    annotator12 & 88.42 (80.1-80.1) & 84.96 (78.3-78.3) & 69.27 (56.5-56.5) & 88.89 (84.1-84.1) \\
    annotator13 & 90.32 (83.9-83.9) & 89.41 (83.5-83.5) & 77.66 (68.4-68.4) & 93.04 (88.5-88.5) \\
    annotator14 & 90.75 (85.1-85.1) & 91.66 (82.5-82.5) & 74.05 (68.3-68.3) & 93.79 (91.8-91.8) \\
    annotator15 & 92.22 (87.2-87.2) & 91.88 (84.7-84.7) & 79.62 (74.5-74.5) & 93.87 (91.5-91.5) \\
    annotator16 & 79.13 (70.2-70.2) & 75.81 (65.1-65.1) & 55.88 (48.0-48.0) & 82.70 (75.5-75.5) \\
    annotator17 & 91.67 (85.3-85.3) & 90.69 (83.8-83.8) & 81.59 (72.3-72.3) & 93.48 (90.6-90.6) \\
    annotator18 & 90.61 (84.6-84.6) & 89.11 (80.7-80.7) & 79.69 (66.1-66.1) & 92.30 (89.0-89.0) \\
    annotator19 & 91.54 (86.7-86.7) & 90.61 (84.3-84.3) & 80.00 (72.4-72.4) & 93.81 (89.8-89.8) \\
    annotator20 & 91.38 (85.4-85.4) & 91.55 (80.8-80.8) & 72.59 (67.7-67.7) & 94.65 (91.6-91.6) \\
    \hline
    \end{tabular}
    }
\end{subtable}

\medskip

\begin{subtable}{\textwidth}
    \centering
    \caption{Dataset Shoulder and corresponding class labels}
    \resizebox{0.6\textwidth}{!}{
    \begin{tabular}{lcccccc}
    \hline
    Annotator & Shoulder & Humerus R & Scapula R & Humerus L & Scapula L \\
    \hline
    all & 87.82 (80.5-80.5) & 84.63 (81.0-81.0) & 93.14 (89.4-89.4) & 84.44 (80.5-80.5) & 92.27 (89.7-89.7) \\
    annotator01 & 88.74 (82.5-82.5) & 84.63 (81.0-81.0) & 93.14 (89.4-89.4) & 84.44 (80.5-80.5) & 92.27 (89.7-89.7) \\
    annotator02 & 90.62 (85.4-85.4) & 87.53 (83.4-83.4) & 92.80 (89.1-89.1) & 87.00 (83.6-83.6) & 92.64 (88.5-88.5) \\
    annotator03 & 75.09 (65.1-65.1) & 67.35 (62.4-62.4) & 86.25 (79.1-79.1) & 68.78 (62.0-62.0) & 80.45 (73.4-73.4) \\
    annotator04 & 91.13 (87.1-87.1) & 89.32 (84.9-84.9) & 92.58 (90.3-90.3) & 89.24 (86.0-86.0) & 91.47 (89.2-89.2) \\
    annotator05 & 80.64 (71.1-71.1) & 71.50 (66.9-66.9) & 85.92 (82.5-82.5) & 72.87 (67.3-67.3) & 85.68 (79.7-79.7) \\
    annotator06 & 89.29 (82.8-82.8) & 84.15 (77.7-77.7) & 93.30 (91.0-91.0) & 84.18 (79.9-79.9) & 91.21 (86.9-86.9) \\
    annotator07 & 91.27 (86.1-86.1) & 88.10 (83.3-83.3) & 93.94 (91.3-91.3) & 89.26 (84.7-84.7) & 92.59 (89.4-89.4) \\
    annotator08 & 92.61 (88.0-88.0) & 89.61 (85.7-85.7) & 94.15 (90.4-90.4) & 92.05 (87.0-87.0) & 94.16 (90.4-90.4) \\
    annotator09 & 89.38 (84.0-84.0) & 86.57 (80.7-80.7) & 93.11 (89.0-89.0) & 85.42 (80.0-80.0) & 91.04 (87.0-87.0) \\
    annotator10 & 92.00 (86.2-86.2) & 89.51 (83.9-83.9) & 93.96 (91.3-91.3) & 87.87 (83.8-83.8) & 93.78 (91.0-91.0) \\
    annotator11 & 78.22 (70.5-70.5) & 72.47 (67.4-67.4) & 82.97 (77.1-77.1) & 73.86 (69.5-69.5) & 81.83 (75.2-75.2) \\
    annotator12 & 80.95 (72.6-72.6) & 75.11 (69.8-69.8) & 86.13 (80.2-80.2) & 77.74 (71.5-71.5) & 83.58 (77.5-77.5) \\
    annotator13 & 86.67 (80.7-80.7) & 81.92 (76.7-76.7) & 90.38 (86.0-86.0) & 83.61 (78.8-78.8) & 90.13 (87.3-87.3) \\
    annotator14 & 88.70 (83.5-83.5) & 85.95 (81.6-81.6) & 90.95 (87.4-87.4) & 85.93 (79.6-79.6) & 90.79 (87.6-87.6) \\
    annotator15 & 92.80 (88.0-88.0) & 90.21 (85.5-85.5) & 95.16 (92.4-92.4) & 91.00 (85.5-85.5) & 93.26 (90.2-90.2) \\
    annotator16 & 73.83 (66.2-66.2) & 72.41 (66.0-66.0) & 75.25 (68.4-68.4) & 71.87 (63.4-63.4) & 75.02 (67.1-67.1) \\
    annotator17 & 87.25 (80.4-80.4) & 81.88 (77.6-77.6) & 90.26 (87.4-87.4) & 82.92 (76.1-76.1) & 90.24 (85.8-85.8) \\
    annotator18 & 88.21 (81.5-81.5) & 84.33 (79.6-79.6) & 91.81 (88.4-88.4) & 82.84 (77.6-77.6) & 90.00 (85.9-85.9) \\
    annotator19 & 89.29 (83.7-83.7) & 87.85 (81.8-81.8) & 91.36 (87.8-87.8) & 88.04 (82.6-82.6) & 89.84 (86.5-86.5) \\
    annotator20 & 90.18 (85.2-85.2) & 86.46 (83.0-83.0) & 93.44 (89.2-89.2) & 89.54 (84.2-84.2) & 92.17 (86.4-86.4) \\
    \hline
    \end{tabular}
    }
\end{subtable}
\end{table*}

\newpage

\begin{table*}[h]\ContinuedFloat
\begin{subtable}{\textwidth}
    \centering
    \caption{Dataset Wrist and corresponding class labels}
    \resizebox{0.95\textwidth}{!}{
    \begin{tabular}{lcccccccc}
    \hline
    Annotator & Wrist & Capitate & Lunate & Radius & Scaphoid & Triquetrum & Ulna \\
    \hline
    all & 92.21 (88.1-88.1) & 92.80 (89.2-89.2) & 94.09 (91.1-91.1) & 94.61 (90.4-90.4) & 94.67 (92.1-92.1) & 93.00 (89.9-89.9) & 91.80 (88.7-88.7) \\
    annotator01 & 93.77 (89.9-89.9) & 92.80 (89.2-89.2) & 94.09 (91.1-91.1) & 94.61 (90.4-90.4) & 94.67 (92.1-92.1) & 93.00 (89.9-89.9) & 91.80 (88.7-88.7) \\
    annotator02 & 91.17 (87.8-87.8) & 91.04 (87.6-87.6) & 91.07 (86.4-86.4) & 92.35 (88.7-88.7) & 91.58 (89.3-89.3) & 89.56 (87.6-87.6) & 90.31 (88.3-88.3) \\
    annotator03 & 90.64 (85.6-85.6) & 91.49 (89.0-89.0) & 89.29 (84.9-84.9) & 91.80 (85.7-85.7) & 92.08 (87.5-87.5) & 88.67 (82.0-82.0) & 87.09 (83.4-83.4) \\
    annotator04 & 95.03 (92.6-92.6) & 94.72 (92.6-92.6) & 95.59 (91.1-91.1) & 95.99 (92.3-92.3) & 95.66 (93.5-93.5) & 94.59 (92.3-92.3) & 95.00 (92.6-92.6) \\
    annotator05 & 89.20 (85.8-85.8) & 89.30 (85.5-85.5) & 89.30 (85.2-85.2) & 90.57 (86.7-86.7) & 89.47 (86.8-86.8) & 88.46 (84.8-84.8) & 87.96 (83.9-83.9) \\
    annotator06 & 94.30 (91.2-91.2) & 94.52 (90.7-90.7) & 94.58 (90.5-90.5) & 94.87 (92.2-92.2) & 94.29 (92.3-92.3) & 92.88 (90.6-90.6) & 94.23 (91.0-91.0) \\
    annotator07 & 94.60 (91.8-91.8) & 95.28 (93.8-93.8) & 94.72 (91.5-91.5) & 94.10 (90.6-90.6) & 94.88 (93.2-93.2) & 94.44 (91.1-91.1) & 93.44 (89.6-89.6) \\
    annotator08 & 94.59 (91.8-91.8) & 95.25 (92.5-92.5) & 94.58 (91.2-91.2) & 95.24 (93.2-93.2) & 94.69 (91.5-91.5) & 94.14 (90.5-90.5) & 94.37 (92.0-92.0) \\
    annotator09 & 91.20 (87.1-87.1) & 91.07 (87.4-87.4) & 91.44 (86.9-86.9) & 91.33 (87.8-87.8) & 91.53 (88.1-88.1) & 90.84 (86.8-86.8) & 90.47 (86.2-86.2) \\
    annotator10 & 94.59 (91.3-91.3) & 94.29 (91.9-91.9) & 94.62 (90.5-90.5) & 95.11 (91.7-91.7) & 95.18 (92.6-92.6) & 94.08 (91.3-91.3) & 93.01 (89.2-89.2) \\
    annotator11 & 88.35 (83.3-83.3) & 87.56 (83.9-83.9) & 90.32 (85.6-85.6) & 86.98 (80.7-80.7) & 89.45 (85.8-85.8) & 87.77 (83.9-83.9) & 83.17 (79.3-79.3) \\
    annotator12 & 92.01 (88.9-88.9) & 92.01 (89.6-89.6) & 92.24 (89.0-89.0) & 92.86 (86.5-86.5) & 92.63 (89.6-89.6) & 91.54 (88.3-88.3) & 89.80 (86.2-86.2) \\
    annotator13 & 91.41 (88.8-88.8) & 91.15 (88.9-88.9) & 91.78 (90.6-90.6) & 92.82 (90.1-90.1) & 91.42 (90.0-90.0) & 90.24 (87.6-87.6) & 90.53 (85.3-85.3) \\
    annotator14 & 91.23 (87.8-87.8) & 90.50 (88.0-88.0) & 90.97 (88.0-88.0) & 93.24 (90.9-90.9) & 91.25 (89.0-89.0) & 89.03 (83.9-83.9) & 92.11 (88.5-88.5) \\
    annotator15 & 92.96 (90.0-90.0) & 92.72 (89.7-89.7) & 94.12 (89.2-89.2) & 94.11 (90.5-90.5) & 94.01 (92.0-92.0) & 91.93 (89.5-89.5) & 92.08 (88.3-88.3) \\
    annotator16 & 80.70 (75.7-75.7) & 81.46 (77.1-77.1) & 80.90 (76.5-76.5) & 79.76 (74.9-74.9) & 82.49 (77.6-77.6) & 79.99 (75.1-75.1) & 76.81 (72.7-72.7) \\
    annotator17 & 93.54 (90.7-90.7) & 93.75 (90.7-90.7) & 92.92 (90.6-90.6) & 94.49 (91.5-91.5) & 93.47 (91.8-91.8) & 92.67 (89.9-89.9) & 93.41 (90.6-90.6) \\
    annotator18 & 92.22 (88.7-88.7) & 91.02 (88.4-88.4) & 93.06 (89.2-89.2) & 92.87 (88.6-88.6) & 92.41 (90.3-90.3) & 91.67 (88.6-88.6) & 92.26 (86.3-86.3) \\
    annotator19 & 93.14 (90.7-90.7) & 92.63 (90.4-90.4) & 93.72 (91.0-91.0) & 93.96 (90.3-90.3) & 93.73 (92.0-92.0) & 93.12 (90.6-90.6) & 91.97 (89.4-89.4) \\
    annotator20 & 92.42 (89.0-89.0) & 92.88 (90.9-90.9) & 92.11 (85.2-85.2) & 94.83 (90.7-90.7) & 92.42 (90.6-90.6) & 91.68 (86.5-86.5) & 91.14 (86.0-86.0) \\
    \hline
    \end{tabular}    
    }
\end{subtable}

\begin{subtable}{\textwidth}
    \centering
    \caption{Dataset Hip and corresponding class labels}
    \resizebox{0.95\textwidth}{!}{
    \begin{tabular}{lcccccccc}
    \hline
    Annotator & Hip & Femur L & Femur R & Hip L & Hip R & Femur Implant L & Femur Implant R \\
    \hline
    all & 90.69 (82.1-82.1) & 91.59 (88.7-88.7) & 90.81 (88.7-88.7) & 90.91 (87.6-87.6) & 91.66 (87.4-87.4) & 67.11 (60.2-60.2) & 65.98 (55.8-55.8) \\
    annotator01 & 90.19 (81.0-81.0) & 91.59 (88.7-88.7) & 90.81 (88.7-88.7) & 90.91 (87.6-87.6) & 91.66 (87.4-87.4) & 67.11 (60.2-60.2) & 65.98 (55.8-55.8) \\
    annotator02 & 91.34 (83.8-83.8) & 91.46 (87.9-87.9) & 91.04 (85.0-85.0) & 93.29 (87.7-87.7) & 93.27 (86.7-86.7) & 75.00 (70.6-70.6) & 77.67 (70.4-70.4) \\
    annotator03 & 85.58 (74.7-74.7) & 85.71 (81.0-81.0) & 84.08 (80.3-80.3) & 89.74 (84.8-84.8) & 90.03 (83.3-83.3) & 60.71 (52.4-52.4) & 55.84 (49.1-49.1) \\
    annotator04 & 92.00 (85.0-85.0) & 91.58 (88.4-88.4) & 90.84 (86.3-86.3) & 95.00 (89.3-89.3) & 93.75 (89.2-89.2) & 83.22 (76.7-76.7) & 83.33 (77.4-77.4) \\
    annotator05 & 89.51 (81.1-81.1) & 89.86 (85.9-85.9) & 88.89 (84.2-84.2) & 91.28 (86.5-86.5) & 91.30 (87.0-87.0) & 65.24 (59.4-59.4) & 65.38 (59.4-59.4) \\
    annotator06 & 92.68 (85.9-85.9) & 91.30 (86.8-86.8) & 92.50 (88.9-88.9) & 95.96 (89.7-89.7) & 93.41 (89.5-89.5) & 74.56 (70.8-70.8) & 77.67 (70.2-70.2) \\
    annotator07 & 92.42 (84.9-84.9) & 94.19 (89.2-89.2) & 92.11 (88.7-88.7) & 93.42 (89.3-89.3) & 92.92 (86.5-86.5) & 74.30 (72.9-72.9) & 70.64 (62.7-62.7) \\
    annotator08 & 90.88 (83.3-83.3) & 90.31 (84.0-84.0) & 90.87 (84.5-84.5) & 92.82 (87.1-87.1) & 92.60 (87.3-87.3) & 77.73 (74.6-74.6) & 75.45 (65.7-65.7) \\
    annotator09 & 90.19 (81.4-81.4) & 91.44 (88.8-88.8) & 90.19 (86.4-86.4) & 91.78 (83.7-83.7) & 90.16 (83.1-83.1) & 83.04 (77.1-77.1) & 76.60 (69.3-69.3) \\
    annotator10 & 92.86 (85.5-85.5) & 93.66 (89.3-89.3) & 93.33 (88.7-88.7) & 94.12 (89.1-89.1) & 94.35 (90.7-90.7) & 77.78 (75.0-75.0) & 77.24 (70.3-70.3) \\
    annotator11 & 89.74 (79.6-79.6) & 90.82 (86.7-86.7) & 90.91 (86.6-86.6) & 91.41 (86.4-86.4) & 90.51 (84.0-84.0) & 62.63 (60.7-60.7) & 60.44 (55.5-55.5) \\
    annotator12 & 90.73 (82.9-82.9) & 91.89 (85.5-85.5) & 90.18 (87.0-87.0) & 91.43 (87.4-87.4) & 92.63 (86.0-86.0) & 75.00 (70.0-70.0) & 69.35 (61.7-61.7) \\
    annotator13 & 91.87 (83.5-83.5) & 91.11 (87.1-87.1) & 91.67 (85.5-85.5) & 92.86 (87.4-87.4) & 93.79 (88.1-88.1) & 75.00 (70.2-70.2) & 76.39 (59.2-59.2) \\
    annotator14 & 91.49 (83.9-83.9) & 92.12 (86.6-86.6) & 91.54 (88.7-88.7) & 93.12 (88.5-88.5) & 94.35 (89.1-89.1) & 77.73 (66.1-66.1) & 77.24 (64.8-64.8) \\
    annotator15 & 90.10 (82.4-82.4) & 89.31 (82.5-82.5) & 89.02 (82.9-82.9) & 90.64 (85.8-85.8) & 92.51 (88.1-88.1) & 81.82 (74.1-74.1) & 79.00 (71.2-71.2) \\
    annotator16 & 82.36 (71.0-71.0) & 86.49 (80.2-80.2) & 82.45 (76.2-76.2) & 85.87 (79.6-79.6) & 82.44 (75.6-75.6) & 62.13 (47.6-47.6) & 53.61 (42.7-42.7) \\
    annotator17 & 92.38 (84.9-84.9) & 92.58 (89.8-89.8) & 91.55 (89.7-89.7) & 94.18 (89.8-89.8) & 94.29 (89.5-89.5) & 77.78 (74.6-74.6) & 77.24 (71.2-71.2) \\
    annotator18 & 91.55 (83.4-83.4) & 94.59 (88.7-88.7) & 90.42 (86.3-86.3) & 93.74 (88.3-88.3) & 93.67 (86.8-86.8) & 81.25 (74.6-74.6) & 75.76 (72.6-72.6) \\
    annotator19 & 91.56 (85.4-85.4) & 91.82 (87.4-87.4) & 90.27 (86.7-86.7) & 93.30 (89.5-89.5) & 93.94 (87.9-87.9) & 81.48 (70.8-70.8) & 75.93 (70.5-70.5) \\
    annotator20 & 90.48 (82.5-82.5) & 90.48 (84.2-84.2) & 90.90 (86.5-86.5) & 92.18 (84.8-84.8) & 92.11 (88.2-88.2) & 77.06 (74.6-74.6) & 76.92 (71.7-71.7) \\
    \hline
    \end{tabular}
    }
\end{subtable}
\end{table*}

\begin{figure*}[h]
    \centering
    \begin{subfigure}[t]{0.9\textwidth}
      \includegraphics[width=1\linewidth]{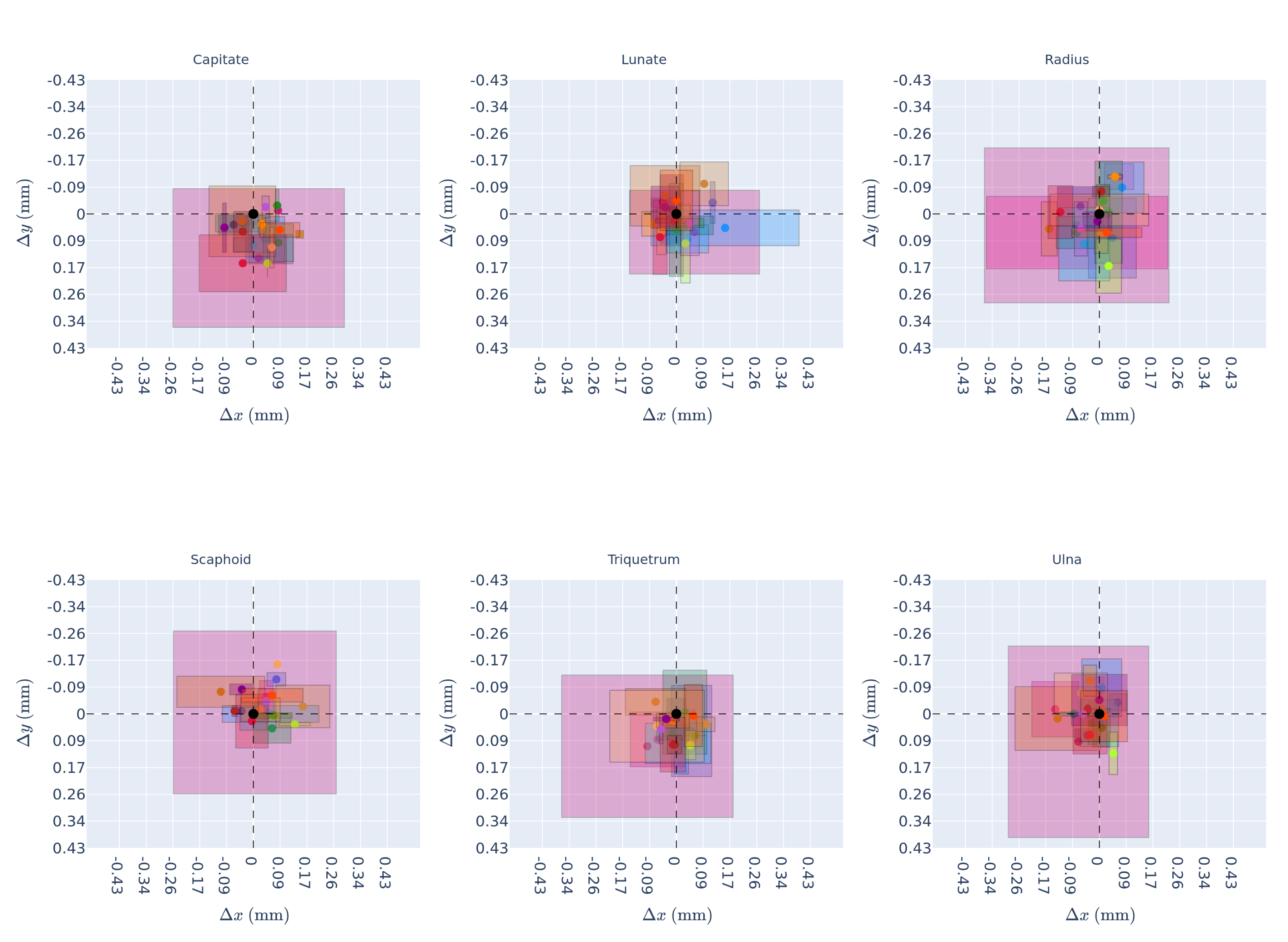}
      \caption{Wrist}
    \end{subfigure} 
    \begin{subfigure}[t]{0.9\textwidth}
      \includegraphics[width=1\linewidth]{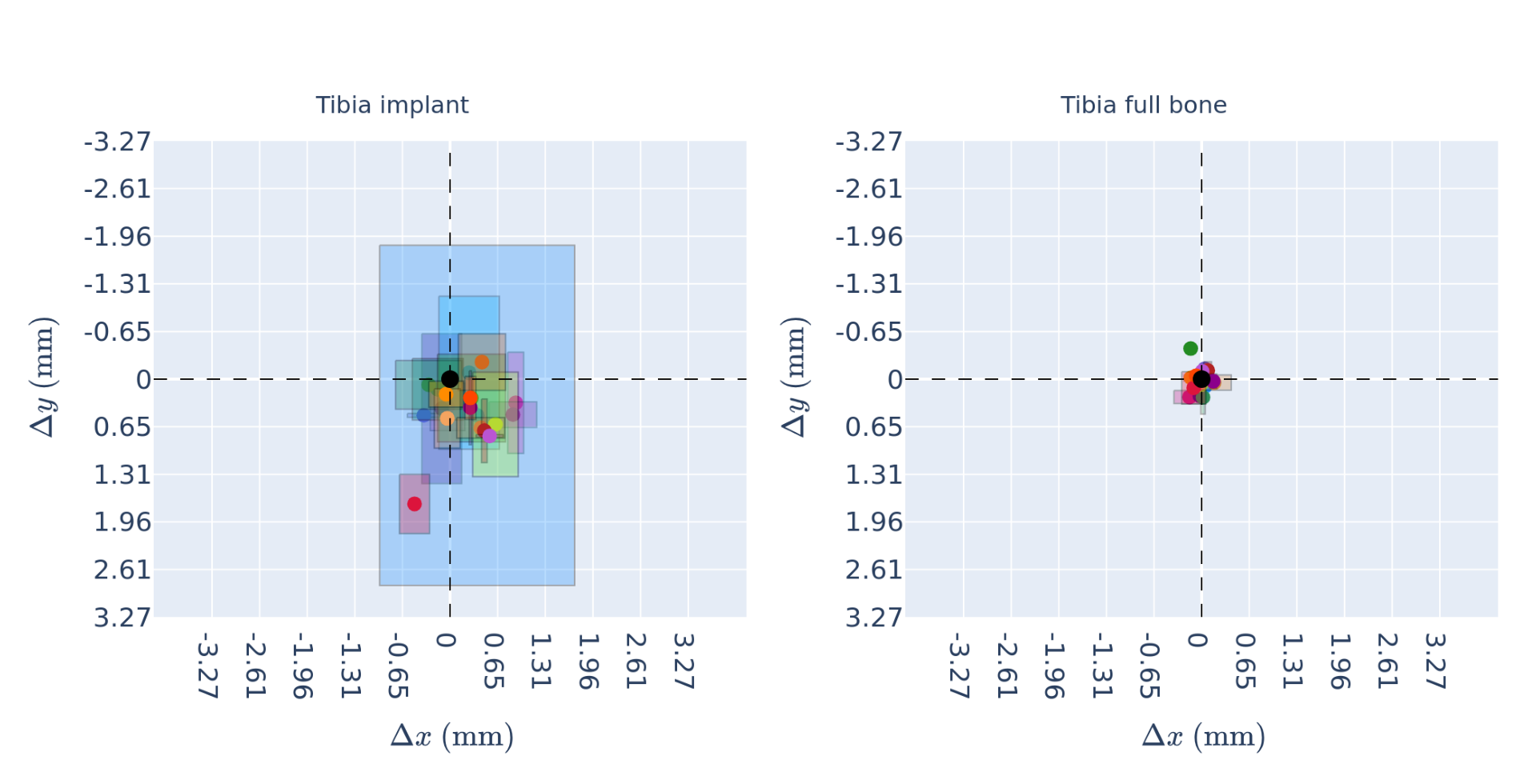}
      \caption{Lower Leg}
     \end{subfigure}
\end{figure*}

\begin{figure*}[h]\ContinuedFloat
    \centering
    \begin{subfigure}[t]{0.9\textwidth}
      \includegraphics[width=1\linewidth]{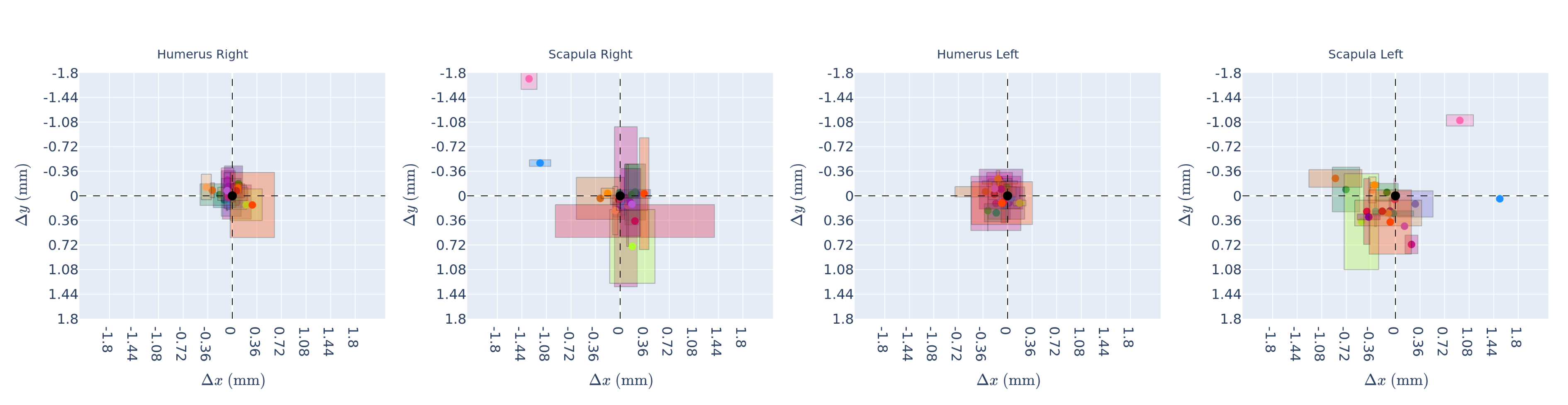}
      \caption{Shoulder}
    \end{subfigure} 
    \begin{subfigure}[t]{0.9\textwidth}
      \includegraphics[width=1\linewidth]{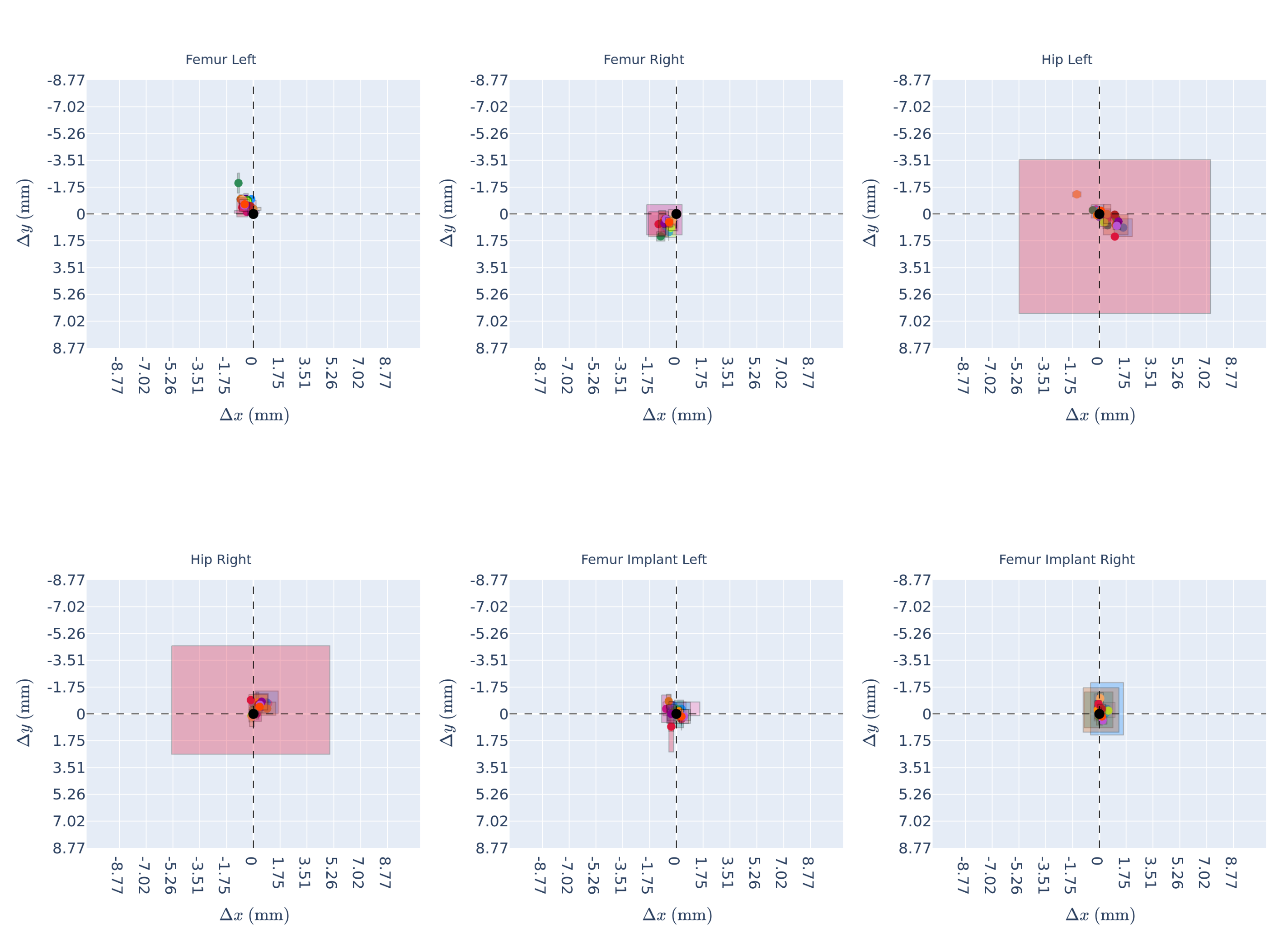}
      \caption{Hip}
    \end{subfigure} 
    \caption{Spatial distribution of the mean $\Delta x$ and $\Delta y$ per annotator per class label. The same-colored (more transparent) rectangle represents each annotator's intra-rater consistency ($\Delta w$, $\Delta h$).}
    \label{fig:ablation_human_box_scatter}
\end{figure*}

\begin{figure*}[h]
    \centering
    \begin{subfigure}[t]{0.48\textwidth}
        \includegraphics[width=1\linewidth]{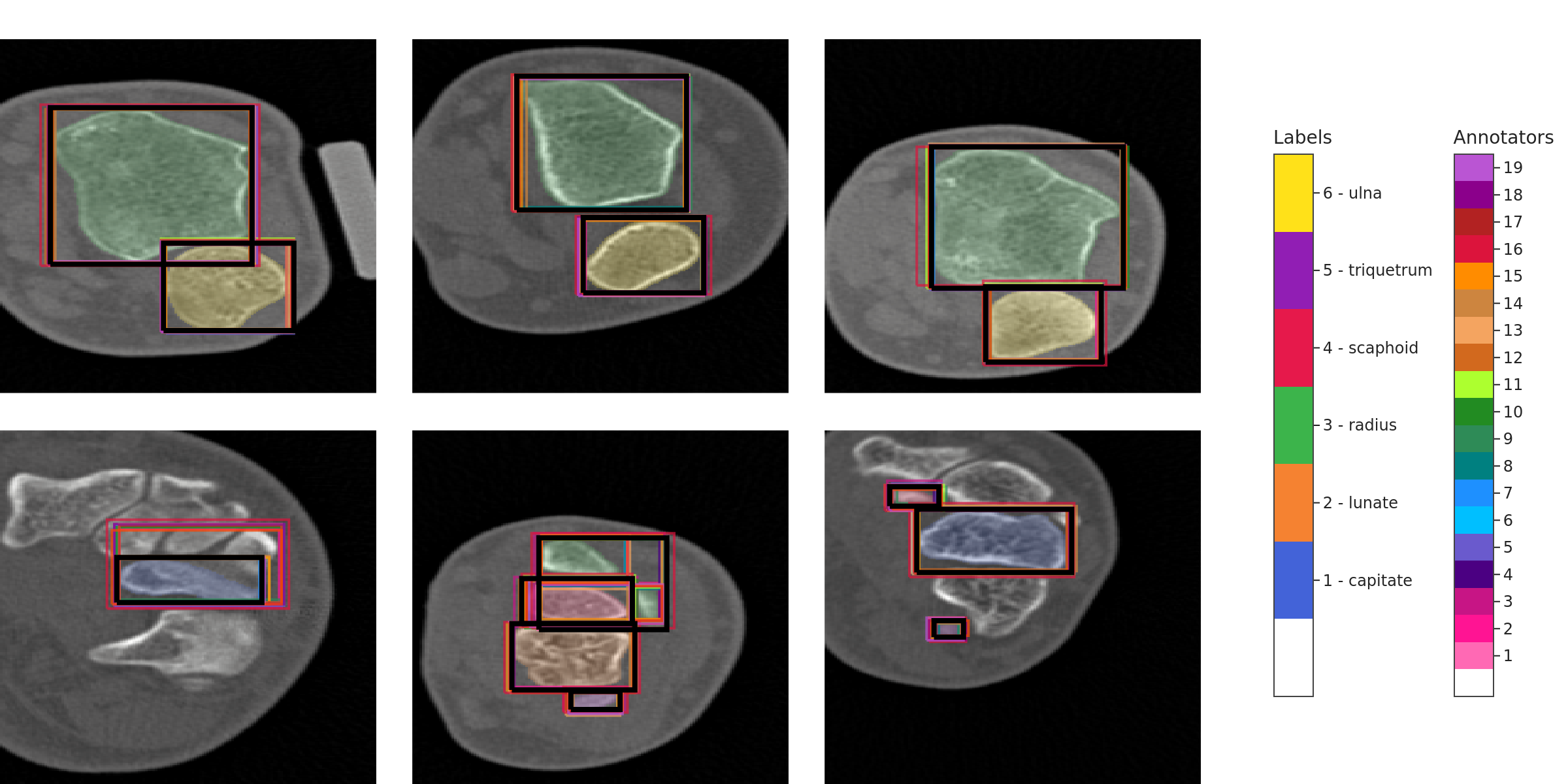}
        \subcaption{Wrist}
    \end{subfigure}\hfill
    \begin{subfigure}[t]{0.48\textwidth}
        \includegraphics[width=1\linewidth]{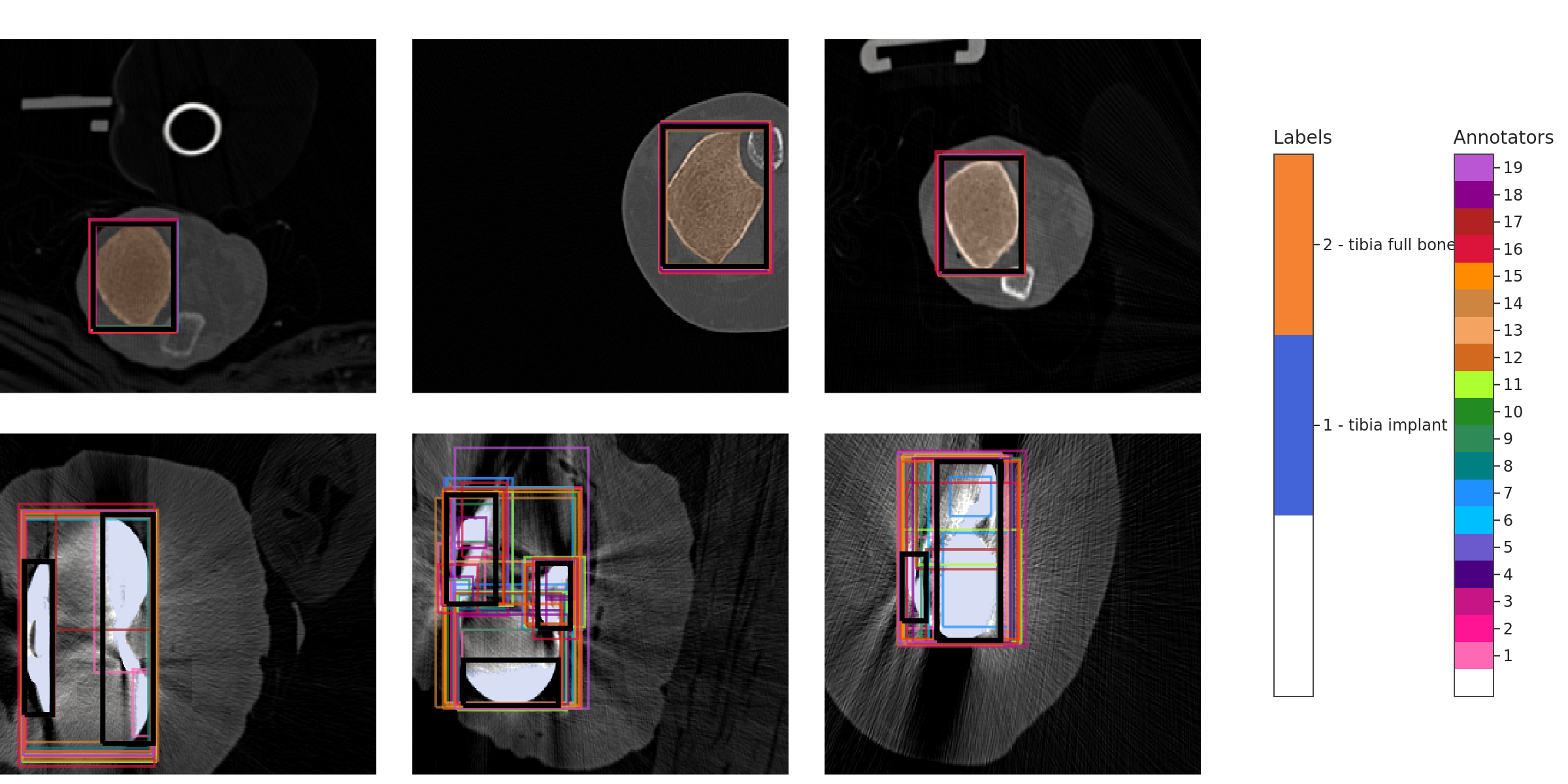}
        \subcaption{Lower Leg}
    \end{subfigure}
    \begin{subfigure}[t]{0.93\textwidth}
        \includegraphics[width=1\linewidth]{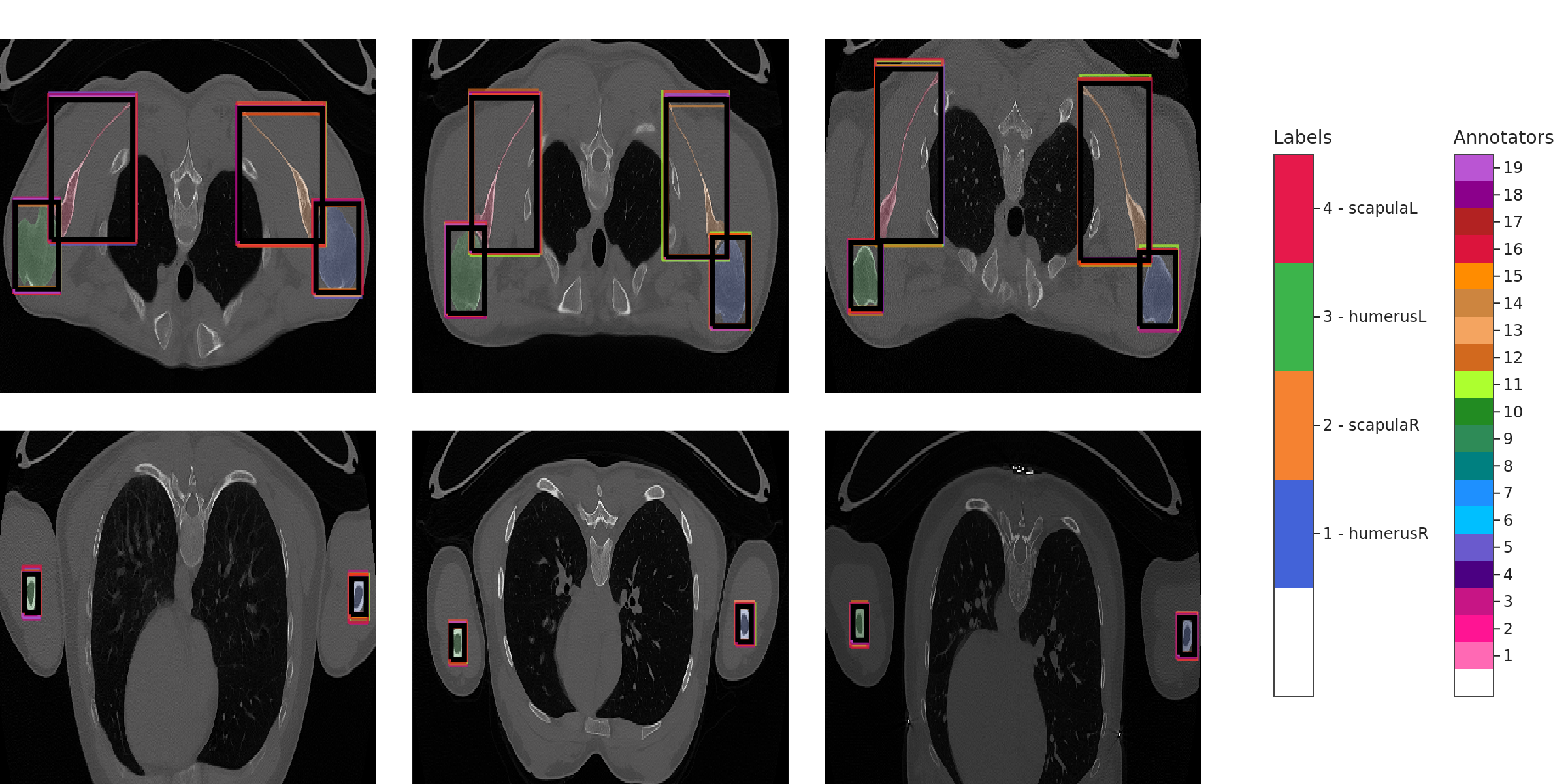}
        \subcaption{Shoulder}
    \end{subfigure}
    \begin{subfigure}[t]{0.93\textwidth}
        \includegraphics[width=1\linewidth]{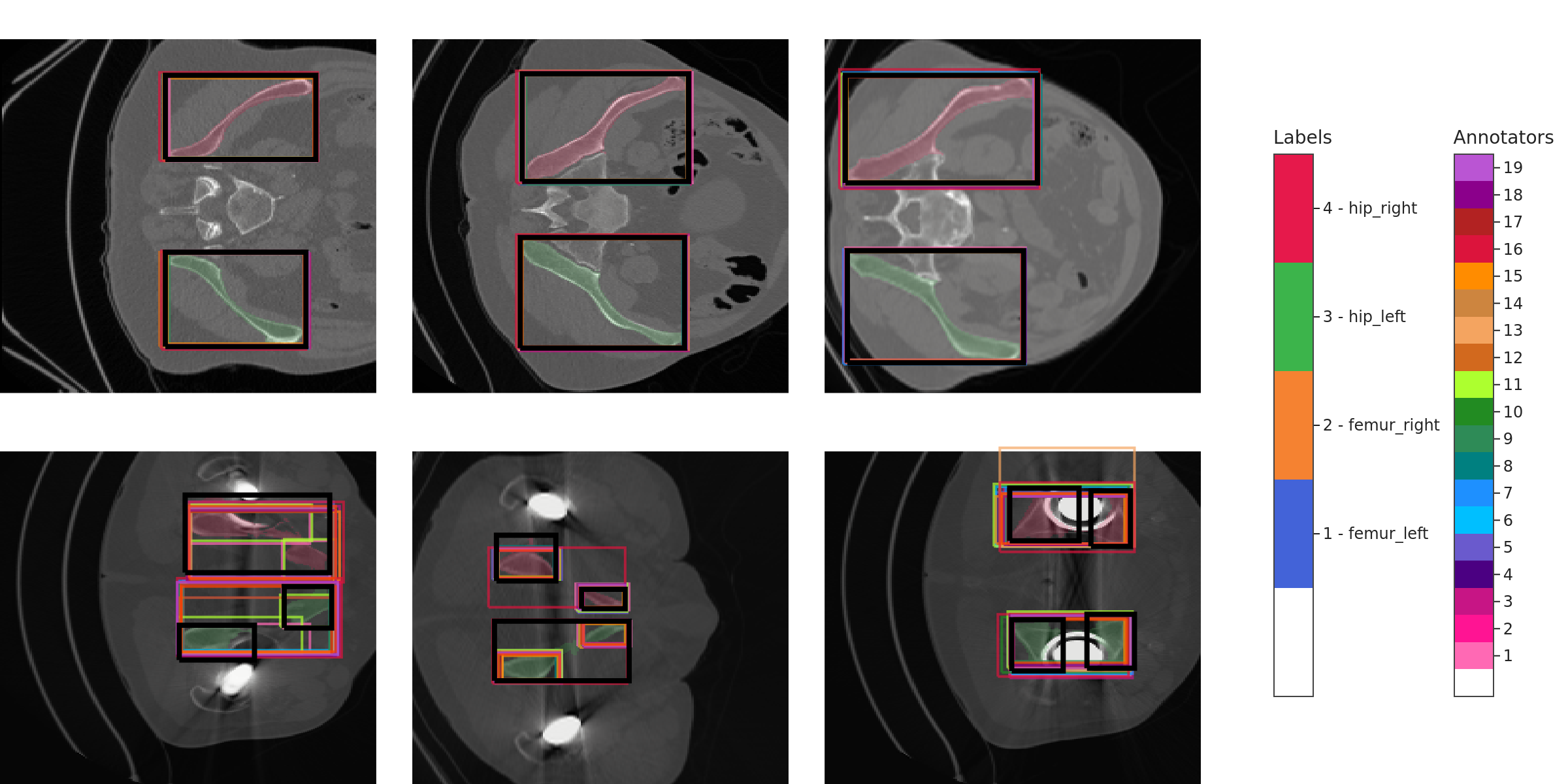}
        \subcaption{Hip}
    \end{subfigure}
    \caption{Examples for bounding box annotations: Boxes with high IoU (\%) (top row) and low values (bottom row) per data subset. Black dots are automatically extracted reference annotation, annotators' annotations are color-encoded.}
    \label{fig:ablation_prompt_eval_boxes_examples}
\end{figure*}

\newpage
\clearpage

\subsection{Inter-rater annotation consistency}

Table \ref{tab:inter_rater_ranking} shows the inter-rater variability ranking, starting with the annotator with the lowest variability to all other annotators. This ranking is used for the iterative search to determine the threshold of model sensitivity to inter-rater variability. The rows highlighted in bold have been tested in the iterative search.

\begin{table*}[h]
    \centering
    \setlength{\tabcolsep}{3pt}
    \renewcommand{\arraystretch}{1.3}
    \caption{Ranking of inter-rater variability, measured by averaged euclidean distance (mm), per annotator, starting with the lowest variability.
    \scriptsize{The euclidean distance (mm) is averaged for all comparisons of one annotator to all other annotators. For the combination prompt, the euclidean distance, averaged from center point and bounding box analysis, is used for the ranking, because it considers both prompts. Annotators highlighted in bold have been used in the iterative search approach.}}
    \label{tab:inter_rater_ranking}
    \begin{subtable}{0.48\linewidth}
    \caption{Center Point}
        \centering
        \begin{tabular}{l:c}
        \hline
        Annotator & Eucl. distance (mm) \\
        \hline
        \textbf{annotator15} & 2.51{\footnotesize±5.5} \\
        annotator02 & 2.54{\footnotesize±5.9} \\
        annotator20 & 2.68{\footnotesize±6.8} \\
        annotator01 & 2.69{\footnotesize±6.8} \\
        annotator14 & 2.70{\footnotesize±6.8} \\
        annotator05 & 2.73{\footnotesize±6.8} \\
        annotator17 & 2.83{\footnotesize±7.2} \\
        annotator18 & 2.88{\footnotesize±6.9} \\
        annotator04 & 2.93{\footnotesize±6.9} \\
        annotator06 & 2.94{\footnotesize±6.9} \\
        annotator08 & 2.95{\footnotesize±7.2} \\
        annotator19 & 2.99{\footnotesize±6.6} \\
        annotator12 & 3.01{\footnotesize±6.7} \\
        annotator03 & 3.05{\footnotesize±7.4} \\
        annotator11 & 3.05{\footnotesize±7.0} \\
        annotator09 & 3.06{\footnotesize±8.2} \\
        annotator16 & 3.12{\footnotesize±7.4} \\
        annotator13 & 3.27{\footnotesize±8.5} \\
        annotator10 & 3.27{\footnotesize±6.9} \\
        annotator07 & 3.63{\footnotesize±7.7} \\
        \hline
        \end{tabular}
    \end{subtable}
    \hfill
    \begin{subtable}{0.48\linewidth}
    \caption{Combination}
    \centering
        \begin{tabular}{l:cc}
        \hline
        Annotator & Eucl. distance (mm) & IoU (\%) \\
        \hline
        \textbf{annotator02} & 1.67{\footnotesize±2.8} & 87.32{\footnotesize±7.9} \\
        annotator15 & 1.68{\footnotesize±2.8} & 88.44{\footnotesize±8.3} \\
        \textbf{annotator05} & 1.85{\footnotesize±3.2} & 87.58{\footnotesize±8.9} \\
        \textbf{annotator14} & 1.86{\footnotesize±3.2} & 87.76{\footnotesize±8.0} \\
        \textbf{annotator20} & 1.86{\footnotesize±3.2} & 87.81{\footnotesize±8.1} \\
        \textbf{annotator01} & 1.87{\footnotesize±3.3} & 89.08{\footnotesize±9.6} \\
        annotator17 & 1.94{\footnotesize±3.4} & 89.07{\footnotesize±8.8} \\
        annotator04 & 1.94{\footnotesize±3.2} & 89.25{\footnotesize±8.2} \\
        annotator06 & 1.94{\footnotesize±3.2} & 89.35{\footnotesize±8.1} \\
        annotator18 & 1.95{\footnotesize±3.2} & 88.64{\footnotesize±9.4} \\
        \textbf{annotator09} & 1.96{\footnotesize±3.1} & 87.95{\footnotesize±8.1} \\
        annotator08 & 2.00{\footnotesize±3.4} & 88.15{\footnotesize±8.8} \\
        annotator19 & 2.00{\footnotesize±3.1} & 88.66{\footnotesize±8.8} \\
        annotator12 & 2.05{\footnotesize±3.2} & 87.67{\footnotesize±7.8} \\
        annotator10 & 2.09{\footnotesize±3.4} & 89.41{\footnotesize±8.6} \\
        annotator11 & 2.09{\footnotesize±3.3} & 86.73{\footnotesize±7.8} \\
        annotator03 & 2.17{\footnotesize±3.4} & 84.13{\footnotesize±9.6} \\
        annotator13 & 2.20{\footnotesize±4.2} & 89.03{\footnotesize±7.6} \\
        annotator16 & 2.24{\footnotesize±3.6} & 81.06{\footnotesize±9.1} \\
        \textbf{annotator07} & 2.40{\footnotesize±3.6} & 87.22{\footnotesize±11.6} \\
        \hline
        \end{tabular}
    \end{subtable}
\end{table*}

\newpage

\section{Segmentation performance with reference prompts} \label{sec:appendix_pareto_front}
\begin{figure*}[h!]
    \centering
        \begin{subfigure}[t]{0.9\textwidth}
          \includegraphics[width=1\linewidth]{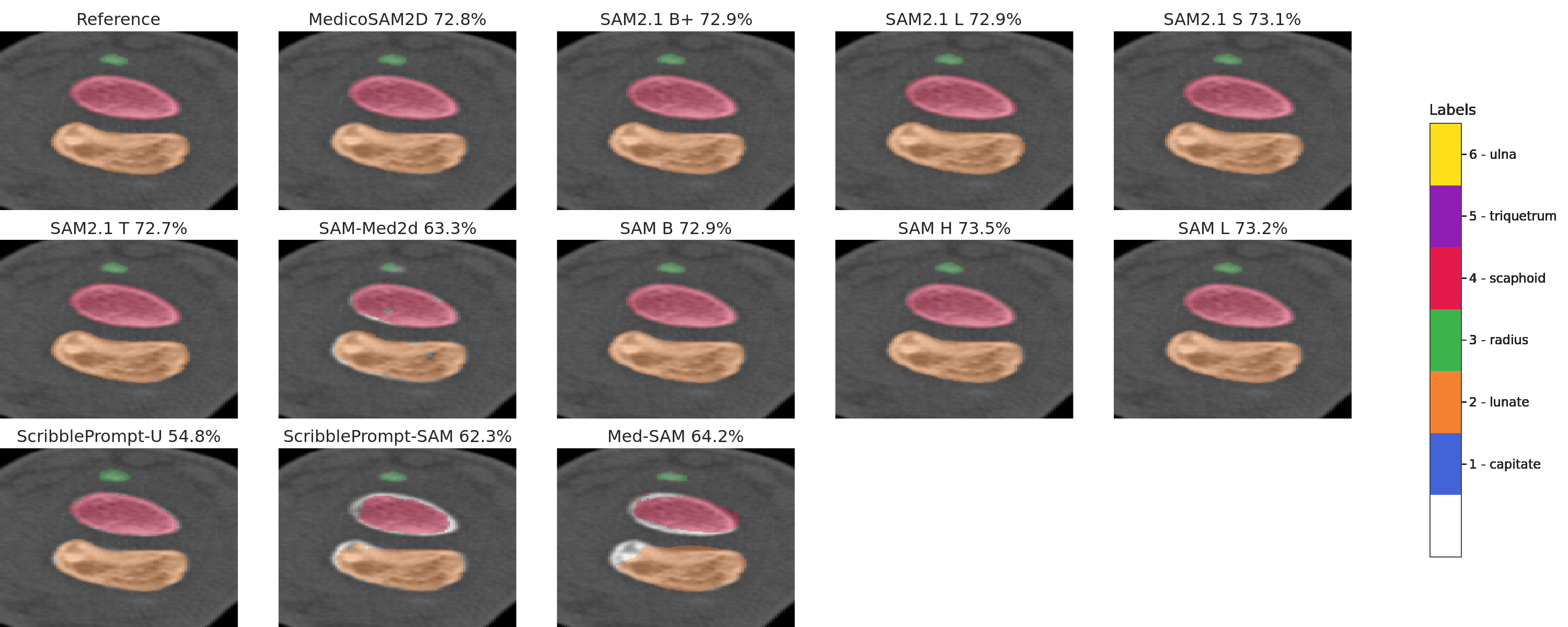}
          \subcaption{Bounding Box}
        \end{subfigure} 
        \begin{subfigure}[t]{0.9\textwidth}
          \includegraphics[width=1\linewidth]{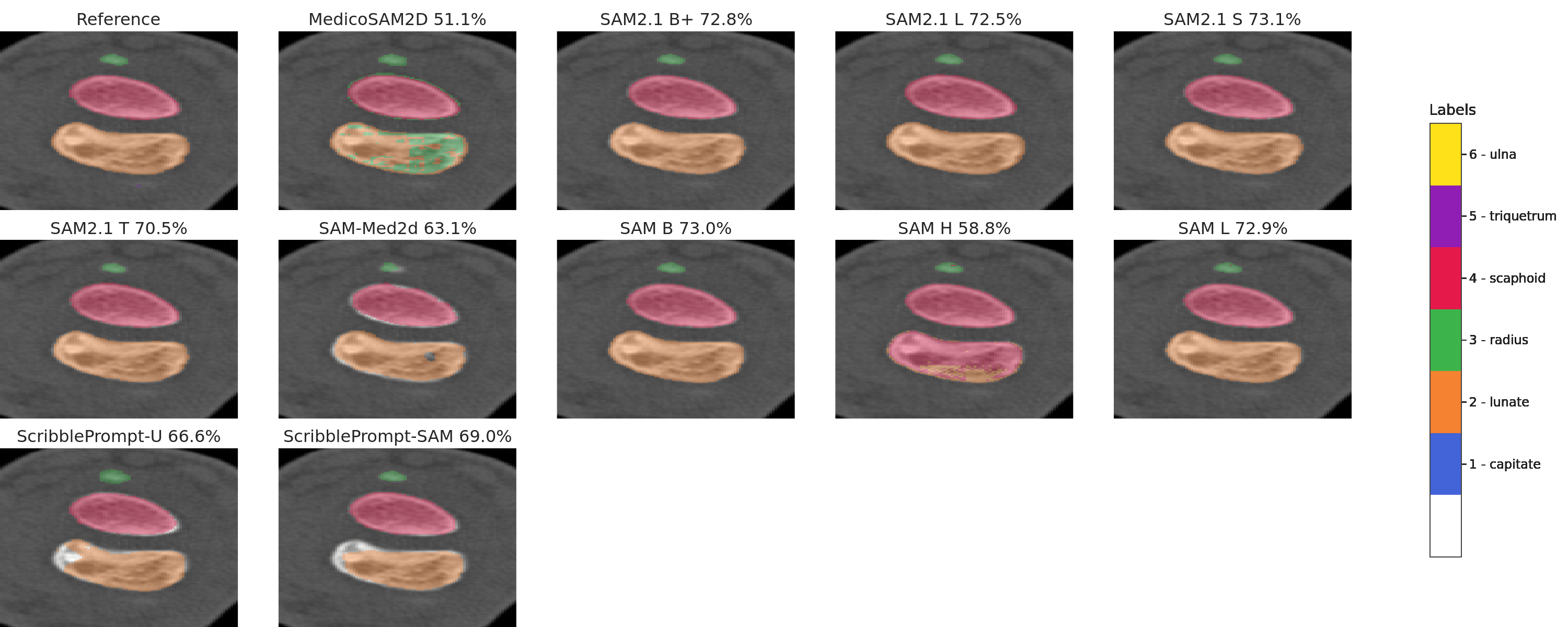}
          \subcaption{Center point}
        \end{subfigure} 
        \begin{subfigure}[t]{0.9\textwidth}
          \includegraphics[width=1\linewidth]{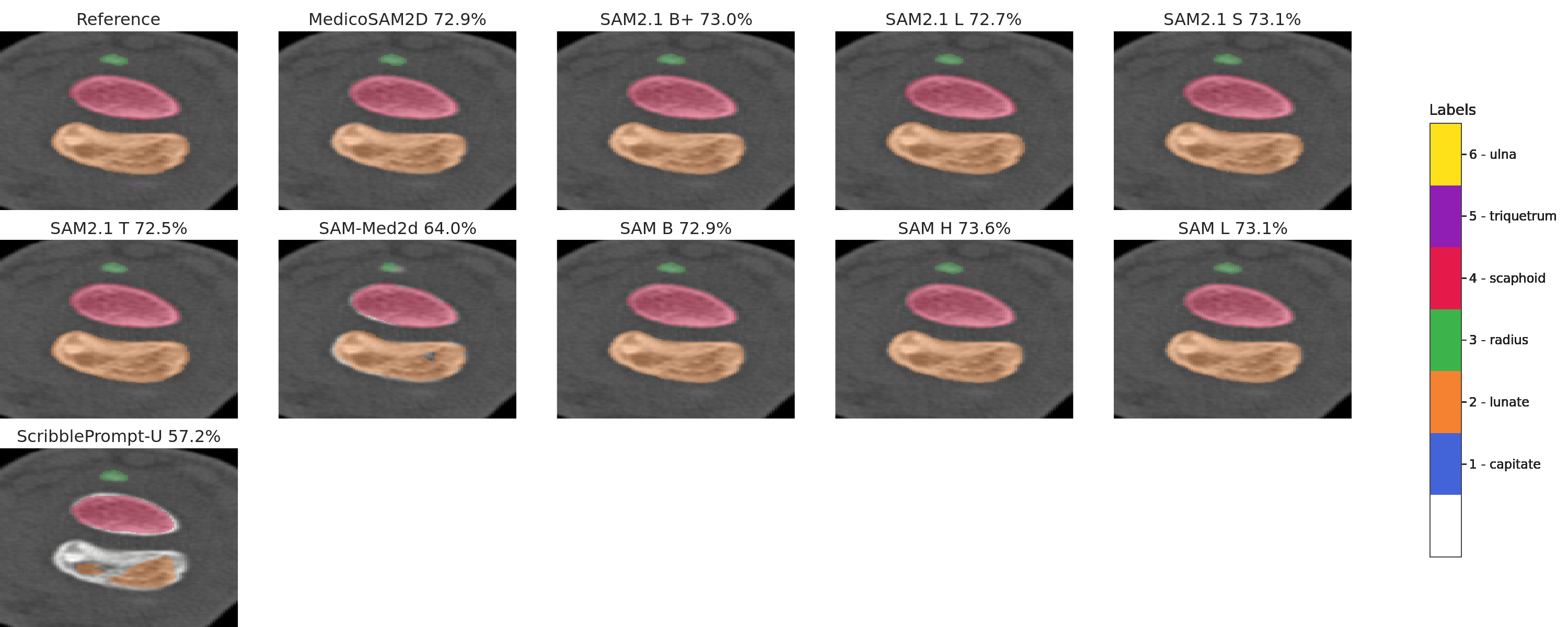}
          \subcaption{Combination}
        \end{subfigure} 
    \caption{Axial slice of Wrist with lowest DSC value (69.9\%) across 2D models. \newline
    {\protect\scriptsize The predictions are binary and were combined for visualization; as a result, some predicted regions may not appear because each pixel can only be assigned a single label.}}
    \label{fig:model_selection_examples_wrist}
\end{figure*}

\begin{figure*}[h!]
    \centering
        \begin{subfigure}[t]{0.98\textwidth}
          \includegraphics[width=1\linewidth]{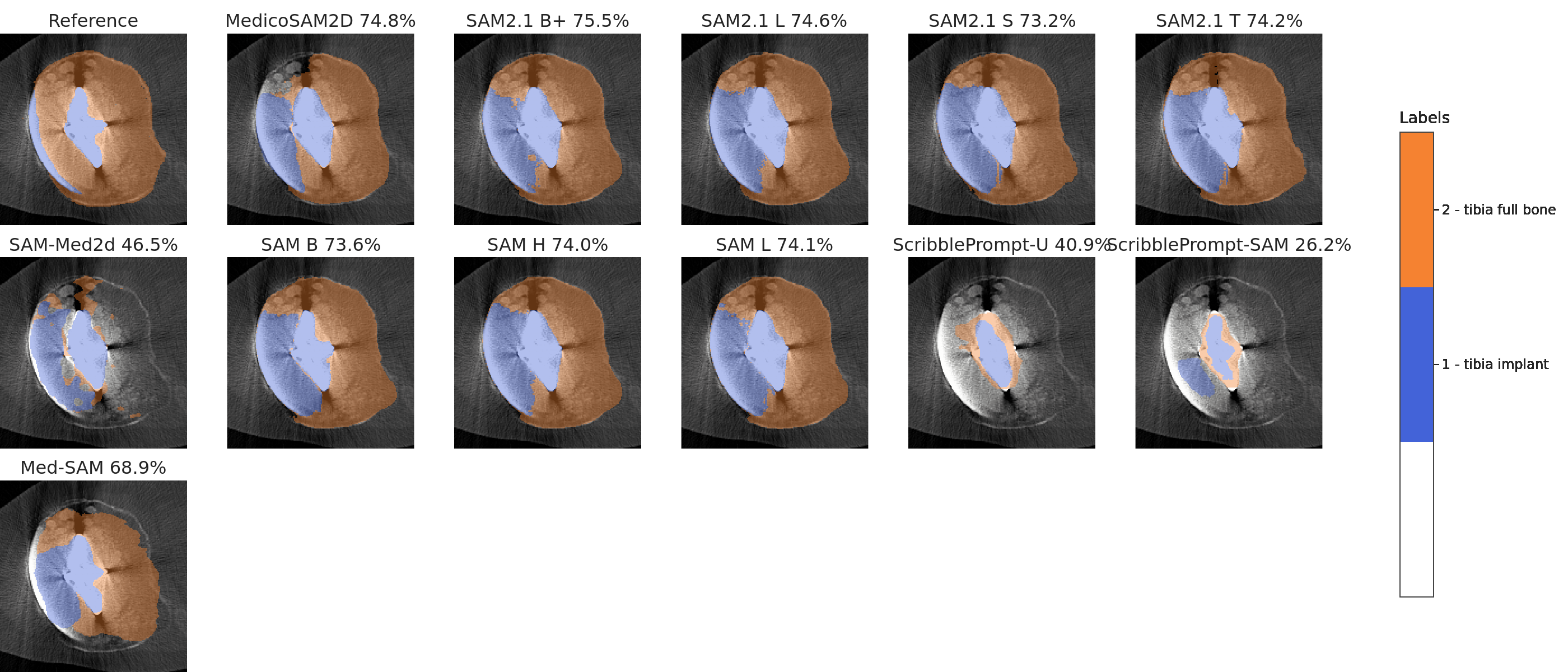}
          \subcaption{Bounding Box}
        \end{subfigure} 
        \begin{subfigure}[t]{0.98\textwidth}
          \includegraphics[width=1\linewidth]{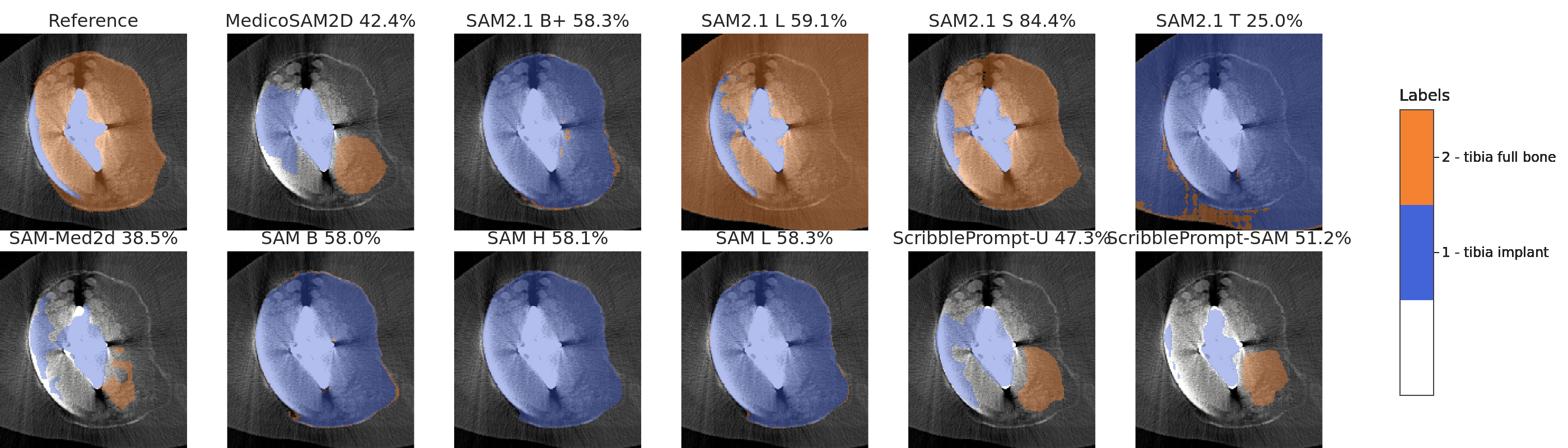}
          \subcaption{Center point}
        \end{subfigure} 
        \begin{subfigure}[t]{0.98\textwidth}
          \includegraphics[width=1\linewidth]{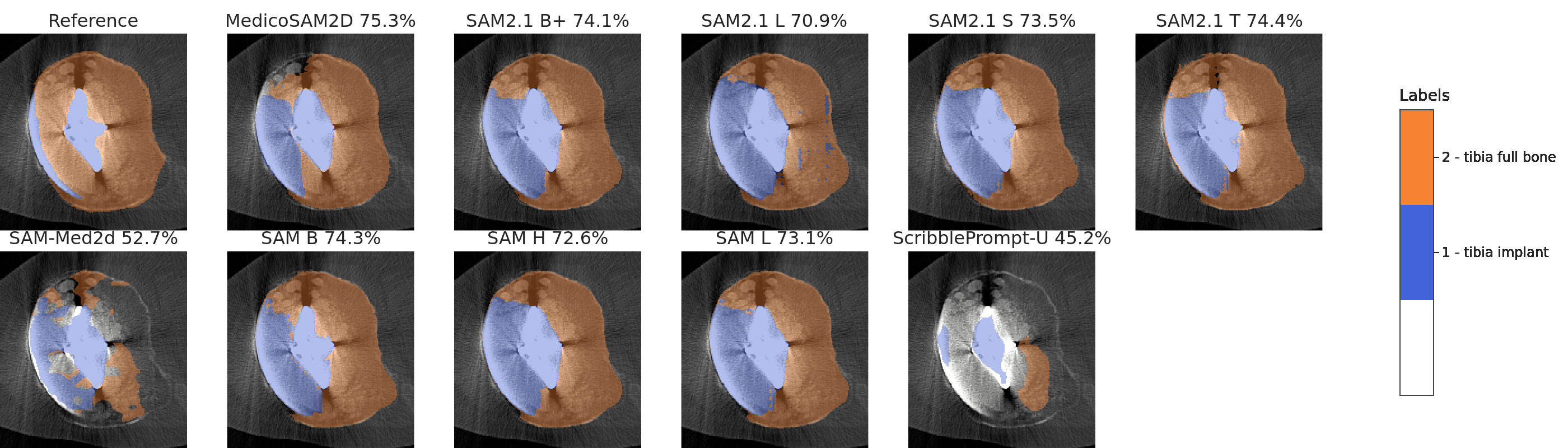}
          \subcaption{Combination}
        \end{subfigure} 
    \caption{Axial slice of Lower Leg with lowest DSC value (62.1\%) across 2D models. \newline
    {\protect\scriptsize The predictions are binary and were combined for visualization; as a result, some predicted regions may not appear because each pixel can only be assigned a single label.}}
    \label{fig:model_selection_examples_lowerleg}
\end{figure*}

\begin{figure*}[h!]
    \centering
        \begin{subfigure}[t]{0.98\textwidth}
          \includegraphics[width=1\linewidth]{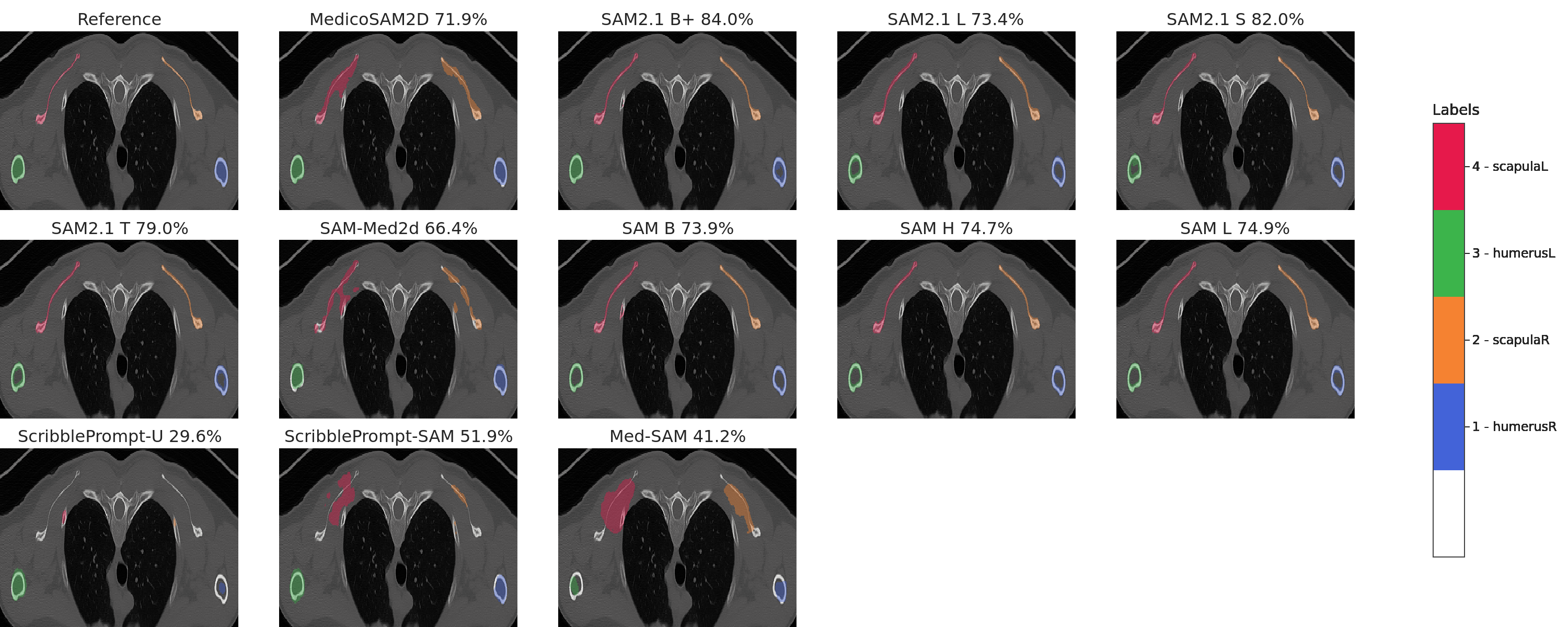}
          \subcaption{Bounding Box}
        \end{subfigure} 
        \begin{subfigure}[t]{0.98\textwidth}
          \includegraphics[width=1\linewidth]{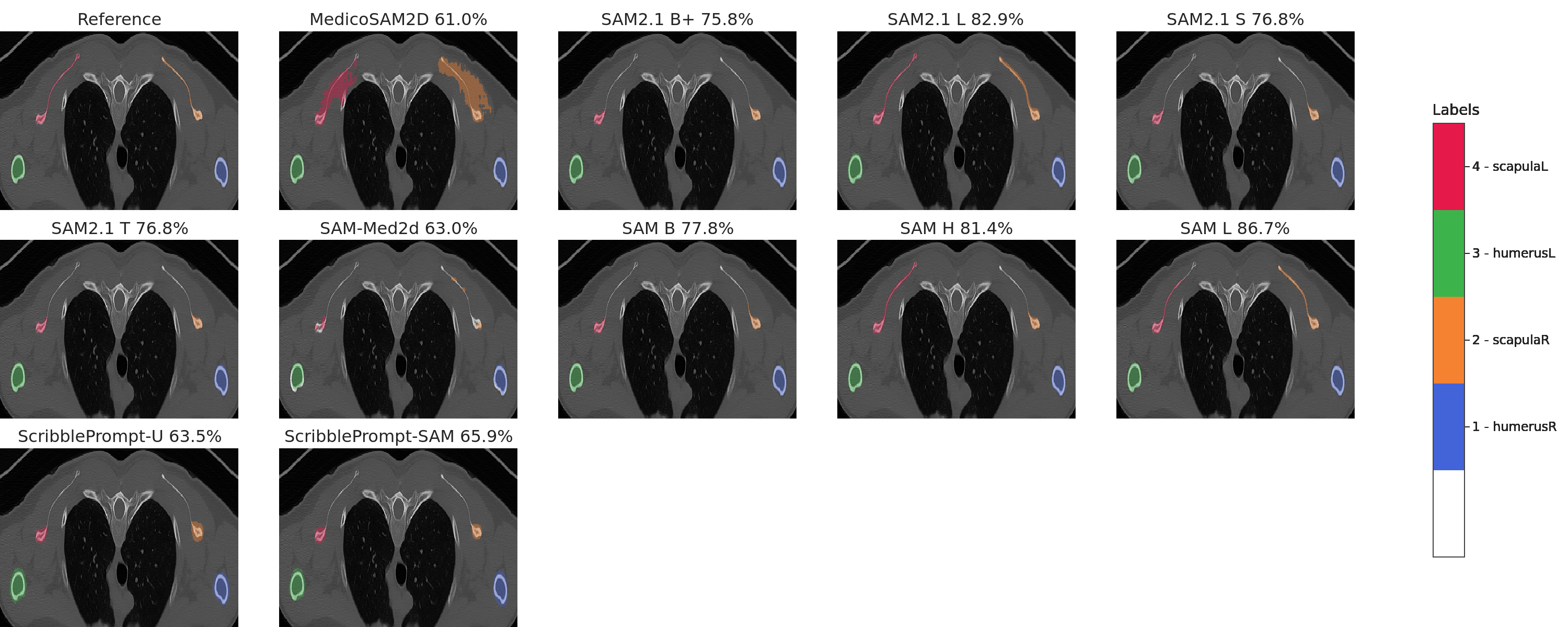}
          \subcaption{Center point}
        \end{subfigure} 
        \begin{subfigure}[t]{0.98\textwidth}
          \includegraphics[width=1\linewidth]{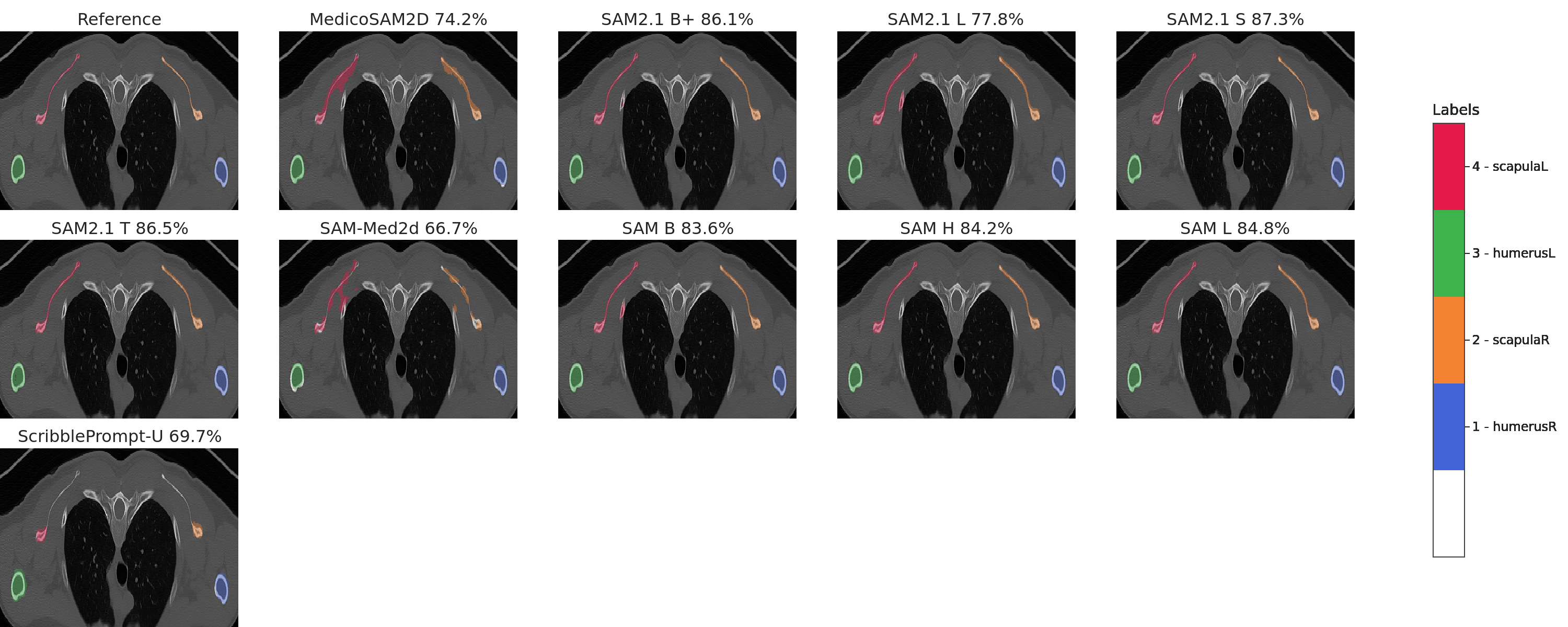}
          \subcaption{Combination}
        \end{subfigure} 
    \caption{Axial slice of Shoulder with lowest DSC value (75.1\%) across 2D models. \newline
    {\protect\scriptsize The predictions are binary and were combined for visualization; as a result, some predicted regions may not appear because each pixel can only be assigned a single label.}}
    \label{fig:model_selection_examples_shoulder}
\end{figure*}

\begin{figure*}[h!]
    \centering
        \begin{subfigure}[t]{0.9\textwidth}
          \includegraphics[width=1\linewidth]{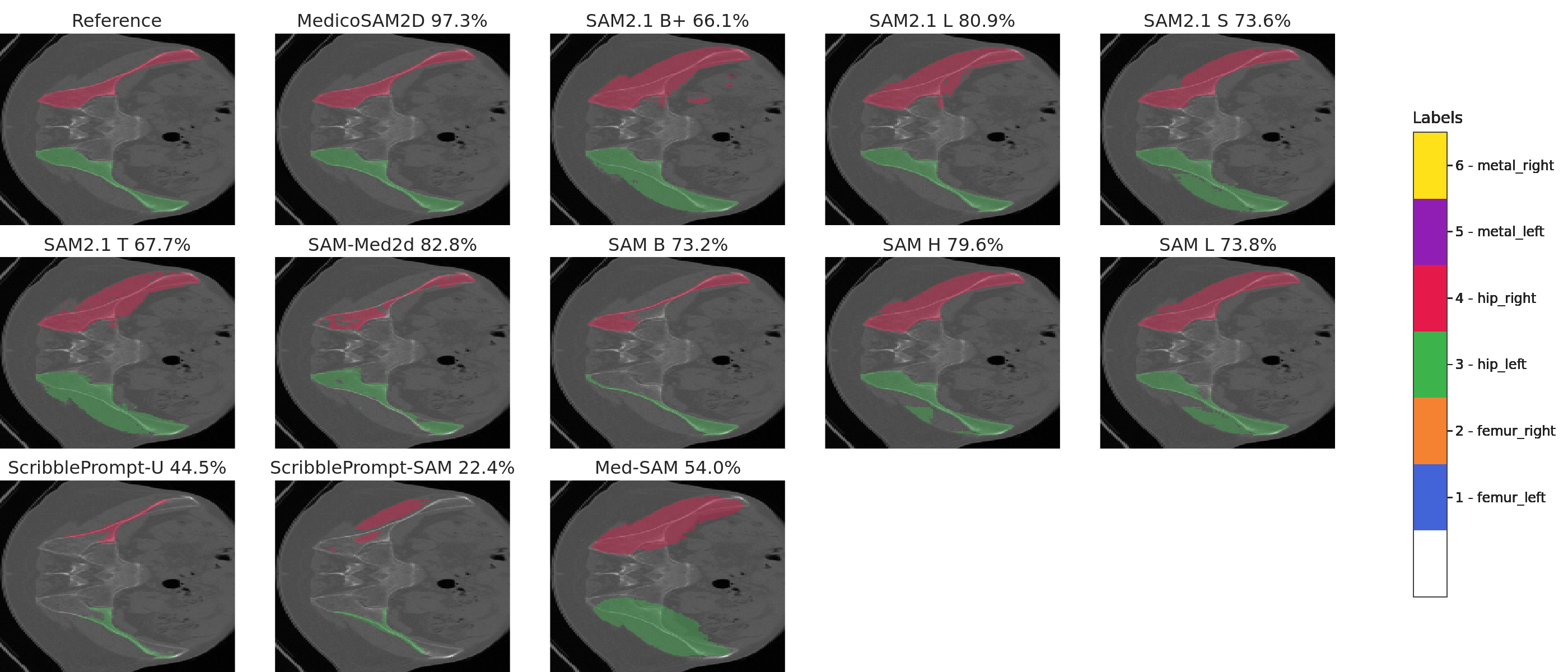}
          \subcaption{Bounding Box}
        \end{subfigure} 
        \begin{subfigure}[t]{0.9\textwidth}
          \includegraphics[width=1\linewidth]{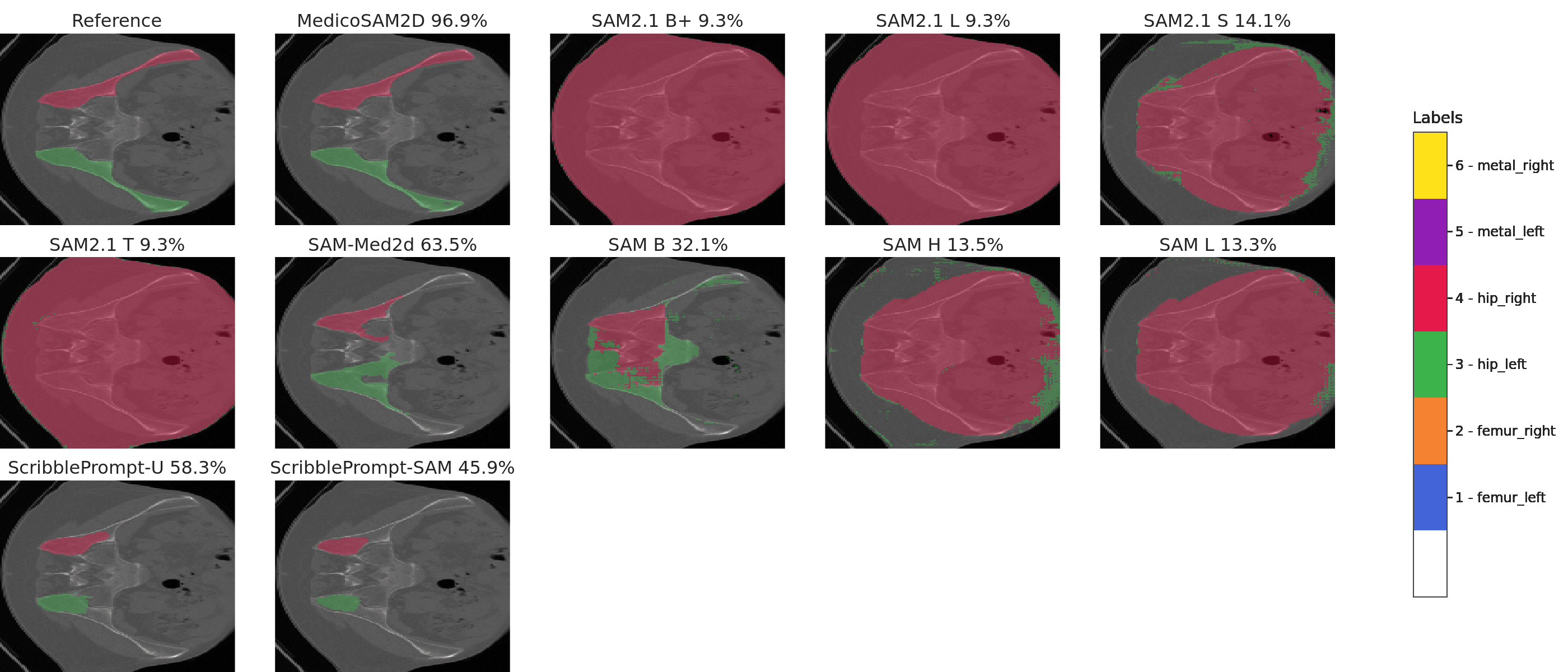}
          \subcaption{Center point}
        \end{subfigure} 
        \begin{subfigure}[t]{0.9\textwidth}
          \includegraphics[width=1\linewidth]{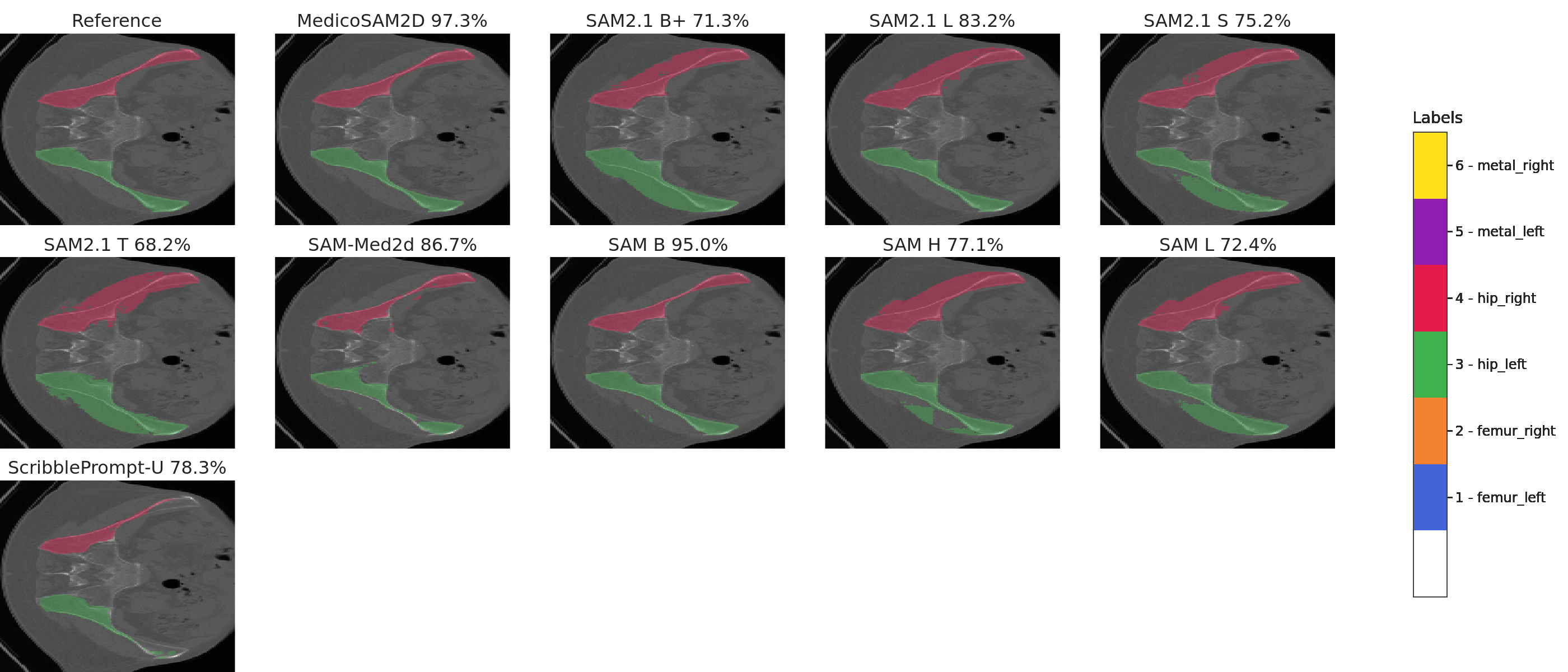}
          \subcaption{Combination}
        \end{subfigure} 
    \caption{Axial slice of Hip with lowest DSC value (58.4\%) across 2D models. \newline
    {\protect\scriptsize The predictions are binary and were combined for visualization; as a result, some predicted regions may not appear because each pixel can only be assigned a single label.}}
    \label{fig:model_selection_examples_hip}
\end{figure*}

Table \ref{tab:appendix_results_all_models} reports the segmentation performance of all 2D and 3D models, with the selected models (i.e., smallest Pareto-optimal models) highlighted as gray-shaded cells. This table is an extension of Table \ref{tab:pareto_front}, where the Pareto-optimal models per category and prompt type are summarized. The axial slices with the lowest average DSC values (i.e., negative examples) across all 2D models are shown in Figures \ref{fig:model_selection_examples_wrist} - \ref{fig:model_selection_examples_hip}.

\begin{table*}[hb!]
\centering
\caption{Segmentation performance of all 2D and 3D models per prompt type.\newline
\scriptsize{Gray-shaded cells indicate the smallest 2D and 3D Pareto-optimal models per prompt type. Omitted results (-) mean that the experiment was not performed, since it was not supported (see Table \ref{tab:model_prompt_overview}).}}
\label{tab:appendix_results_all_models}
\setlength{\tabcolsep}{3pt}
\renewcommand{\arraystretch}{1.3}
\resizebox{.95\textwidth}{!}{
\begin{tabular}{llc:ccc:ccc:ccc}
    \hline
    & \textbf{Model} & & \multicolumn{3}{:c:}{\textbf{Bounding Box} 2D \cblacksquare[0.2]{black} or 3D \cblacksquare[0.2]{white}} & \multicolumn{3}{c}{\textbf{Center Point} \cblackcircledot[0.25]{black} (2D) or \cblackcircledot[0.25]{white} (3D)} & \multicolumn{3}{:c}{\textbf{Combination} \cblacksquaredot[0.25]{black} (2D) or \cwhitesquaredot[0.25]{black} (3D)} \\ 
    & & Size & DSC $\uparrow$ & NSD $\uparrow$ & HD95 $\downarrow$ & DSC $\uparrow$ & NSD $\uparrow$ & HD95 $\downarrow$ & DSC $\uparrow$ & NSD $\uparrow$ & HD95 $\downarrow$ \\
    &  & (M) & (\%) & (\%) & (mm) & (\%) & (\%) & (mm) & (\%) & (\%) & (mm) \\    
    \hline \hline \rule{0pt}{2.6ex}

    & \multicolumn{11}{c}{\textbf{2D Models}} \\
    \hline
    
    \multirow{5}{*}{\rotatebox[origin=c]{90}{medical}} & Med-SAM & 94 & 66.89{\footnotesize±14.7} & 79.47{\footnotesize±11.2} & 4.59{\footnotesize±2.5} & - & - & - & - & - & - \\
    & MedicoSAM2D & 94 & 90.74{\footnotesize±7.7} & 97.36{\footnotesize±3.6} & 0.76{\footnotesize±0.9} & 77.46{\footnotesize±19.3} & 83.23{\footnotesize±18.4} & 5.00{\footnotesize±5.9} & 91.27{\footnotesize±7.4} & 97.74{\footnotesize±3.3} & 0.69{\footnotesize±0.8} \\
    & SAM-Med2d & 271 & 78.87{\footnotesize±13.8} & 91.43{\footnotesize±8.5} & 2.27{\footnotesize±1.8} & 73.69{\footnotesize±17.0} & 84.48{\footnotesize±14.9} & 5.35{\footnotesize±5.0} & 79.88{\footnotesize±13.2} & 91.59{\footnotesize±8.2} & 2.47{\footnotesize±2.1} \\
    & ScribblePrompt-SAM & 94 & 66.23{\footnotesize±20.1} & 78.25{\footnotesize±16.8} & 5.48{\footnotesize±4.1} & 74.19{\footnotesize±14.6} & 84.22{\footnotesize±12.6} & 6.30{\footnotesize±5.3} & - & - & - \\
    & ScribblePrompt-UNet & 4 & 66.14{\footnotesize±24.6} & 79.83{\footnotesize±18.2} & 5.87{\footnotesize±4.6} & 71.15{\footnotesize±16.2} & 80.85{\footnotesize±14.7} & 6.96{\footnotesize±5.3} & 72.46{\footnotesize±13.8} & 83.27{\footnotesize±12.0} & 5.91{\footnotesize±4.4} \\
    
    \hdashline
    \multirow{7}{*}{\rotatebox[origin=c]{90}{natural}} & SAM B & 94 & 89.03{\footnotesize±9.7} & 96.89{\footnotesize±4.7} & 1.10{\footnotesize±1.4} & \fcolorbox{gray!50}{gray!50}{85.43{\footnotesize±14.4}} & \fcolorbox{gray!50}{gray!50}{90.82{\footnotesize±13.0}} & \fcolorbox{gray!50}{gray!50}{4.83{\footnotesize±6.3}} & 91.80{\footnotesize±8.0} & 97.84{\footnotesize±4.4} & 1.07{\footnotesize±1.6} \\
    & SAM H & 641 & 90.44{\footnotesize±8.8} & 97.68{\footnotesize±3.9} & 0.84{\footnotesize±1.0} & 81.83{\footnotesize±17.7} & 87.61{\footnotesize±17.6} & 6.32{\footnotesize±9.4} & 91.56{\footnotesize±7.7} & 98.01{\footnotesize±3.8} & 0.78{\footnotesize±1.1} \\
    & SAM L & 312 & 89.53{\footnotesize±9.4} & 97.34{\footnotesize±4.2} & 0.91{\footnotesize±1.1} & 79.34{\footnotesize±20.0} & 84.83{\footnotesize±19.8} & 6.92{\footnotesize±10.7} & 91.41{\footnotesize±8.1} & 97.92{\footnotesize±4.3} & 0.80{\footnotesize±1.2} \\
    & SAM2.1 B+ & 81 & \fcolorbox{gray!50}{gray!50}{90.60{\footnotesize±8.1}} & \fcolorbox{gray!50}{gray!50}{97.84{\footnotesize±3.5}} & \fcolorbox{gray!50}{gray!50}{0.82{\footnotesize±1.0}} & 83.20{\footnotesize±16.5} & 88.87{\footnotesize±15.1} & 7.59{\footnotesize±9.7} & 91.98{\footnotesize±7.2} & 98.21{\footnotesize±3.6} & 0.73{\footnotesize±1.1} \\
    & SAM2.1 L & 224 & 88.39{\footnotesize±8.7} & 97.30{\footnotesize±3.9} & 0.92{\footnotesize±1.0} & 81.72{\footnotesize±17.4} & 88.44{\footnotesize±16.4} & 6.60{\footnotesize±10.7} & 90.90{\footnotesize±6.9} & 98.36{\footnotesize±3.2} & 0.69{\footnotesize±1.0} \\
    & SAM2.1 S & 46 & 89.40{\footnotesize±8.3} & 97.43{\footnotesize±3.8} & 0.91{\footnotesize±1.0} & 82.26{\footnotesize±15.6} & 88.46{\footnotesize±14.2} & 6.64{\footnotesize±8.4} & 91.51{\footnotesize±7.0} & 98.40{\footnotesize±3.3} & 0.69{\footnotesize±0.9} \\
    & SAM2.1 T & 39 & 89.57{\footnotesize±8.4} & 97.55{\footnotesize±3.8} & 0.88{\footnotesize±1.0} & 82.12{\footnotesize±16.3} & 88.62{\footnotesize±14.8} & 6.16{\footnotesize±8.6} & \fcolorbox{gray!50}{gray!50}{91.83{\footnotesize±6.9}} & \fcolorbox{gray!50}{gray!50}{98.38{\footnotesize±3.2}} & \fcolorbox{gray!50}{gray!50}{0.71{\footnotesize±1.0}} \\

    \hline
    & \multicolumn{11}{c}{\textbf{3D Models evaluated volumetric}} \\
    \hline
    
    \multirow{9}{*}{\rotatebox[origin=c]{90}{medical}} & Med-SAM2 & 39 & \fcolorbox{gray!50}{gray!50}{79.56{\footnotesize±11.1}} & \fcolorbox{gray!50}{gray!50}{80.25{\footnotesize±10.5}} & \fcolorbox{gray!50}{gray!50}{13.49{\footnotesize±11.1}} & - & - & - & - & - & - \\
    & MedicoSAM3D & 94 & 51.78{\footnotesize±15.1} & 52.73{\footnotesize±13.9} & 34.85{\footnotesize±13.6} & 54.39{\footnotesize±16.4} & 53.70{\footnotesize±15.4} & 36.89{\footnotesize±17.9} & 52.16{\footnotesize±15.0} & 53.04{\footnotesize±13.9} & 34.65{\footnotesize±14.0} \\
    & SAM-Med3d-Turbo-crop & 101 & - & - & - & 25.22{\footnotesize±9.0} & 18.20{\footnotesize±7.3} & 57.92{\footnotesize±9.7} & - & - & - \\
    & SAM-Med3d-Turbo-resample & 101 & - & - & - & 4.85{\footnotesize±5.3} & 4.20{\footnotesize±3.6} & 121.30{\footnotesize±20.9} & - & - & - \\
    & SAM-Med3d-crop & 101 & - & - & - & 28.76{\footnotesize±8.6} & 19.91{\footnotesize±5.8} & 52.93{\footnotesize±10.2} & - & - & - \\
    & SAM-Med3d-resample & 101 & - & - & - & 3.52{\footnotesize±2.5} & 3.02{\footnotesize±1.6} & 116.75{\footnotesize±20.0} & - & - & - \\
    & SegVol & 181 & - & - & - & 33.47{\footnotesize±13.5} & 32.97{\footnotesize±12.1} & 62.53{\footnotesize±22.4} & - & - & - \\
    & Vista3D & 218 & - & - & - & 25.70{\footnotesize±13.1} & 22.32{\footnotesize±11.6} & 58.14{\footnotesize±16.0} & - & - & - \\
    & nnInteractive & 102 & 76.15{\footnotesize±9.3} & 77.51{\footnotesize±9.2} & 25.36{\footnotesize±9.9} & \fcolorbox{gray!50}{gray!50}{69.40{\footnotesize±11.2}} & \fcolorbox{gray!50}{gray!50}{68.23{\footnotesize±12.0}} & \fcolorbox{gray!50}{gray!50}{30.98{\footnotesize±9.4}} & \fcolorbox{gray!50}{gray!50}{75.92{\footnotesize±9.4}} & \fcolorbox{gray!50}{gray!50}{76.60{\footnotesize±9.6}} & \fcolorbox{gray!50}{gray!50}{26.53{\footnotesize±10.3}} \\    

    \hdashline
    \multirow{4}{*}{\rotatebox[origin=c]{90}{natural}} & SAM2.1 B+ & 81 & 66.11{\footnotesize±10.1} & 66.59{\footnotesize±10.0} & 24.77{\footnotesize±18.1} & 53.38{\footnotesize±18.1} & 50.31{\footnotesize±19.6} & 48.14{\footnotesize±29.5} & 68.33{\footnotesize±9.4} & 67.86{\footnotesize±10.2} & 26.04{\footnotesize±18.2} \\
    & SAM2.1 L & 224 & 58.98{\footnotesize±11.8} & 57.27{\footnotesize±11.4} & 55.04{\footnotesize±30.1} & 48.41{\footnotesize±20.0} & 44.29{\footnotesize±20.9} & 69.02{\footnotesize±34.4} & 62.42{\footnotesize±11.3} & 59.79{\footnotesize±11.4} & 55.14{\footnotesize±29.8} \\
    & SAM2.1 S & 46 & 67.69{\footnotesize±10.2} & 68.48{\footnotesize±10.0} & 31.67{\footnotesize±21.6} & 56.90{\footnotesize±19.1} & 53.96{\footnotesize±20.2} & 47.84{\footnotesize±31.2} & 70.22{\footnotesize±10.1} & 69.88{\footnotesize±10.7} & 32.21{\footnotesize±22.0} \\
    & SAM2.1 T & 39 & 61.87{\footnotesize±11.9} & 63.40{\footnotesize±11.0} & 34.24{\footnotesize±22.6} & 54.74{\footnotesize±15.9} & 52.92{\footnotesize±16.9} & 46.40{\footnotesize±28.5} & 65.89{\footnotesize±9.8} & 66.34{\footnotesize±9.8} & 33.41{\footnotesize±21.4} \\
    \hline
\end{tabular}
}
\end{table*}

\clearpage

\clearpage

\clearpage

\clearpage

\clearpage

\section{Ablation Studies}
\label{sec:ablation_studies}

\subsection{SAM2 vs. SAM2.1}
\label{sec:sam2_vs_sam21}

Comparing SAM2 (released July 29, 2024) and SAM2.1 (released September 29, 2024) showed only marginal differences in segmentation performance for the same prompt type and model size (Table \ref{tab:ablation:sam2_vs_sam21}). Using the paired Wilcoxon signed-rank test with Bonferroni correction ($n=12$), none of the model pairs showed a statistically significant difference on any of the three metrics, except for the comparison between SAM2 T and SAM2.1 T prompted with bounding box.

\begin{table*}[h]
\centering
\setlength{\tabcolsep}{3pt}
\renewcommand{\arraystretch}{1.3}

\caption{Comparison of 2D segmentation performance of all model sizes of SAM2 and SAM2.1 per prompt type.\newline
\scriptsize{$\nearrow$ indicates that all metrics improve, whereas -- denotes no consistent trend across metrics. Asterisk ($*$) marks statistically significant differences between models ($p$-value$<0.05/12=0.0042$.}}
\label{tab:ablation:sam2_vs_sam21}

    \begin{tabular}{l:ccc:c:ccc}
    \hline
    Model & \multicolumn{3}{c:}{SAM2} & Trend & \multicolumn{3}{c}{SAM2.1} \\
     & DSC $\uparrow$ & NSD $\uparrow$ & HD95 $\downarrow$ & & DSC $\uparrow$ & NSD $\uparrow$ & HD95 $\downarrow$ \\
    & (\%) & (\%) & (mm) & & (\%) & (\%) & (mm)  \\
    \hline \hline

    \multicolumn{8}{c}{Bounding box \cblacksquare[0.25]{black}} \\
    \hline
    B+ & 90.40{\footnotesize±8.0} & 97.81{\footnotesize±3.4} & 0.81{\footnotesize±0.9} & $\nearrow$ & 90.60{\footnotesize±8.1} & 97.84{\footnotesize±3.5} & 0.82{\footnotesize±1.0} \\
    L & 88.23{\footnotesize±8.8} & 97.20{\footnotesize±4.0} & 0.93{\footnotesize±1.0} & $\nearrow$ & 88.39{\footnotesize±8.7} & 97.30{\footnotesize±3.9} & 0.92{\footnotesize±1.0} \\
    S & 89.06{\footnotesize±8.8} & 97.28{\footnotesize±4.0} & 0.93{\footnotesize±1.0} & $\nearrow$ & 89.40{\footnotesize±8.3} & 97.43{\footnotesize±3.8} & 0.91{\footnotesize±1.0} \\
    T & 89.07{\footnotesize±8.5} & 97.39{\footnotesize±3.9} & 0.92{\footnotesize±1.0} & $\nearrow$* & 89.57{\footnotesize±8.4} & 97.55{\footnotesize±3.8} & 0.88{\footnotesize±1.0} \\
    \hline
    
    \multicolumn{8}{c}{Center Point \cblackcircledot[0.25]{black}} \\
    \hline
    B+ & 83.39{\footnotesize±16.6} & 89.12{\footnotesize±15.3} & 7.45{\footnotesize±9.9} & -- & 83.20{\footnotesize±16.5} & 88.87{\footnotesize±15.1} & 7.59{\footnotesize±9.7} \\
    L & 78.45{\footnotesize±21.2} & 85.49{\footnotesize±21.0} & 8.30{\footnotesize±13.4} & $\nearrow$ & 81.72{\footnotesize±17.4} & 88.44{\footnotesize±16.4} & 6.60{\footnotesize±10.7} \\
    S & 81.51{\footnotesize±16.9} & 87.56{\footnotesize±16.3} & 7.22{\footnotesize±9.5} & $\nearrow$ & 82.26{\footnotesize±15.6} & 88.46{\footnotesize±14.2} & 6.64{\footnotesize±8.4} \\
    T & 80.38{\footnotesize±18.0} & 86.84{\footnotesize±16.9} & 7.53{\footnotesize±10.9} & $\nearrow$ & 82.12{\footnotesize±16.3} & 88.62{\footnotesize±14.8} & 6.16{\footnotesize±8.6} \\
    \hline
    
    \multicolumn{8}{c}{Combination \cblacksquaredot[0.25]{black}} \\
    \hline
    B+ & 91.82{\footnotesize±7.1} & 98.32{\footnotesize±3.3} & 0.70{\footnotesize±1.0} & -- & 91.98{\footnotesize±7.2} & 98.21{\footnotesize±3.6} & 0.73{\footnotesize±1.1} \\
    L & 90.78{\footnotesize±7.0} & 98.28{\footnotesize±3.1} & 0.68{\footnotesize±0.9} & -- & 90.90{\footnotesize±6.9} & 98.36{\footnotesize±3.2} & 0.69{\footnotesize±1.0} \\
    S & 91.48{\footnotesize±7.1} & 98.28{\footnotesize±3.5} & 0.71{\footnotesize±1.0} & -- & 91.51{\footnotesize±7.0} & 98.40{\footnotesize±3.3} & 0.69{\footnotesize±0.9} \\
    T & 91.33{\footnotesize±6.9} & 98.26{\footnotesize±3.3} & 0.73{\footnotesize±1.0} & $\nearrow$ & 91.83{\footnotesize±6.9} & 98.38{\footnotesize±3.2} & 0.71{\footnotesize±1.0} \\
    \hline
    \end{tabular}
\end{table*}

\newpage

\subsection{Limited vs. unlimited volume propagation}
\label{sec:limited_vs_unlimited}

SAM2.1 and Med-SAM2 generate volumetric predictions via memory bank and a propagation mechanism, which can be restricted to known start and/or end slices (see Table \ref{tab:model_prompt_overview}).
Although MedicoSAM3D also employs slice-by-slice propagation, the original method does not include a volume restriction for prediction and was therefore not including in our analysis.
Applying the prediction volume restriction requires knowing the object’s top and bottom slices, which adds two extra annotations to the required input information. However, limiting the propagation yielded better performance compared to unlimited propagation for all models (Table \ref{tab:ablation:limited_vs_unlimited_propagation}).

\begin{table*}[h]
\centering
\setlength{\tabcolsep}{3pt}
\renewcommand{\arraystretch}{1.3}

\caption{Comparison of volumetric prediction without (default setting) and with propagation limitation, per prompt type.}
\label{tab:ablation:limited_vs_unlimited_propagation}

    \begin{tabular}{l:ccc:ccc}
    \hline
    Model & \multicolumn{3}{c:}{unlimited propagation} & \multicolumn{3}{c}{limited propagation} \\
     & DSC $\uparrow$ & NSD $\uparrow$ & HD95 $\downarrow$ & DSC $\uparrow$ & NSD $\uparrow$ & HD95 $\downarrow$ \\
    & (\%) & (\%) & (mm) & (\%) & (\%) & (mm)  \\
    \hline \hline

    \multicolumn{7}{c}{Bounding box \cwhitesquare[0.25]{black}} \\
    \hline
    Med-SAM2 & 79.56{\footnotesize±11.1} & 80.25{\footnotesize±10.5} & 13.49{\footnotesize±11.1} & 84.00{\footnotesize±7.3} & 84.03{\footnotesize±7.6} & 7.76{\footnotesize±4.8} \\
    SAM2.1 B+ & 66.11{\footnotesize±10.1} & 66.59{\footnotesize±10.0} & 24.77{\footnotesize±18.1} & 83.47{\footnotesize±6.5} & 84.07{\footnotesize±6.9} & 6.75{\footnotesize±4.4} \\
    SAM2.1 L & 58.98{\footnotesize±11.8} & 57.27{\footnotesize±11.4} & 55.04{\footnotesize±30.1} & 80.97{\footnotesize±7.2} & 80.99{\footnotesize±7.4} & 8.41{\footnotesize±6.1} \\
    SAM2.1 S & 67.69{\footnotesize±10.2} & 68.48{\footnotesize±10.0} & 31.67{\footnotesize±21.6} & 82.70{\footnotesize±6.8} & 84.15{\footnotesize±6.8} & 7.85{\footnotesize±6.7} \\
    SAM2.1 T & 61.87{\footnotesize±11.9} & 63.40{\footnotesize±11.0} & 34.24{\footnotesize±22.6} & 81.50{\footnotesize±9.8} & 83.09{\footnotesize±9.3} & 8.91{\footnotesize±9.2} \\
    \hline

    \multicolumn{7}{c}{Center point \cwhitecircledot[0.25]{black}} \\
    \hline
    SAM2.1 B+ & 53.38{\footnotesize±18.1} & 50.31{\footnotesize±19.6} & 48.14{\footnotesize±29.5} & 69.15{\footnotesize±16.7} & 65.92{\footnotesize±19.2} & 27.68{\footnotesize±18.8} \\
    SAM2.1 L & 48.41{\footnotesize±20.0} & 44.29{\footnotesize±20.9} & 69.02{\footnotesize±34.4} & 67.98{\footnotesize±19.8} & 64.41{\footnotesize±22.4} & 25.60{\footnotesize±21.1} \\
    SAM2.1 S & 56.90{\footnotesize±19.1} & 53.96{\footnotesize±20.2} & 47.84{\footnotesize±31.2} & 70.76{\footnotesize±17.4} & 67.50{\footnotesize±19.7} & 22.96{\footnotesize±19.2} \\
    SAM2.1 T & 54.74{\footnotesize±15.9} & 52.92{\footnotesize±16.9} & 46.40{\footnotesize±28.5} & 73.38{\footnotesize±15.5} & 71.38{\footnotesize±17.2} & 22.13{\footnotesize±19.0} \\
    \hline

    \multicolumn{7}{c}{Combination \cwhitesquaredot[0.25]{black}} \\
    \hline
    SAM2.1 B+ & 68.33{\footnotesize±9.4} & 67.86{\footnotesize±10.2} & 26.04{\footnotesize±18.2} & 86.47{\footnotesize±4.7} & 86.87{\footnotesize±5.8} & 6.59{\footnotesize±4.2} \\
    SAM2.1 L & 62.42{\footnotesize±11.3} & 59.79{\footnotesize±11.4} & 55.14{\footnotesize±29.8} & 84.98{\footnotesize±5.3} & 84.44{\footnotesize±6.5} & 8.37{\footnotesize±6.2} \\
    SAM2.1 S & 70.22{\footnotesize±10.1} & 69.88{\footnotesize±10.7} & 32.21{\footnotesize±22.0} & 86.16{\footnotesize±5.7} & 86.77{\footnotesize±6.4} & 7.47{\footnotesize±6.0} \\
    SAM2.1 T & 65.89{\footnotesize±9.8} & 66.34{\footnotesize±9.8} & 33.41{\footnotesize±21.4} & 86.35{\footnotesize±5.6} & 87.23{\footnotesize±6.2} & 6.96{\footnotesize±5.7} \\
    \hline
    \end{tabular}

\end{table*}

\newpage

\subsection{Single vs. multiple initial slices}
\label{sec:single_vs_multiple_initial_slices}

For medical FMs (Med-SAM2, SegVol, Vista3D, nnInteractive), using multiple initial slices improved the performance for all prompt types, whereas for SAM2.1 models (except SAM2.1 L box-prompted), the performance was better for a single initial slice (Table \ref{tab:ablation:single_vs_multiple_slices}). nnInteractive box-prompted outperformed Med-SAM2, which was the Pareto-optimal model for the default settings (i.e., single initial slice). Using the paired Wilcoxon signed-rank test with Bonferroni correction ($n=18$), all model pairs showed a statistically significant difference in all three metrics, except for SAM2.1 L and SegVol.

\begin{table*}[h]
\centering
\setlength{\tabcolsep}{3pt}
\renewcommand{\arraystretch}{1.3}
\caption{Comparison of volumetric prediction with a single initial slice (default setting) or all initial slices, per prompt type. \newline
\scriptsize{$\nearrow$ indicates that all metrics improve, whereas $\searrow$ indicates that all metrics deteriorate. Asterisk ($*$) marks statistically significant differences between models.}}
\label{tab:ablation:single_vs_multiple_slices}

    \begin{tabular}{l:ccc:c:ccc}
    \hline
    Model & \multicolumn{3}{c:}{1 initial slice} & Trend & \multicolumn{3}{c}{$N_S$ initial slices} \\
     & DSC $\uparrow$ & NSD $\uparrow$ & HD95 $\downarrow$ & & DSC $\uparrow$ & NSD $\uparrow$ & HD95 $\downarrow$ \\
    & (\%) & (\%) & (mm) & & (\%) & (\%) & (mm)  \\
    \hline \hline

    \multicolumn{8}{c}{Bounding box \cwhitesquare[0.25]{black}} \\
    \hline
    Med-SAM2 & 84.00{\footnotesize±7.3} & 84.03{\footnotesize±7.6} & 7.76{\footnotesize±4.8} & $\nearrow$ * & 86.57{\footnotesize±6.3} & 87.54{\footnotesize±6.3} & 4.75{\footnotesize±3.1} \\
    SAM2.1 B+ & 66.11{\footnotesize±10.1} & 66.59{\footnotesize±10.0} & 24.77{\footnotesize±18.1} & $\searrow$ * & 59.80{\footnotesize±9.0} & 60.19{\footnotesize±7.2} & 38.10{\footnotesize±20.4} \\
    SAM2.1 L & 58.98{\footnotesize±11.8} & 57.27{\footnotesize±11.4} & 55.04{\footnotesize±30.1} & $\nearrow$ & 60.01{\footnotesize±11.9} & 59.17{\footnotesize±10.5} & 51.21{\footnotesize±31.3} \\
    SAM2.1 S & 67.69{\footnotesize±10.2} & 68.48{\footnotesize±10.0} & 31.67{\footnotesize±21.6} & $\searrow$ * & 60.84{\footnotesize±9.2} & 60.10{\footnotesize±8.1} & 54.76{\footnotesize±25.9} \\
    SAM2.1 T & 61.87{\footnotesize±11.9} & 63.40{\footnotesize±11.0} & 34.24{\footnotesize±22.6} & $\searrow$ * & 55.64{\footnotesize±8.9} & 57.06{\footnotesize±8.2} & 46.79{\footnotesize±24.4} \\
    nnInteractive & 76.15{\footnotesize±9.3} & 77.51{\footnotesize±9.2} & 25.36{\footnotesize±9.9} & $\nearrow$ * & 90.02{\footnotesize±5.8} & 92.08{\footnotesize±5.9} & 2.69{\footnotesize±1.8} \\
    \hline

    \multicolumn{8}{c}{Center point \cwhitecircledot[0.25]{black}} \\
    \hline
    SAM2.1 B+ & 53.38{\footnotesize±18.1} & 50.31{\footnotesize±19.6} & 48.14{\footnotesize±29.5} & $\searrow$ * & 41.87{\footnotesize±13.8} & 40.96{\footnotesize±12.6} & 57.80{\footnotesize±16.3} \\
    SAM2.1 L & 48.41{\footnotesize±20.0} & 44.29{\footnotesize±20.9} & 69.02{\footnotesize±34.4} & $\searrow$ & 38.92{\footnotesize±22.1} & 37.36{\footnotesize±20.2} & 74.30{\footnotesize±38.0} \\
    SAM2.1 S & 56.90{\footnotesize±19.1} & 53.96{\footnotesize±20.2} & 47.84{\footnotesize±31.2} & $\searrow$ * & 44.84{\footnotesize±16.8} & 42.52{\footnotesize±14.3} & 71.08{\footnotesize±29.5} \\
    SAM2.1 T & 54.74{\footnotesize±15.9} & 52.92{\footnotesize±16.9} & 46.40{\footnotesize±28.5} & $\searrow$ * & 42.53{\footnotesize±15.1} & 43.22{\footnotesize±14.0} & 62.72{\footnotesize±30.3} \\
    SegVol & 33.47{\footnotesize±13.5} & 32.97{\footnotesize±12.1} & 62.53{\footnotesize±22.4} & $\nearrow$ & 38.32{\footnotesize±14.2} & 37.42{\footnotesize±14.2} & 19.86{\footnotesize±8.4} \\
    Vista3D & 25.70{\footnotesize±13.1} & 22.32{\footnotesize±11.6} & 58.14{\footnotesize±16.0} & $\nearrow$ * & 44.98{\footnotesize±14.8} & 35.88{\footnotesize±12.9} & 28.00{\footnotesize±14.7} \\
    nnInteractive & 69.40{\footnotesize±11.2} & 68.23{\footnotesize±12.0} & 30.98{\footnotesize±9.4} & $\nearrow$ * & 85.67{\footnotesize±7.1} & 82.89{\footnotesize±9.7} & 4.44{\footnotesize±2.7} \\
    \hline

    \multicolumn{8}{c}{Combination \cwhitesquaredot[0.25]{black}} \\
    \hline
    SAM2.1 B+ & 68.33{\footnotesize±9.4} & 67.86{\footnotesize±10.2} & 26.04{\footnotesize±18.2} & $\searrow$ * & 60.65{\footnotesize±8.4} & 60.91{\footnotesize±7.1} & 39.73{\footnotesize±20.4} \\
    SAM2.1 L & 62.42{\footnotesize±11.3} & 59.79{\footnotesize±11.4} & 55.14{\footnotesize±29.8} & $\searrow$ & 62.37{\footnotesize±11.3} & 62.10{\footnotesize±10.8} & 50.94{\footnotesize±30.1} \\
    SAM2.1 S & 70.22{\footnotesize±10.1} & 69.88{\footnotesize±10.7} & 32.21{\footnotesize±22.0} & $\searrow$ * & 61.01{\footnotesize±8.7} & 60.67{\footnotesize±8.1} & 56.47{\footnotesize±25.8} \\
    SAM2.1 T & 65.89{\footnotesize±9.8} & 66.34{\footnotesize±9.8} & 33.41{\footnotesize±21.4} & $\searrow$ * & 58.48{\footnotesize±8.4} & 59.77{\footnotesize±7.5} & 46.36{\footnotesize±24.8} \\
    nnInteractive & 75.92{\footnotesize±9.4} & 76.60{\footnotesize±9.6} & 26.53{\footnotesize±10.3} & $\nearrow$ * & 89.81{\footnotesize±5.2} & 91.37{\footnotesize±6.3} & 2.70{\footnotesize±1.7} \\
    \hline
    \end{tabular}

\end{table*}

\newpage

\subsection{Single vs. multiple prompts}
\label{sec:single_vs_multiple_prompts}

The support for multiple prompts varies for 3D models, with more models supporting multiple point (see Table  \ref{tab:model_prompt_overview}). The multiple prompt setting was equivalent to the default setting for 2D models. Comparing the volumetric segmentation performance for single vs. multiple prompts per prompt type showed only marginal differences per model (Table \ref{tab:ablation:single_vs_multiple_prompts}). Using the paired Wilcoxon signed-rank test with Bonferroni correction ($n=6$ for bounding box, $n=24$ for center point), only MedicoSAM3D showed statistically significant difference in all three metrics.

\begin{table*}[h]
\centering
\setlength{\tabcolsep}{3pt}
\renewcommand{\arraystretch}{1.3}

\caption{Comparison of volumetric prediction with a single (default setting) or multiple (up to 5) prompts, per prompt type. \scriptsize{$\nearrow$ indicates that all metrics improve, $\searrow$ indicates that all metrics deteriorate, whereas -- denotes no consistent trend across metrics. An asterisk (*) marks statistically significant differences between models.}}
\label{tab:ablation:single_vs_multiple_prompts}

    \begin{tabular}{l:ccc:c:ccc}
    \hline
    Model & \multicolumn{3}{c:}{1 prompt} & Trend & \multicolumn{3}{c}{up to 5 prompts} \\
     & DSC $\uparrow$ & NSD $\uparrow$ & HD95 $\downarrow$ & & DSC $\uparrow$ & NSD $\uparrow$ & HD95 $\downarrow$ \\
    & (\%) & (\%) & (mm) & (\%) & (\%) & (mm)  \\
    \hline \hline

    &\multicolumn{7}{c}{Bounding box } \\[-3mm]
    & \multicolumn{3}{c}{\cwhitesquare[0.25]{black}} & &  \multicolumn{3}{c}{\cblacksquare[0.25]{black}} \\
    \hline
    MedicoSAM3D & 51.78{\footnotesize±15.1} & 52.73{\footnotesize±13.9} & 34.85{\footnotesize±13.6} & $\searrow$* & 51.63{\footnotesize±15.1} & 52.59{\footnotesize±14.0} & 35.15{\footnotesize±13.9} \\
    nnInteractive & 76.15{\footnotesize±9.3} & 77.51{\footnotesize±9.2} & 25.36{\footnotesize±9.9} & -- & 76.63{\footnotesize±8.7} & 78.09{\footnotesize±8.6} & 25.26{\footnotesize±9.7} \\
     \hline

    & \multicolumn{7}{c}{Center point} \\[-3mm]
    & \multicolumn{3}{c}{ \cwhitecircledot[0.25]{black}} & &  \multicolumn{3}{c}{ \cblackcircledot[0.25]{black}} \\ 
    \hline
    MedicoSAM3D & 54.39{\footnotesize±16.4} & 53.70{\footnotesize±15.4} & 36.89{\footnotesize±17.9} & $\searrow$ * & 54.11{\footnotesize±16.5} & 53.45{\footnotesize±15.6} & 37.32{\footnotesize±18.3} \\
    SAM2.1 B+ & 53.38{\footnotesize±18.1} & 50.31{\footnotesize±19.6} & 48.14{\footnotesize±29.5} & -- & 53.27{\footnotesize±18.6} & 50.28{\footnotesize±20.1} & 47.91{\footnotesize±29.8} \\
    SAM2.1 L & 48.41{\footnotesize±20.0} & 44.29{\footnotesize±20.9} & 69.02{\footnotesize±34.4} & -- & 48.43{\footnotesize±20.2} & 44.35{\footnotesize±21.1} & 68.72{\footnotesize±34.2} \\
    SAM2.1 S & 56.90{\footnotesize±19.1} & 53.96{\footnotesize±20.2} & 47.84{\footnotesize±31.2} & $\searrow$ & 56.76{\footnotesize±19.4} & 53.83{\footnotesize±20.6} & 48.38{\footnotesize±31.7} \\
    SAM2.1 T & 54.74{\footnotesize±15.9} & 52.92{\footnotesize±16.9} & 46.40{\footnotesize±28.5} & $\searrow$ & 54.23{\footnotesize±16.5} & 52.37{\footnotesize±17.4} & 47.53{\footnotesize±29.2} \\
    SegVol & 33.47{\footnotesize±13.5} & 32.97{\footnotesize±12.1} & 62.53{\footnotesize±22.4} & -- & 33.63{\footnotesize±13.4} & 33.12{\footnotesize±12.0} & 62.90{\footnotesize±22.6} \\
    Vista3D & 25.70{\footnotesize±13.1} & 22.32{\footnotesize±11.6} & 58.14{\footnotesize±16.0} & $\searrow$ & 25.63{\footnotesize±13.0} & 22.31{\footnotesize±11.5} & 58.34{\footnotesize±16.1} \\
    nnInteractive & 69.40{\footnotesize±11.2} & 68.23{\footnotesize±12.0} & 30.98{\footnotesize±9.4} & $\nearrow$ & 69.66{\footnotesize±10.8} & 68.50{\footnotesize±11.6} & 30.80{\footnotesize±9.2} \\
    \bottomrule
    \end{tabular}
\end{table*}

\newpage

\section{Comparison segmentation with reference and human prompts} \label{app:automatic_vs_human_prompts}

Table \ref{tab:automatic_vs_human_performance} shows the average difference for the performance of FMs prompted with reference and human prompts. The paired Wilcoxon signed-rank test showed a statistically significant difference for the overall comparison of 2D and 3D models, with $p$-value smaller than the Bonferroni-corrected $\alpha$-value ($0.05/6=0.0083$).

\begin{table*}[h!]
\centering
\caption{Difference in segmentation performance between reference and human prompts, per prompt type.\newline
{\protect\scriptsize The models with the least difference per prompt type are highlighted in bold. The selected models are the smallest Pareto-optimal models prompted with reference prompts per category highlighted in bold in Table \ref{tab:pareto_front}}.}
\label{tab:automatic_vs_human_performance}
\setlength{\tabcolsep}{3pt}
\renewcommand{\arraystretch}{1.3}
\resizebox{.98\textwidth}{!}{
\begin{tabular}{llc:ccc:ccc:ccc}
    \hline
    & \textbf{Model} & & \multicolumn{3}{:c:}{\textbf{Bounding Box} 2D \cblacksquare[0.2]{black} or 3D \cblacksquare[0.2]{white}} & \multicolumn{3}{c}{\textbf{Center Point} \cblackcircledot[0.25]{black} (2D) or \cblackcircledot[0.25]{white} (3D)} & \multicolumn{3}{:c}{\textbf{Combination} \cblacksquaredot[0.25]{black} (2D) or \cwhitesquaredot[0.25]{black} (3D)} \\ 
    & & Size & DSC $\uparrow$ & NSD $\uparrow$ & HD95 $\downarrow$ & DSC $\uparrow$ & NSD $\uparrow$ & HD95 $\downarrow$ & DSC $\uparrow$ & NSD $\uparrow$ & HD95 $\downarrow$ \\
    &  & (M) & (\%) & (\%) & (mm) & (\%) & (\%) & (mm) & (\%) & (\%) & (mm) \\    
    \hline \hline \rule{0pt}{2.6ex}
    
    & \multicolumn{11}{c}{\textbf{2D Models}} \\
    \hline
    \multirow{2}{*}{\rotatebox[origin=c]{25}{\footnotesize{medical}}}& MedicoSAM2D & 94 & 3.39 ± 6.3 & 1.41 ± 4.4 & -0.41 ± 1.1 & \textbf{1.24} ± 5.5 & 0.28 ± 3.5 & -0.16 ± 2.4 & 3.47 ± 6.3 & 2.15 ± 5.3 & -0.62 ± 1.5 \\
    & ScribblePrompt-SAM & 94 & - & - & - & 1.33 ± 6.3 & 0.19 ± 3.6 & \textbf{-0.05} ± 2.2 & - & - & - \\
    \multirow{3}{*}{\rotatebox[origin=c]{25}{\footnotesize{natural}}} & SAM B & 94 & - & - & - & 1.39 ± 5.5 & \textbf{0.13} ± 2.9 & 0.16 ± 3.2 & - & - & - \\
    & SAM2.1 B+ & 81 & \textbf{2.05} ± 5.8 & \textbf{0.92} ± 3.7 & \textbf{-0.31} ± 1.0 & - & - & - & - & - & - \\
    & SAM2.1 T & 39 & - & - & - & - & - & - & \textbf{1.64} ± 5.2 & \textbf{0.99} ± 4.1 & \textbf{-0.36} ± 1.2 \\
    \hdashline
    & Average per prompt type & & 2.72 ± 6.1 & 1.16 ± 4.1 & -0.36 ± 1.1 & 1.32 ± 5.8 & 0.20 ± 3.3 & -0.02 ± 2.6 & 2.56 ± 5.8 & 1.57 ± 4.8 & -0.49 ± 1.4 \\
    & Average 2D Models & & \multicolumn{3}{c}{2.07 ± 1.0 \% DSC ($p<0.001$)} & \multicolumn{3}{c}{0.87 ± 0.7 \% NSD ($p<0.001$)} & \multicolumn{3}{c}{-0.25 ± 0.3 mm HD95 ($p<0.001$)} \\
    \hline
    & \multicolumn{11}{c}{\textbf{3D Models evaluated volumetric}} \\
    \hline
    \multirow{2}{*}{\rotatebox[origin=c]{25}{\footnotesize{medical}}} & Med-SAM2 & 39 & 76.80{\footnotesize±13.5} & 79.27{\footnotesize±11.2} & 14.46{\footnotesize±11.8} & - & - & - & - & - & - \\
    & nnInteractive & 102 & - & - & - & 68.12{\footnotesize±12.6} & 68.63{\footnotesize±11.5} & 30.10{\footnotesize±8.8} & 75.59{\footnotesize±10.6} & 77.29{\footnotesize±9.1} & 25.65{\footnotesize±9.5} \\
    \multirow{2}{*}{\rotatebox[origin=c]{25}{\footnotesize{natural}}} & SAM2.1 S & 46 & 65.93{\footnotesize±11.6} & 67.83{\footnotesize±10.2} & 32.71{\footnotesize±21.6} & - & - & - & 68.80{\footnotesize±11.2} & 69.19{\footnotesize±10.9} & 33.88{\footnotesize±22.4} \\
    & SAM2.1 T & 39 & - & - & - & 53.72{\footnotesize±16.3} & 52.93{\footnotesize±16.5} & 46.84{\footnotesize±27.8} & - & - & - \\
    \hdashline
    & Average per prompt type & & 1.76 ± 5.8 & 0.96 ± 4.8 & -0.89 ± 10.2 & 0.80 ± 7.4 & 0.07 ± 6.6 & 0.20 ± 7.8  & 0.63 ± 4.5 & 0.40 ± 4.2 & -0.48 ± 8.2\\
    & Average 3D Models & & \multicolumn{3}{c}{1.06 ± 0.7 \% DSC ($p<0.001$)} & \multicolumn{3}{c}{0.47 ± 0.6 \% NSD ($p<0.001$)} & \multicolumn{3}{c}{-0.39 ± 0.7 mm HD95 ($p<0.001$)} \\
    \hline
\end{tabular}
}
\end{table*}

\clearpage

\end{document}